\documentclass[]{mihoyo}

\usepackage{hyperref}
\usepackage{cleveref}
\usepackage{verbatim}

\usepackage{graphicx}
\usepackage{floatrow}
\usepackage{subcaption}
\usepackage{listings}
\usepackage{algorithm}
\usepackage{mathtools}
\usepackage{amssymb}
\usepackage{wrapfig}  
\usepackage{xspace}
\usepackage{pifont}

\usepackage{subcaption} %

\usepackage[toc,page,header]{appendix}

\usepackage{minitoc}

\newcommand{\ourmodel}{Gamer\xspace}

\title{\ourmodel: A Generalist Vision-Language-Action Foundation Model for Long-Horizon Real-Time Game Agent}
\author{%
\parbox{\textwidth}{\centering
miHoYo Honkai AI R\&D Team
}}

\affiliation{%
\parbox{\textwidth}{\centering\small
See Contributions section for a full author list
}}

\renewcommand{\thefootnote}{\arabic{footnote}}  

\abstract{
We introduce \ourmodel, a generalist agent framework built for long-horizon tasks in complex interactive environments. Its end-to-end architecture combines perception, memory management, and multi-step reasoning. A shared training and deployment pipeline supports interaction with both conventional 2D graphical interfaces and complex 3D simulated worlds. We first continue pre-training on a dataset constructed from tens of thousands of hours of device-level keyboard and mouse interaction. These trajectories span simulated games and embodied environments, allowing the model to learn environmental dynamics across both physical and virtual domains. Sparse-thinking instruction tuning then substantially reduces redundant computation while preserving the reasoning needed for complex decisions. We train the policy for long-horizon interaction with a stable multi-turn reinforcement learning method that addresses finite context windows, accumulated rollout errors, and sparse rewards. For real-time interaction, a cloud–edge inference engine serves numerous simulated game environments concurrently, supports unbounded-context execution, and keeps online inference latency low. Across multiple challenging 3D open-world games, \ourmodel demonstrates solid generalization and operational efficiency comparable to skilled players. It also achieves competitive results on a range of conventional 2D game benchmarks. These results suggest that even repetitive keyboard and mouse trajectories can support substantial gains from scaling data and learning algorithms. Training on these trajectories helps \ourmodel learn complex skills more steadily in environments with both physical and digital properties. This approach offers a path toward generalist agents that can interact across digital and physical worlds.
}
\checkdata[Corresponding]{\url{yinhe.zheng@mihoyo.com}, \url{xiaoboxia@ustc.edu.cn}, \url{sqren@ustc.edu.cn}}
\checkdata[Project Page]{\url{https://mihoyo-gamer.github.io/}}

\begin{document}
\maketitle

\addtocontents{toc}{\protect\setcounter{tocdepth}{-1}}
\section{Introduction}

Multimodal large language models (MLLMs)~\cite{qwen2vl,qwenvl,dong2025seedvl,internvl2,internvl2.5,luo2026next} are rapidly evolving from static input processors into interactive agents that can operate in digital and physical environments. In particular, vision-language-action (VLA) models~\cite{wang2024gui,nguyen2025gui,luo2025gui,gamescaling} offer a promising path toward embodied artificial general intelligence. However, high trial-and-error costs and the lack of environmental controllability significantly hinder both the evaluation and algorithm scaling of real-world physical agents.

Modern large-scale games~\cite{fan2022minedojo,baker2022vpt, gameverse, wang2026gamehorizon,towards,videoarena, gameworld2026,luo2026ai} provide an ideal testbed for developing generalist VLA agents and studying their scaling behaviors because they combine realistic complexity with rigorous experimental control. Compared with traditional 2D games, web interfaces, and simplified simulators, contemporary games require agents to parse complex visual scenes, navigate intricate 3D layouts, and advance multi-stage quests. Agents must continually make strategic decisions and manage resources under partial observability and delayed feedback, all through fine-grained, low-level keyboard and mouse actions. Tasks may last hours or longer, requiring reliable memory, planning, and decision-making throughout. Such environments therefore serve as an uncompromising benchmark for coordinating perception, reasoning, memory, and control.

\begin{figure*}[!t]
\centering
\includegraphics[width=0.91\textwidth]{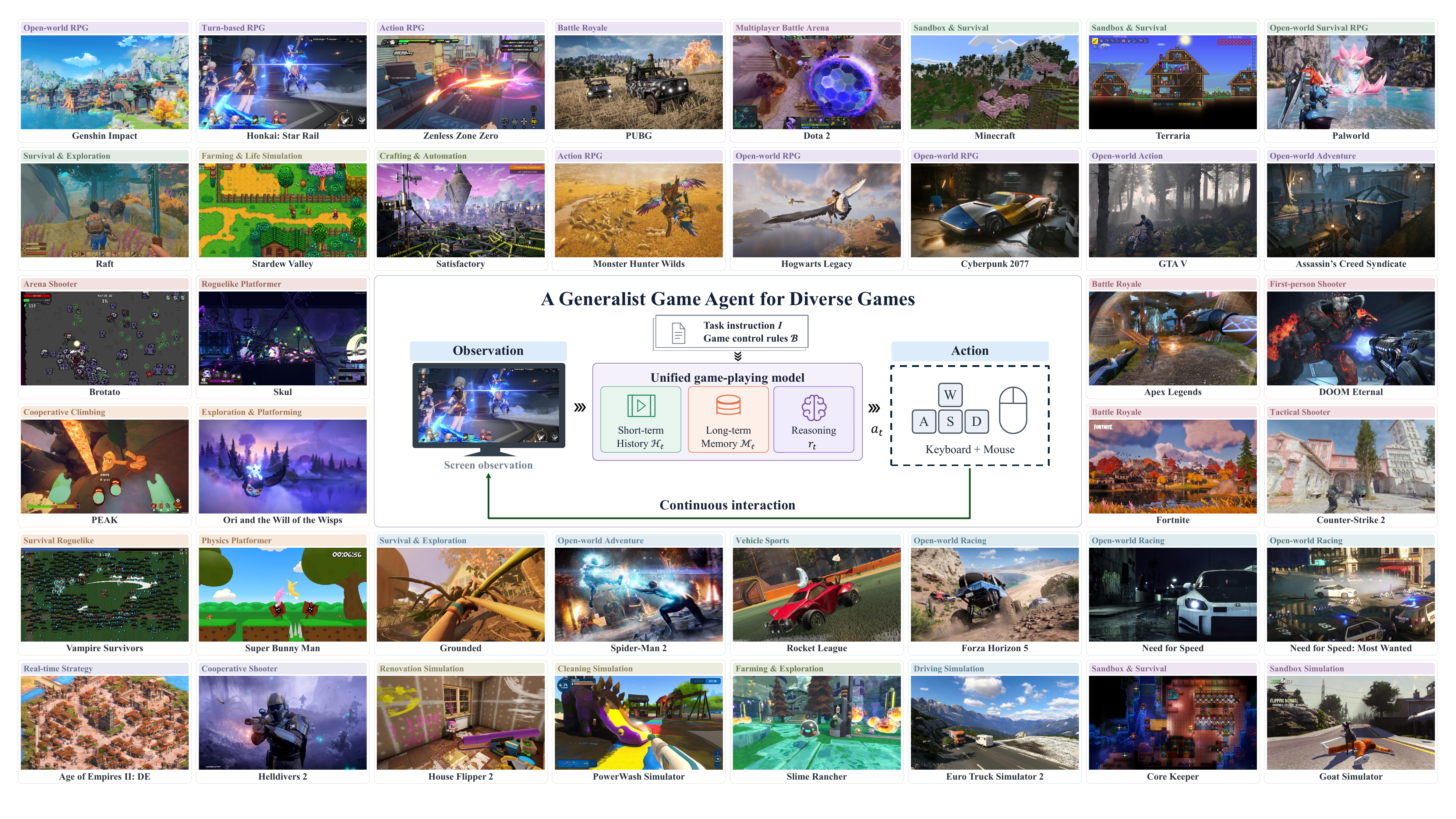}
\caption{
\textbf{Overview of \ourmodel.}
A generalist game agent framework operates across a broad range of visually and mechanically diverse games. At each step, the agent uses the current screen, task context, recent interaction history, long-term memory, and reasoning to generate native keyboard and mouse actions in real time. The agent interacts continuously through this loop. Cross-game training, multi-level supervision, and persistent memory support transferable control and sustained task execution.
}
\vspace{-10pt}
\label{fig:overview}
\end{figure*}

\textit{Honkai: Star Rail} (HSR) exemplifies this setting by condensing these demands into a continuously evolving game environment. Its gameplay combines narrative-driven exploration across diverse regions with turn-based combat, squad management, resource allocation, and multi-level UI operation. This environment features nested objectives and long-term progression that challenge an agent's cross-task generalization, long-term memory, and closed-loop execution. Developing agents in this system compels MLLMs to transition from passive visual understanding to sustained, proactive interaction. Mastering such challenges paves a concrete pathway toward generalist agents capable of operating across both complex digital worlds and embodied physical environments. As illustrated in Figure~\ref{fig:overview}, \ourmodel is designed to operate across visually and mechanically diverse environments via a single screen-to-action interface.

\begin{table*}[t]
\centering
\caption{Comparison between \ourmodel and other representative game agents in terms of model initialization, training recipe, training data, real-time inference, and open-source availability. Here, BC, SFT, CPT, and RL denote behavior cloning, supervised fine-tuning, continued pre-training, and reinforcement learning, respectively.}
\label{tab:comparison}
\small
\setlength{\tabcolsep}{8pt}
\renewcommand{\arraystretch}{1.15}
\resizebox{\textwidth}{!}{
\begin{tabular}{@{}lllccc@{}}
\toprule
Method & Initialization & Training Recipe & Training Data & Real-Time & Open-Source \\
\midrule
VPT       & From scratch & BC + RL         & Minecraft      & \ding{51} & \ding{51} \\
Open-P2P  & From scratch & BC              & Diverse        & \ding{51} & \ding{51} \\
NitroGen  & From scratch & BC              & Diverse        & \ding{55} & \ding{51} \\
Game-TARS & Seed-VL-1.5  & CPT + SFT       & Diverse        & \ding{55} & \ding{55} \\
Lumine    & Qwen2-VL     & CPT + SFT       & Genshin Impact & \ding{51} & \ding{55} \\
SIMA 2    & Gemini 2.5 Pro + Flash-Lite & SFT + RL & Diverse & \ding{55} & \ding{55} \\
\midrule
\textbf{\ourmodel (Ours)} & Qwen3.5/Qwen3-VL & CPT + SFT + RL & Diverse & \ding{51} & \ding{51}\textsuperscript{*} \\
\bottomrule
\end{tabular}
}
\vspace{4pt}
\begin{minipage}{\textwidth}
\footnotesize
\raggedright
\textsuperscript{*}For \ourmodel, we release our inference engine
and the checkpoints used in the CPT scaling study.
For data security reasons, we do not release Honkai: Star Rail
data or any model checkpoints trained or fine-tuned on such data.
\end{minipage}
\end{table*}

Although games have long served as standard testbeds for AI, extending these successes to modern, complex games remains challenging. With domain-specific policy optimization, prior agents have matched or exceeded human performance in benchmarks such as Atari~\cite{atari}, Minecraft~\cite{fan2022minedojo}, and StarCraft~\cite{starcraft}. However, these methods typically use privileged internal states, tailored action spaces, or dedicated APIs, which generalize poorly beyond their narrow training domains. As summarized in Table~\ref{tab:comparison}, existing game agents differ substantially in their training recipes, game coverage, and real-time execution capabilities. Because modern games rarely expose internal states or dedicated interaction APIs, a true generalist agent should operate like a human player~\cite{tan2025lumine,wang2025gametars, player1}: perceiving the world through raw visual pixels and acting through standard keyboard-and-mouse interfaces. Unifying transferable control, long-horizon decision-making, and non-invasive execution within a single generalist model remains largely unaddressed.

Unlike static benchmarks, dynamic game environments evolve continuously while an agent observes, reasons, and acts. Therefore, real-time interaction introduces a critical system bottleneck. Conventional inference engines introduce high inference latency, causing actions to be executed after the observed game state has changed. Over long trajectories, these delays can lead to control errors, unnecessary interactions, and early failures. This latency penalty also affects online reinforcement learning (RL): delayed rollouts not only inflate wall-clock collection times but also dilute the quality of policy feedback. Consequently, real-time inference is not merely a deployment requirement, but an important part of effective online learning.

To address these challenges, we present \ourmodel, a scalable generalist agent framework for complex interactive games. \ourmodel observes raw screen pixels and executes native keyboard and mouse controls. This unified device-level interface enables non-invasive interaction across diverse environments, eliminating reliance on game-specific APIs or customized action spaces. We first conduct continued pre-training (CPT) on thousands of hours of multi-source gameplay trajectories and map heterogeneous control formats into a unified keyboard-and-mouse representation to establish a transferable control prior. A scalable data-construction pipeline is built to transform raw gameplay trajectories into multi-level instruction data. Using LLM-assisted annotation, the pipeline adds task instructions, reasoning at critical decisions, and memory summaries while preserving the original observation-action sequences. We produce over a thousand hours of instruction data covering local operations, intermediate subgoals, and complete tasks. We leverage this corpus for sparse-thinking instruction tuning, enabling the policy to organize executable behavior around long-horizon objectives while retaining task-relevant memory. Finally, we introduce Temporal-Aware Group Relative Policy Optimization (TA-GRPO) to further optimize the game policy through online environment feedback and address accumulated execution errors and sparse rewards over extended interactions.

We support large-scale training and real-time deployment through a unified training, evaluation, and inference infrastructure. We extend Verl~\cite{sheng2024hybridflow} to support CPT, SFT, and online RL within a shared training framework, with self-developed Streaming-vLLM serving as the inference engine for interactive evaluation and rollout generation. Streaming-vLLM combines stateful KV-cache reuse, architecture-aware execution, FSM-guided speculative decoding with BF16 verification, and streaming action dispatch to reduce inference latency. Equipped with a sandbox supporting concurrent game instances and a data pipeline that collects trajectories and identifies failures, this system enables iterative policy improvement under real-time interaction constraints. \ourmodel demonstrates solid generalization and operational efficiency comparable to skilled human players on several challenging 3D game tasks and achieves competitive performance on a range of traditional 2D game benchmarks in real-time. These results demonstrate that a unified keyboard-and-mouse action space which is coupled with cross-domain trajectory scaling and real-time online learning can offer a viable pathway toward generalist interactive agents.

We make the following contributions:
\begin{itemize}
    \item \textbf{A generalist agent framework for complex games.}
    \ourmodel interacts with complex game environments non-invasively, using screen images as input and standard keyboard and mouse operations as a unified action space. We demonstrate how learning in this shared action space generalizes across diverse game.

    \item \textbf{A scalable pipeline for long-horizon instruction data construction.}
    We develop a pipeline that transforms multi-source and multi-format gameplay trajectories into instruction data. Through LLM-assisted annotation and quality validation, the pipeline enriches recorded interactions with task instructions, reasoning at critical decisions, and persistent memory summaries. Applied to tens of thousands of hours of gameplay data, it produces over a thousand hours of high-quality instruction data spanning operations, subgoals, and complete quests, providing supervision for long-horizon reasoning and control.

    \item \textbf{Stable and scalable training, evaluation, and real-time inference.}
    We extend Verl into a unified training and evaluation framework spanning CPT, SFT, and TA-GRPO and introduce Streaming-vLLM to support efficient online rollouts and deployment. Our RL approach optimizes memory-linked trajectory segments using verifiable environment feedback at scale.

    \item \textbf{Strong empirical results.}
    \ourmodel demonstrates solid generalization and operational efficiency comparable to skilled human players on challenging 3D commercial game tasks, and achieves competitive performance on several traditional 2D game benchmarks. These results support its potential for industrial-scale deployment.
\end{itemize}
\section{Continued Pre-Training for Cross-Game Control}
\label{sec:cpt}

\begin{figure*}[!htbp]
    \centering
    \includegraphics[width=0.88\linewidth]{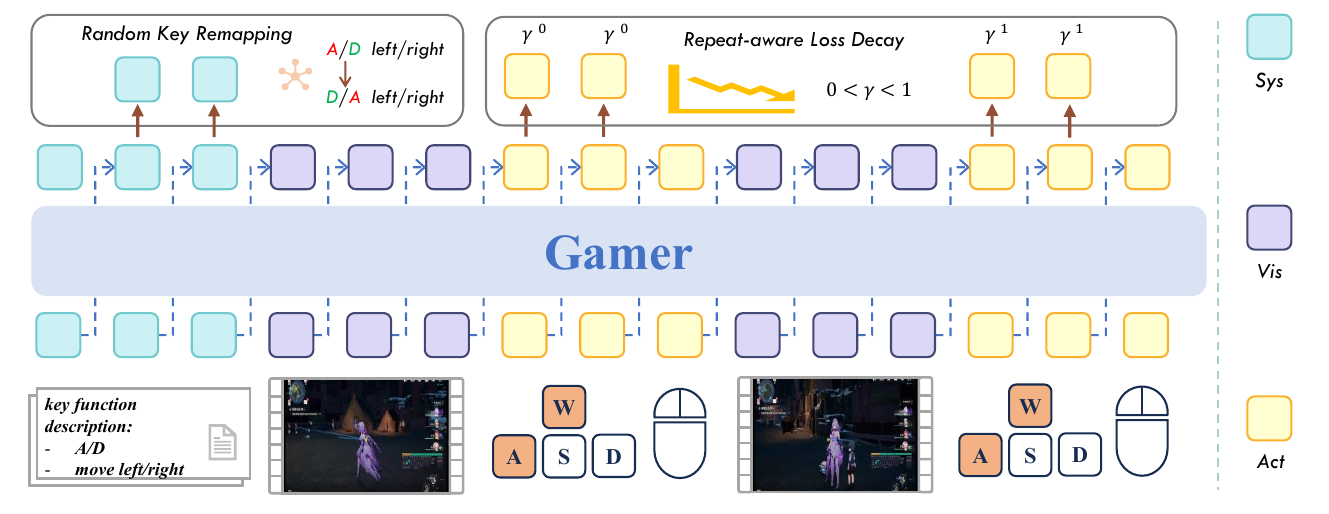}
    \caption{
    \textbf{CPT for unified keyboard-mouse control.}
    Synchronized gameplay observations and device events are mapped to a shared action representation. Random key remapping reduces reliance on key identity, while repeat-aware loss decay reweights temporally persistent actions. The resulting control prior is used by the subsequent task-conditioned training stages.
    }
    \label{fig:cpt_overview}
\end{figure*}

Games differ in world, mechanics, and objectives, but most games share a unified device-level interface: screen observations and native keyboard and mouse inputs. As shown in Figure~\ref{fig:cpt_overview}, CPT adapts a pretrained multimodal model to this interface and learns transferable control from trajectories collected across environments. We first unify action representations across games and reduce biases caused by key frequencies and repeated control states. We then summarize findings from GameScaling~\cite{gamescaling} on how model capacity, gameplay data, and game composition affect control learning and cross-game generalization.

\subsection{Unified Observation-Action Representation}
\label{sec:unified_action}

Training across games requires a common action format that preserves the meaning of each game's controls. We represent each gameplay trajectory as interleaved screen observations and native keyboard-mouse actions. For each game, a control specification $\mathcal{B}$ maps physical inputs to their logical functions:
\[
\mathcal{B}=\{(k_i,d_i)\}_{i=1}^{|\mathcal{B}|},
\]
where $k_i$ is a keyboard key or mouse button and $d_i$ describes its in-game function. We include this specification in the policy input so that the model can interpret the shared input vocabulary using the target game's controls.

Following previous works~\cite{tan2025lumine,wang2025gametars}, we decouple visual decision making from low-level action execution. Observations are sampled every $200$\,ms ($5$\,Hz), and each observation interval is divided into six keyboard-mouse action blocks. Let $\mathcal{E}_t$ denote as the set of mouse movement events in interval $t$, where each event is represented by $\{ (\delta x_i,\delta y_i,\delta z_i)\}$, corresponding to horizontal movement, vertical movement, and scrolling, respectively. We define the action for interval $t$ as
\begin{equation}
a_t=\left(\Delta\mathbf{P}_t;\mathcal{K}_{t,1},\ldots,\mathcal{K}_{t,6}\right),
\qquad
\Delta\mathbf{P}_t=\sum_{\mathcal{E}_t}(\delta x_i,\delta y_i,\delta z_i),
\label{eq:action_representation}
\end{equation}
where $\Delta\mathbf{P}_t$ is the accumulated mouse movement in the interval $t$, and $\mathcal{K}_{t,j}$ ($j=1,\dots,6$) is the set of keyboard keys and mouse buttons pressed in action block $j$. We show more details of the action format in Figure~\ref{fig:action_format}. Each block lasts approximately $33$\,ms, yielding a $30$\,Hz action rate. Keys and buttons shared by consecutive blocks remain pressed, while those absent from the next block are released. This representation supports held movement, brief taps, dragging, and simultaneous input combinations without increasing the visual observation rate. Actions are serialized using the pretrained language model's output vocabulary, without an additional output head.

\begin{figure}[t]
    \centering
    \begin{tcolorbox}[
        enhanced,
        width=0.96\linewidth,
        colback=gray!6,
        colframe=black!65,
        boxrule=0.5pt,
        arc=1.2mm,
        left=3mm,right=3mm,top=4mm,bottom=3mm,
        title={Unified Keyboard-Mouse Action Serialization},
        fonttitle=\bfseries\normalsize,
        coltitle=white,colbacktitle=black!85
    ]
    \centering
    \normalsize\ttfamily
    \detokenize{<|action_start|>}
    $\Delta \mathbf{P}_t$
    \detokenize{;}
    $\mathcal{K}_1$
    \detokenize{;}
    $\mathcal{K}_2$
    \detokenize{;}
    $\mathcal{K}_3$
    \detokenize{;}
    $\mathcal{K}_4$
    \detokenize{;}
    $\mathcal{K}_5$
    \detokenize{;}
    $\mathcal{K}_6$
    \detokenize{<|action_end|>}
    \end{tcolorbox}
    \caption{\textbf{Serialized device-level action.} For each observation interval, the policy predicts accumulated mouse movement and six sequential sets of active keys and buttons.}
    \label{fig:action_format}
\end{figure}

Given $\mathcal{B}$ and a recent interaction history $\mathcal{H}_t$ containing the current observation and preceding observation-action pairs, the policy autoregressively generates an output sequence $y_t=[r_t, a_t]$:
\begin{equation}
\pi_\theta(y_t\mid\mathcal{B},\mathcal{H}_t)
=\prod_{j=1}^{|y_t|}
\pi_\theta(y_{t,j}\mid\mathcal{B},\mathcal{H}_t,y_{t,<j}),
\label{eq:autoregressive_policy}
\end{equation}
where $r_t$ is an optional reasoning sequence preceding the executable action $a_t$, and  $y_{t,j}$ is the $j$-th output token. For a small fraction of CPT turns, $r_t$ contains retrospective reasoning annotated from the observation and recorded action. These annotations preserve the backbone's optional reason-then-act format without changing the action targets. The same executable action representation is retained during SFT, online RL, and deployment.

\subsection{Trajectory Debiasing and Temporal Reweighting}
\label{sec:trajectory_debiasing}

Gameplay data exhibit two biases that can limit transferable control learning. Uneven key frequencies encourage the model to associate actions with specific inputs rather than their functions, while long runs of repeated actions can dominate the token-level supervision. We address these biases with random key remapping and repeat-aware loss decay, respectively.

\paragraph{Random key remapping.}
For each trajectory, we sample a random bijection $\sigma:\mathcal{K}_{\mathrm{bind}}\rightarrow\mathcal{K}_{\mathrm{bind}}$ over rebindable keys. Inputs with fixed semantics remain unchanged. The same mapping is applied to the control specification and the actions:
\begin{equation}
\begin{aligned}
\mathcal{B}^\sigma
  =\{(\sigma(k_i),d_i)\}_{i=1}^{|\mathcal{B}|},\quad
\mathcal{K}_{t,j}^\sigma
=\{\sigma(k)\mid k\in\mathcal{K}_{t,j}\},\quad
a_t^\sigma
    =\left(\Delta\mathbf{P}_t; \mathcal{K}_{t,1}^\sigma,\ldots,\mathcal{K}_{t,6}^\sigma\right).
\end{aligned}
\end{equation}
Historical actions are remapped consistently to obtain $\mathcal{H}_t^\sigma$, whereas visual observations and mouse displacements remain unchanged. The policy is trained to predict $a_t^\sigma \sim \pi_\theta(a_t^\sigma\mid\mathcal{B}^\sigma,\mathcal{H}_t^\sigma)$. The mapping remains fixed within a trajectory, preserving its control semantics while reducing shortcuts based on frequent key identities.

\paragraph{Repeat-aware loss decay.}
More than $85\%$ of the action steps in our interaction data repeat the preceding action, which is common in gameplay since players often hold the same inputs across multiple time steps to perform sustained behaviors such as moving or waiting. These repetitions record how long such behaviors last, so we retain the full trajectories while reducing the contribution of prolonged repetitions. Following the decaying-loss formulation in Game-TARS~\cite{wang2025gametars}, we down-weight consecutive repeated actions. Specifically, the action-token component of the CPT objective we used is
\begin{equation}
\begin{aligned}
\mathcal{L}_{\mathrm{action}}
    &=-\sum_{t=1}^{T}\omega_t
      \sum_{j=1}^{|a_t|}
      \log\pi_\theta(a_{t,j}\mid\mathcal{B},\mathcal{H}_t,r_t,a_{t,<j}),\quad
\omega_t=\gamma^{n_t-1},
\end{aligned}
\label{eq:cpt_objective_1}
\end{equation}
where $n_t$ counts consecutive repetitions of the current action and resets to $1$ whenever the action changes. We use $\gamma=0.9$ and omit the remapping superscript for clarity. When present, retrospective reasoning tokens are additionally supervised as part of the same autoregressive response. Eq.~\eqref{eq:cpt_objective_1} specifies the reweighting applied to action tokens, not the full mixed-token objective.

A newly issued action receives full weight, while successive repetitions contribute progressively less. This preserves action duration while emphasizing behavior transitions, such as stopping, turning, and interacting. Random key remapping discourages reliance on key identity, while repeat-aware decay shifts supervision toward changes in behavior.

The CPT objective combines the reweighted action-token loss with a standard autoregressive loss for the optional reasoning tokens:
\begin{equation}
\begin{aligned}
\mathcal{L}_{\mathrm{CPT}}
&=\mathcal{L}_{\mathrm{action}}+\mathcal{L}_{\mathrm{reason}},\\
\mathcal{L}_{\mathrm{reason}}
&=-\sum_{t=1}^{T}
\sum_{j=1}^{|r_t|}
\log\pi_\theta(r_{t,j}\mid\mathcal{B},\mathcal{H}_t,r_{t,<j}).
\end{aligned}
\label{eq:cpt_objective}
\end{equation}
Here, $r_t=\varnothing$ for turns without reasoning, so the corresponding reasoning loss is zero. The repeat-aware weight $\omega_t$ is applied only to action tokens. For clarity, we omit the remapping superscript; when key remapping is applied, the control specification, interaction history, and action targets are remapped consistently.

\subsection{Scaling Laws and Cross-Game Generalization}
\label{sec:cpt_scaling}

Although a shared control interface makes cross-game CPT possible, the benefits of additional gameplay data depend on how well the training data cover the target environment. GameScaling \cite{gamescaling} examines this dependence by varying model size, gameplay budget, and game composition. We summarize its findings here to contextualize Gamer’s CPT stage.

To characterize these effects, a composition-aware scaling relation based on D-CPT Law~\cite{que2024dcpt} is adopted. Let $N$ be model size, $D$ the gameplay budget in hours, and $q\in[0,1]$ the proportion of gameplay relevant to a specified validation distribution. The validation loss is modeled as
\begin{equation}
\mathcal{L}_{\mathrm{val}}(N,D,q)
=E+\frac{C_N}{N^a}
+\frac{C_D q^\eta}{D^b}
+\frac{C_q}{(q+\epsilon)^c},
\label{eq:cpt_scaling_law}
\end{equation}
where $E$, $C_N$, $C_D$, $C_q$, $a$, $b$, $c$, and $\eta$ are fitted variables. The small constant $\epsilon>0$ keeps the expression finite when the target game's share is zero. In view of the widespread use of Minecraft in prior work~\cite{wang2025gametars,fan2022minedojo}, Gamescaling \cite{gamescaling} varies its proportion in the gameplay mixture to study the effect of data composition. We use $q=w_{\mathrm{MC}}$ for Minecraft validation and $q=1-w_{\mathrm{MC}}$ for validation over the remaining training-game distribution, where $w_{\mathrm{MC}}$ is the Minecraft fraction of the gameplay mixture.

The data term describes how validation loss changes with the gameplay budget, and the final term captures differences between data mixtures. The scaling relation fits the training-game distributions and supports held-out prediction within the measured range. However, distant unseen games do not exhibit sustained gains from additional CPT data. Sec.~\ref{sec:exp_cpt_scaling} provides the controlled study and its empirical limits. CPT thus provides transferable control, while more distant worlds and longer tasks motivate additional task-conditioned supervision and online interaction.
\section{Supervised Fine-Tuning for Sparse Reasoning and Cross-Context Memory}
\label{sec:sft}

\begin{figure}[H]
    \centering
    \includegraphics[width=0.94\linewidth]{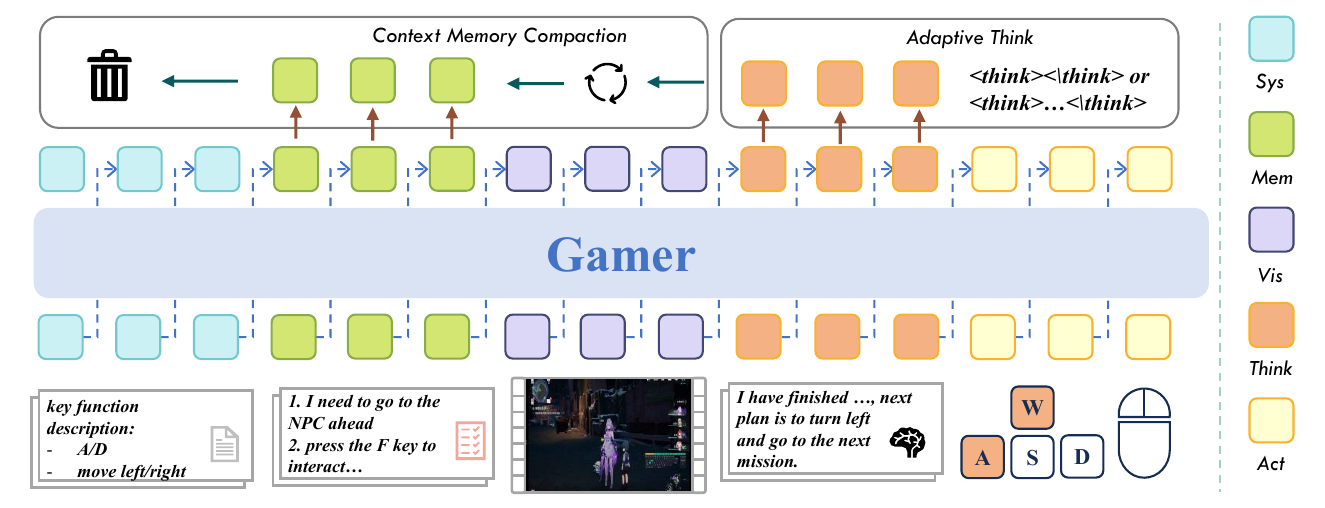}
    \caption{
    \textbf{Sparse reasoning and cross-context memory in supervised fine-tuning.}
    The shared autoregressive policy integrates visual observations, task memory, optional reasoning, and executable keyboard-mouse actions. Context memory compaction preserves task progress across long interactions, while adaptive reasoning allows the policy to act directly at routine steps and generate explicit reasoning when needed.
    }
    \label{fig:sft_overview}
\end{figure}

CPT learns transferable action control from visual observations, while supervised fine-tuning (SFT) teaches the model to follow explicit task objectives. We train on tasks at different difficulty levels, from local operations and intermediate subgoals to complete objectives, with reasoning and memory summaries at key decision points. The need for reasoning is inherently sparse in long-horizon gameplay. Routine movement and repeated attacks favor direct action, while route failures, target switches, and resource decisions often require replanning. \ourmodel therefore uses a single autoregressive policy to alternate between direct control, sparse reasoning, and memory updates. Persistent summaries carry task state across context boundaries. Figure~\ref{fig:sft_overview} illustrates how adaptive reasoning and memory compaction are integrated into an interaction sequence.

\subsection{Multi-level Task Conditioning and Sparse Reasoning}
\label{sec:sparse_reasoning}

SFT retains the control specification $\mathcal{B}$ and recent interaction history $\mathcal{H}_t$ introduced during CPT, and additionally conditions the policy on a task instruction $I$ and long-term memory $\mathcal{M}_t$. The resulting context is
\[
(\mathcal{B}, I, \mathcal{M}_t, \mathcal{H}_t),
\]
where $\mathcal{H}_t$ contains the current observation together with recent observations, actions, and previously generated reasoning. To expose the policy to task structure at different temporal scales, the instruction $I$ is instantiated at three difficulty levels:
\[
I
\in
\left\{
I^{\mathrm{op}},
I^{\mathrm{sub}},
I^{\mathrm{task}}
\right\},
\]
in which $I^{\mathrm{op}}$, $I^{\mathrm{sub}}$, and $I^{\mathrm{task}}$ correspond to local operations, intermediate subgoals, and complete tasks, respectively. Training data at all three levels use the same observation–action interface and are used to train a single policy that produces executable keyboard and mouse actions.

At each interaction step, the model first generates an optional reasoning sequence $r_t$ and then produces the executable action $a_t$. Their joint distribution follows the autoregressive generation order:
\begin{equation}
\pi_\theta
\left(
r_t,a_t
\mid
\mathcal{B},I,\mathcal{M}_t,\mathcal{H}_t
\right)
=
\pi_\theta
\left(
r_t
\mid
\mathcal{B},I,\mathcal{M}_t,\mathcal{H}_t
\right)\cdot
\pi_\theta
\left(
a_t
\mid
\mathcal{B},I,\mathcal{M}_t,\mathcal{H}_t,r_t
\right).
\label{eq:sft_joint_policy}
\end{equation}

The reasoning sequence takes one of three forms:
\begin{equation}
r_t
\in
\left\{
\varnothing,
r_t^{\mathrm{cur}},
r_t^{\mathrm{mem}}
\right\},
\label{eq:reasoning_types}
\end{equation}
when $r_t = \emptyset$, $\pi_\theta(a_t \mid \mathcal{B}, I, \mathcal{M}_t, \mathcal{H}_t, r_t)$ naturally simplifies to $\pi_\theta(a_t \mid \mathcal{B}, I, \mathcal{M}_t, \mathcal{H}_t)$.
$r_t^{\mathrm{cur}}$ denotes reasoning for the current decision and $r_t^{\mathrm{mem}}$ denotes a memory summary for future contexts. At routine control steps, the model sets $r_t=\varnothing$ and generates an action directly. At decision-critical or memory-update steps, the model generates reasoning before the action. Current reasoning captures immediate judgment, next subgoal, or short-term plan, such as revising a route, selecting an interaction target, or switching combat strategy. Memory summaries, by contrast, retain information needed beyond the current context, including task progress, important locations, resource states, failed attempts, and remaining plans. The policy learns when to generate reasoning within the same autoregressive process, without a separate classifier or routing head. Here, $r_t$ refers only to the explicitly supervised reasoning text generated by the policy and does not characterize all latent computation within the model.

SFT trains current reasoning, memory summaries, and executable actions within the same autoregressive output space. Let
\[
y_t=[r_t,a_t]
\]
denote the complete output sequence at step $t$, and let $y_{t,<j}$ denote the generated prefix before its $j$-th token. The SFT objective is
\begin{equation}
\mathcal{L}_{\mathrm{SFT}}
=
-
\sum_t
\sum_{j=1}^{|y_t|}
\log
\pi_\theta
\left(
y_{t,j}
\mid
\mathcal{B},
I,
\mathcal{M}_t,
\mathcal{H}_t,
y_{t,<j}
\right).
\label{eq:sft_objective}
\end{equation}

The objective uses the same token-level autoregressive formulation as CPT, with the target extended from the action $a_t$ to the reasoning-action sequence $[r_t,a_t]$. It also introduces the task instruction $I$ and long-term memory $M_t$ into the conditioning context.

\subsection{Summary-Based Cross-Context Memory}
\label{sec:cross_context_memory}

Long-horizon tasks require the agent to retain relevant information after earlier observations and actions leave the active context. We update long-term memory $\mathcal{M}_t$ when the task changes stage, a subgoal is completed, or older interactions are about to leave the active context. Consider a memory-update step in which the short-term history $\mathcal{H}_t$ is partitioned as
\[
\mathcal{H}_t
=
\mathcal{H}_t^{\mathrm{old}}
\oplus
\mathcal{H}_t^{\mathrm{keep}},
\]
where $\mathcal{H}_t^{\mathrm{old}}$ contains earlier interactions that will leave the active context, $\mathcal{H}_t^{\mathrm{keep}}$ contains recent interactions that remain directly accessible, and $\oplus$ denotes temporal concatenation.

Before $\mathcal{H}_t^{\mathrm{old}}$ is discarded, the model generates a memory summary from the complete information currently available. The long-term memory and short-term history are then updated as
\begin{equation}
r_t^{\mathrm{mem}}
\sim
\pi_\theta
\left(
\cdot
\mid
\mathcal{B},I,\mathcal{M}_t,\mathcal{H}_t
\right),
\qquad
\left(
\mathcal{M}_t^{+},
\mathcal{H}_t^{+}
\right)
=
\left(
\bar {\mathcal{M}_t}
\oplus
r_t^{\mathrm{mem}},
\mathcal{H}_t^{\mathrm{keep}}
\right).
\label{eq:memory_update}
\end{equation}

Here, $\mathcal{M}_t^{+}$ and $\mathcal{H}_t^{+}$ denote the updated long-term memory and short-term history. $\bar {\mathcal{M}_t}$ denotes the portion of the previous memory retained after reserving space for the new summary. When the memory budget is exceeded, the oldest complete summaries are removed in temporal order.

The update preserves overall task progress in memory and recent interaction details in the active history. Task-relevant information contained in $\mathcal{H}_t^{\mathrm{old}}$ is carried forward through $r_t^{\mathrm{mem}}$, while fine-grained recent evidence remains directly accessible in $\mathcal{H}_t^{\mathrm{keep}}$. When memory is updated at a task boundary without context eviction, we set $\mathcal{H}_t^{\mathrm{old}}=\varnothing$ and $\mathcal{H}_t^{\mathrm{keep}}=\mathcal{H}_t$.

A memory-update step remains an ordinary interaction step. After generating $r_t^{\mathrm{mem}}$, the policy still produces the corresponding executable action $a_t$. Memory generation remains part of the reasoning–action process, allowing closed-loop control to continue without help from another memory model. During training, each step uses only the information that would be available at inference time. Each memory summary is generated only from interactions available at the corresponding update point. Subsequent trajectory segments are conditioned on the updated memory together with the retained recent history. The policy thus learns to continue a task after parts of its interaction history are no longer directly accessible.

\section{Temporal-Aware GRPO for Long-Horizon Policy Optimization}
\label{sec:ta_grpo}

CPT and SFT provide cross-game control and task-following capabilities, while online RL improves the policy on states encountered during its own interactions. Online RL in real-time games faces two challenges: long interactions can exceed the model's context window, and delayed model responses can change the environment transitions observed during rollout collection. We address the first using Temporal-Aware GRPO (TA-GRPO) and the second using the shared Streaming-vLLM infrastructure (Sec.~\ref{sec:streaming_vllm}). Together, they enable policy optimization over long sequences of real-time keyboard and mouse interaction.

\begin{figure*}[!t]
    \centering
    \includegraphics[width=0.89\linewidth]{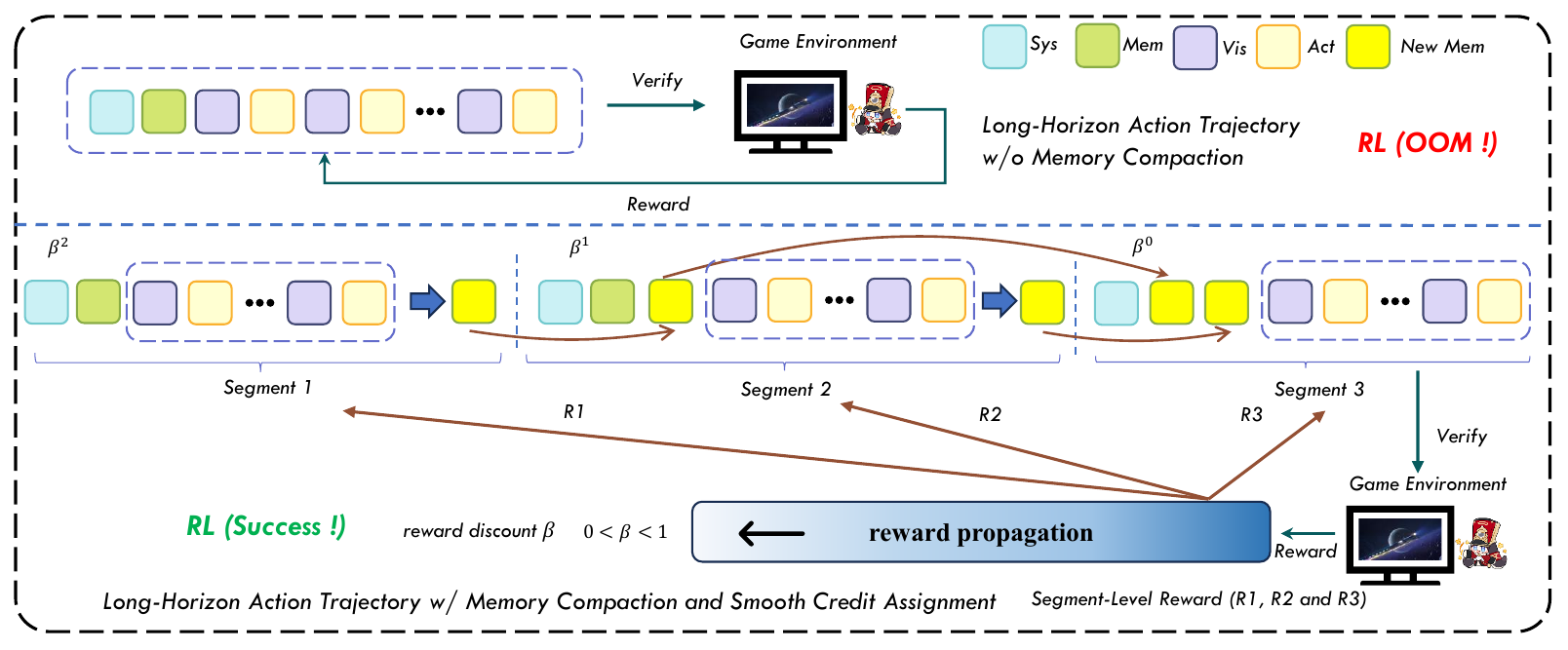}
    \caption{\textbf{Memory-linked long-horizon optimization with TA-GRPO}. A long interaction may exceed the model's context budget (top). Memory compaction divides online rollouts into finite segments whose task states remain connected by generated summaries (bottom). Segments at corresponding stages are compared using environment-verifiable outcomes, and temporal credit assignment distributes their group-relative advantage across decisions. The figure is schematic, and model execution uses the real-time serving backend in Sec.~\ref{sec:streaming_vllm}.}
    \label{fig:ta_grpo}
\end{figure*}

As illustrated in Figure~\ref{fig:ta_grpo}, TA-GRPO divides long rollouts into finite segments linked by the same memory compaction mechanism used during SFT. It scores segments from separate rollouts of the same task using verifiable outcomes and assigns group-relative advantages~\cite{shao2024deepseekmath} to the decisions within each segment. The policy can therefore learn from long interactions without fitting a full trajectory into one context or repeatedly restoring an intermediate game state. The policy retains the same input and output format, generating reasoning, memory updates, and executable actions as needed.

\subsection{Memory-Linked Rollouts}
\label{sec:ta_grpo_segments}

For a task instruction $I$, the old policy $\pi_{\theta_{\mathrm{old}}}$ samples $G$ long-horizon rollouts from the same task condition. Memory compaction partitions rollout $g$ into $J_g$ consecutive segments $\tau_{g,1},\ldots,\tau_{g,J_g}$, each containing $T_{g,m}$ interaction steps. At step $t$ of segment $m$, the policy receives $x_{g,m,t}=(\mathcal B,I,\mathcal M_{g,m,t},\mathcal H_{g,m,t})$ and generates $y_{g,m,t}=(r_{g,m,t},a_{g,m,t})$. The memory produced at a segment boundary carries completed objectives, failures, and remaining plans into subsequent segments. Earlier detailed observations may leave the active context without resetting the task.

For each segment index $m$, we form a comparison group from rollouts that reach that stage, with index set $\mathcal V_m$ and size $G_m=|\mathcal V_m|$. Unlike state-level branching, these segments are not required to start from identical intermediate game states. Grouping rollouts by task condition and compaction stage avoids costly game-state restoration. The shared action format keeps output structure and interaction granularity consistent. This grouping makes computation more regular, although the segments may still differ in difficulty and state distribution. Environment-verifiable feedback determines the relative outcomes used for learning.

\subsection{Verifiable Rewards and Temporal Credit Assignment}
\label{sec:ta_grpo_reward}
\label{sec:temporal_credit}

Long-horizon tasks often expose sparse terminal feedback, although some intermediate objectives admit direct verification. We use the verified outcome $R_{g,m}^{\mathrm{env}}$ of the segment's current objective as its score $S_{g,m}=R_{g,m}^{\mathrm{env}}$. When milestone completion is verifiable, it can provide feedback before the entire quest ends. Otherwise, the score remains sparse. No learned process reward model or value critic is required. Environment-side verification may access task state for training and evaluation, while the deployed policy still observes only the screen and issues keyboard-mouse actions.

Within the group at stage $m$, scores are normalized to obtain a relative segment advantage:
\begin{equation}
\label{eq:segment_advantage}
A_{g,m}=\frac{S_{g,m}-\mu_m}{s_m},\qquad
\mu_m=\frac{1}{G_m}\sum_{g\in\mathcal V_m}S_{g,m},\quad
s_m=\sqrt{\frac{1}{G_m}\sum_{g\in\mathcal V_m}(S_{g,m}-\mu_m)^2+\epsilon_{\mathrm{grp}}}.
\end{equation}
This critic-free estimate follows GRPO~\cite{shao2024deepseekmath}. Because verification typically occurs at a segment boundary, the segment score does not identify the contributions of individual decisions. TA-GRPO assigns larger weights to later steps, which are temporally closer to that outcome, while normalizing the weights to preserve their mean:
\begin{equation}
\label{eq:temporal_advantage}
\widetilde d_{g,m,t}=\frac{\beta^{T_{g,m}-1-t}}{\frac{1}{T_{g,m}}\sum_{u=0}^{T_{g,m}-1}\beta^{T_{g,m}-1-u}},\qquad
A_{g,m,t}=\widetilde d_{g,m,t}A_{g,m},\quad \beta\in(0,1).
\end{equation}
The temporal weights average to one, redistributing the segment-level advantage across steps while preserving the mean. This weighting encodes a temporal prior, and it does not imply that later actions contribute more to the outcome.

\subsection{Joint Reasoning-Action Policy Update}
\label{sec:ta_grpo_update}

TA-GRPO updates the autoregressive policy over the complete output $y_{g,m,t}=(r_{g,m,t},a_{g,m,t})$, where $r_{g,m,t}$ may be empty, contain reasoning for the current decision, or provide a memory summary. The low-level action $a_{g,m,t}$ is always generated. For output token $j$, we define the ratio between the current and old policy, conditioned on the preceding output tokens:
\begin{equation}
\label{eq:token_policy_ratio}
\rho_{g,m,t,j}(\theta)=\frac{\pi_\theta(y_{g,m,t,j}\mid x_{g,m,t},y_{g,m,t,<j})}{\pi_{\theta_{\mathrm{old}}}(y_{g,m,t,j}\mid x_{g,m,t},y_{g,m,t,<j})}.
\end{equation}
We use the standard token-level clipped GRPO surrogate, with $\rho=\rho_{g,m,t,j}(\theta)$ and $A=A_{g,m,t}$ for readability:
\[
\ell_{g,m,t,j}(\theta)
=\min\!\left\{\rho A,\,
\operatorname{clip}\!\left(\rho,1-\epsilon_{\mathrm{clip}},1+\epsilon_{\mathrm{clip}}\right)A\right\}.
\]
The final objective averages this surrogate over valid segments, interaction steps, and output tokens:
\begin{equation}
\label{eq:ta_grpo_objective}
\mathcal L_{\mathrm{TA\text{-}GRPO}}=-\frac{1}{\sum_m G_m}\sum_m\sum_{g\in\mathcal V_m}\frac{1}{T_{g,m}}\sum_{t=0}^{T_{g,m}-1}\frac{1}{|y_{g,m,t}|}\sum_{j=1}^{|y_{g,m,t}|}\ell_{g,m,t,j}(\theta).
\end{equation}
Token-level clipping avoids a single sequence-wide probability ratio whose scale would depend strongly on the combined reasoning and action length. The same learning signal can update memory summaries, sparse reasoning, and executable control. Streaming-vLLM generates these rollouts across concurrent game instances and dispatches actions promptly.

\section{Streaming-vLLM for Real-Time and Scalable Gameplay}
\label{sec:streaming_vllm}

Real-time gameplay introduces timing constraints that are absent when evaluation pauses the game during inference. The results in Sec.~\ref{sec:exp_cpt_scaling} show that a policy can perform well in a paused game but degrade sharply when the game state changes before its actions arrive. This temporal mismatch also affects online policy optimization: delayed actions change subsequent states and the verified feedback in collected trajectories. At the same time, long-horizon training requires many independent environments to advance concurrently. Practical gameplay and efficient online rollout collection therefore require low-latency serving that preserves state across turns.

We develop \textbf{Streaming-vLLM} as a common execution backend for training, evaluation, and deployment. The backend retains KV states across turns under memory compaction and reduces model execution cost. It also verifies structural tokens proposed by a fast finite-state machine (FSM) and streams verified action blocks before the full packet is complete. The model-facing interface is shared across Qwen3-VL~\cite{qwen3} and Qwen3.5~\cite{qwen35}, including dense and mixture-of-experts (MoE) backbones. Architecture-specific kernels and schedules support each backbone while preserving the same observation–action protocol. Appendix \ref{app:streaming_details} reports representative Dense and MoE serving configurations on Ampere (NVIDIA A800) and Hopper (NVIDIA H200).

\begin{figure*}[!htbp]
    \centering
    \includegraphics[width=0.89\linewidth]{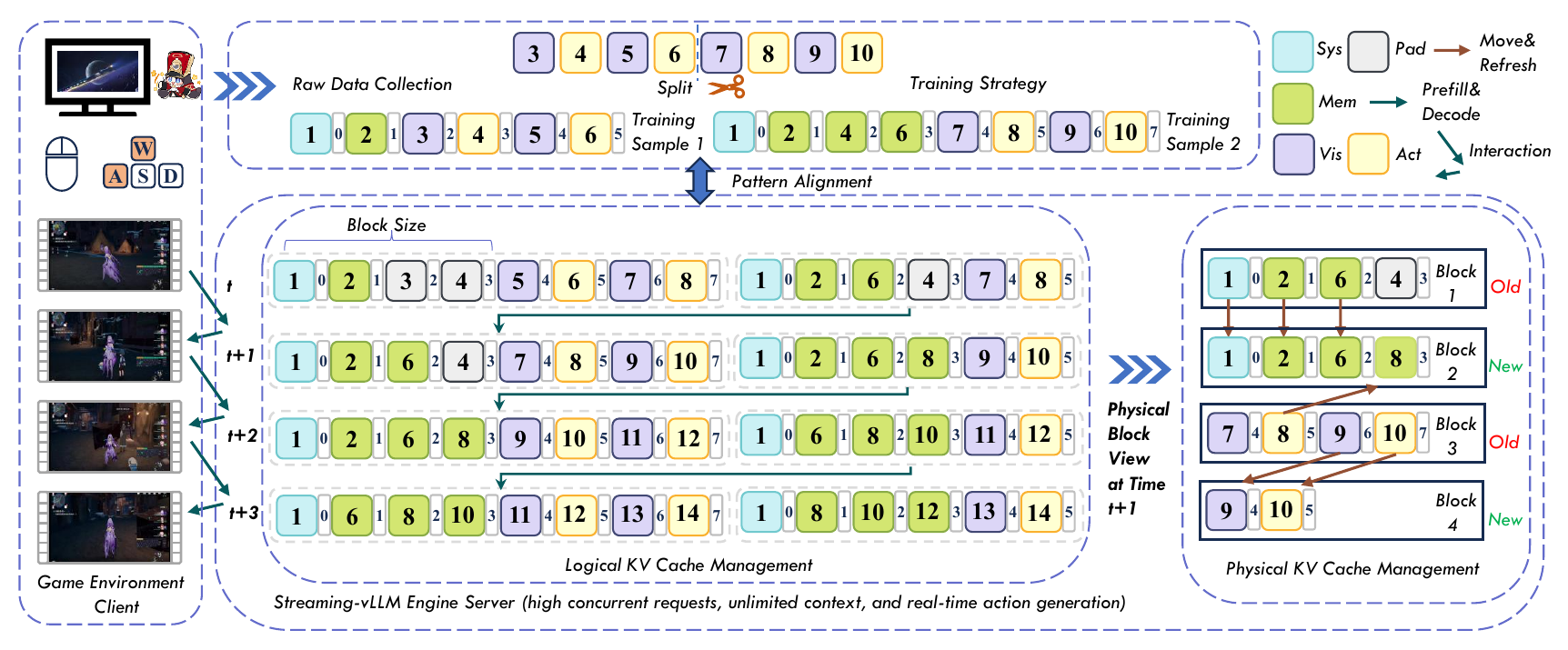}
    \caption{\textbf{Streaming-vLLM for persistent multi-environment inference.} Each game client maintains separate short-term history, long-term memory, and logical KV-cache state. Two cache forms coexist. Full-Attention layers keep a \emph{growing} paged KV cache: blocks are allocated and reused on demand, and released when context is compacted. In hybrid backbones, gated delta network (GDN) layers instead hold \emph{fixed-size} linear recurrent and convolution states (\texttt{ssm\_state}, \texttt{conv\_state}) whose footprint is independent of sequence length and which are carried across compaction in place. Paged allocation and memory-aware KV-cache reuse avoid repeated context processing, while incremental rotary position embedding (RoPE) refresh corrects retained keys after compaction. The runtime combines architecture-aware execution, BF16-verified FSM speculation, and early action dispatch to serve concurrent real-time interactions.}
    \label{fig:streaming_vllm}
\end{figure*}

\subsection{Memory-Compacted Streaming Context}
\label{sec:streaming_context}

At each interaction step, the model consumes the control specification, task instruction, long-term memory, and recent observation-action history $(\mathcal B,I,\mathcal M_t,\mathcal H_t)$. Re-encoding the visual and textual history at every turn would repeat the same prefill work. As shown in Figure~\ref{fig:streaming_vllm}, Streaming-vLLM instead preserves KV states of retained context and computes new states only for newly appended tokens, following stateful streaming inference~\cite{xiao2024streamingllm,xu2026streamingvlm}. A short-term queue holds recent complete interaction rounds, while a long-term queue stores generated memory summaries. At a compaction boundary, the policy summarizes history before eviction. The discarded rounds release their KV pages, and recent rounds retain cached states. Each environment keeps its own logical-to-physical block table over the paged allocator~\cite{kwon2023pagedattention}, allowing requests with different history lengths to coexist.

Compaction also changes the logical token positions of retained states. We update the rotary phase of the corresponding cached keys using the positional shift, without recomputing their projections or modifying the value cache; in hybrid backbones, gated-delta-network (GDN) layers carry no rotary phase and instead retain their fixed-size linear recurrent and convolution states in place across eviction boundaries. The retained cache preserves state across turns, so it need not be numerically equivalent to a full re-encoding of the shortened context. The precise RoPE refresh and its scope are detailed in Appendix~\ref{app:rope_refresh}. The same bounded history and persistent summaries are used in SFT and to link TA-GRPO trajectory segments.

\subsection{Architecture-Aware Execution Across Backbones}
\label{sec:streaming_backend}

Frequent short requests incur costs beyond prefill, including decoder launches, inter-device communication, and architecture-specific operations. Streaming-vLLM separates a common serving interface from specialized execution paths. Dense and MoE variants share action serialization, context compaction, and client scheduling, while the compute backend accommodates their respective attention, expert, and recurrent or gated-delta-network operations. We combine operator fusion, tensor-parallel execution, communication overlap, and CUDA Graph replay to reduce serial overhead. We apply tensor parallelism selectively. It can reduce decoder compute time, but communication and synchronization may outweigh the savings for the smaller ViT workload. The measured tradeoff is analyzed in Table~\ref{tab:streaming_vllm_ablation}. Across Ampere and Hopper GPUs, decode and speculative verification are each implemented as an all-layer megakernel that combines fine-grained L2 prefetching, in-place cross-rank reduction and normalization, and a fused sampling and state-machine tail to minimize CPU and synchronization overhead. For vision encoding and context prefill, persistent kernels overlap tile-level communication with computation and fold cross-rank reduction and normalization into the GEMM epilogue. Finally, W8A8 quantization yields only modest end-to-end gains and is applied selectively: ViT remains in BF16 to avoid low-precision representation instability, prefill quantizes only large QKV and MoE expert projections to prevent router flips and GDN state drift across short single-frame extensions, and low-batch decode is latency- and barrier-bound rather than compute-bound.

We assess portability across model families and hardware configurations. Our supported model set spans Qwen3-VL~\cite{qwen3} dense 2B, 4B, 8B, and 32B and MoE 30B-A3B, together with Qwen3.5~\cite{qwen35} dense 0.8B, 2B, 9B, and 27B and MoE 35B-A3B. Our experiments and analysis cover representative Qwen3-VL Dense and Qwen3.5 MoE workloads across datacenter Ampere and Hopper GPUs. The appendix compares numerical precision and tensor-parallel settings on each GPU. Beyond these datacenter architectures, because the kernel pipelines and tile schedules are parameterized by SM count, register budget, shared-memory ceiling, and L2 capacity at runtime, the same execution paths also extend to Ada Lovelace and Blackwell GPUs. Although these platforms face much tighter inter-GPU communication constraints without NVLink peer-to-peer interconnect, a custom communication layer encapsulates zero-copy host-mapped shared memory, system-scope lock-free release/acquire primitives, and payload-adaptive collective schedules behind an unchanged model-facing interface. These extensions are included for architectural portability and compatibility rather than benchmark comparison.

\subsection{Verified FSM Decoding and Streaming Actions}
\label{sec:streaming_action}

After execution overhead is reduced, autoregressive generation remains sequential. A fast finite-state-machine (FSM) drafter proposes predictable structural tokens in the reasoning–action output. The FSM tracks \texttt{<think>}, \texttt{</think>}, \texttt{<|action\_start|>}, and \texttt{<|action\_end|>} boundaries. It proposes closing an empty reasoning span or starting an action packet after reasoning ends. Within the action, it drafts whitespace after mouse-coordinate numbers, the separator following the Z coordinate, and predictable delimiters and empty-key patterns across the six keyboard-mouse groups. \textbf{Mouse-coordinate values remain model-generated.} During generation, the FSM enforces constrained sampling over valid transitions via state-dependent vocabulary bitmasks at every decoding and verification position, while simultaneously packing candidate continuations from the next grammar states into a draft tree that the target model verifies through a single shared-prefix pass~\cite{leviathan2023fast,zheng2024sglang,miao2024specinfer,cai2024medusa,li2024eagle}, committing only accepted prefixes. The policy retains control over action selection, while verification reduces the number of sequential decoding iterations. See Appendix~\ref{app:speculative_sampling} for more details.

Streaming-vLLM also reduces the time to the first executable action. Once the mouse fields and first verified key block are ready, the server sends that prefix to the client while the remaining blocks continue decoding. The client executes the six blocks in order, maintaining key press-hold-release semantics. Action execution can therefore begin while the remaining blocks are still being generated. The theoretical block-generation rate reported in the system ablation and the nominal 30~Hz client control protocol are distinct measurements, as clarified in Appendix~\ref{app:streaming_latency}.

\subsection{Concurrent Serving for Online Rollouts}
\label{sec:streaming_serving}

Real-time rollouts require scheduling that accounts for both action deadlines and throughput. Each active environment is assigned an isolated resident state slot that holds its short-term history, long-term memory, and model states, avoiding context rebuilds when environments alternate turns. The runtime time-division multiplexes incoming frames' vision encoding and context extension across slots over a shared workspace, and then performs batched token generation and verification across all ready environments. Within each slot, paged KV caches and linear recurrent (GDN) states follow different update rules during scheduling: KV caches are position-indexed, so speculative tokens write directly to trailing slots and rejected suffixes are simply overwritten on the next step, whereas fixed-size recurrent states update in place and are committed only for the accepted prefix after verification. This balances aggregate throughput against the progress rate of each active environment. Real-time evaluation and TA-GRPO training share the same serving path, so training rollouts are collected under the same execution protocol while the game continues to advance.

Stateful cache reuse reduces the cost of long contexts, architecture-specific backends support different models and GPUs, and verified FSM drafting reduces action-generation cost. Combined with early action dispatch, these components enable responsive, high-frequency control and parallel RL rollouts linked by memory summaries. Section~\ref{sec:exp_streaming} reports latency and closed-loop evaluations, and Appendix~\ref{app:streaming_latency} provides representative GPU, precision, and parallelism settings.

\section{Data Construction for Cross-Game Learning}
\label{sec:data_construction}

\begin{figure*}[!htbp]
    \centering
    \includegraphics[width=0.92\linewidth]{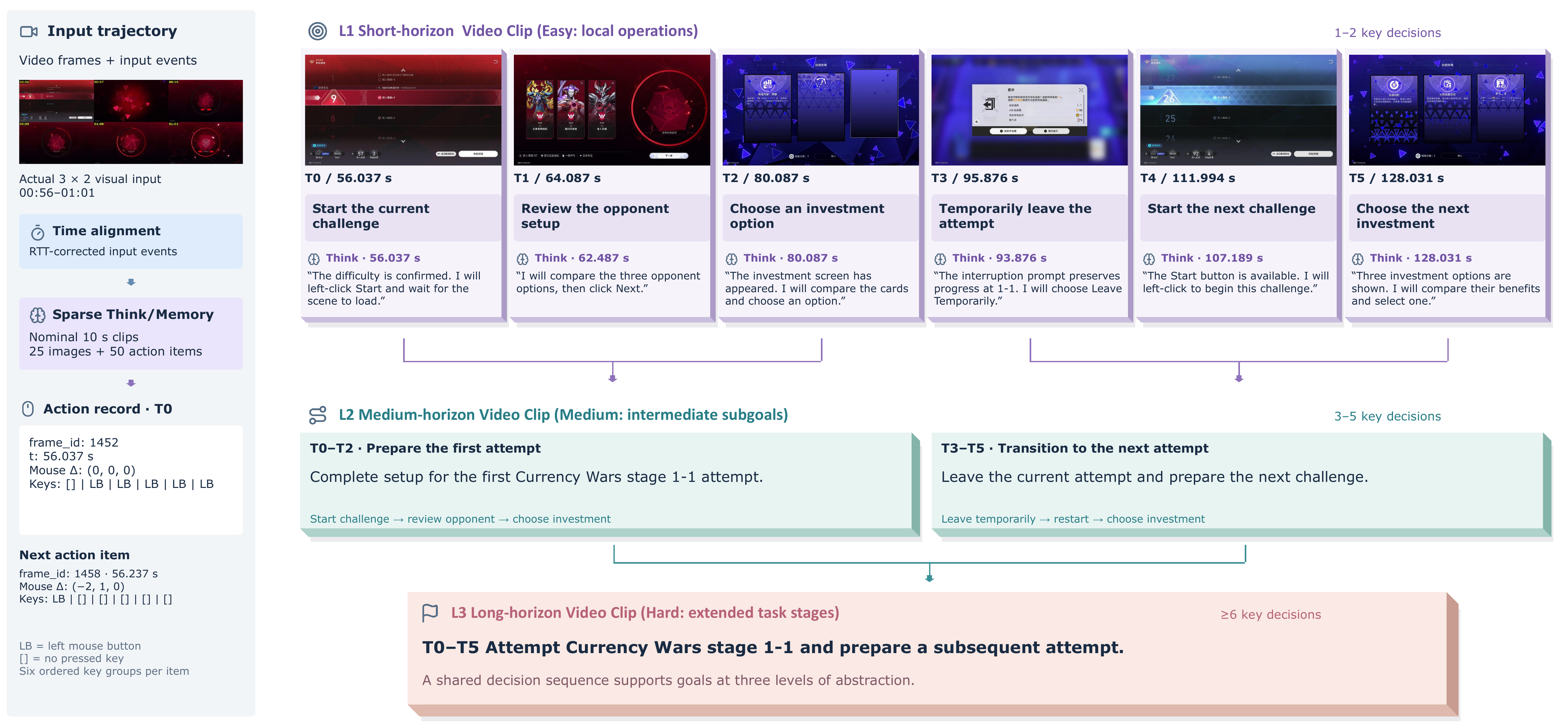}
    \caption{
    \textbf{Bottom-up construction of multi-level task supervision.}
    The diagram illustrates the organization of aligned gameplay segments into local operations, intermediate subgoals, and extended task stages. Separate GPT-5.6-Luna passes produce sparse Think annotations, segment-level instructions and scene descriptions, and memory updates as detailed in Secs.~\ref{sec:data_annotation}-\ref{sec:data_memory}. The diagram focuses on hierarchy assembly rather than depicting every annotation call.
    }
    \label{fig:data_construction}
\end{figure*}

Raw gameplay videos provide rich visual and control experience, but they rarely specify the player's objective, the decisions that require reasoning, or the information that should be retained beyond the recent context. We develop a scalable annotation pipeline that turns heterogeneous recordings into supervision for long-horizon task following, as shown in Figure~\ref{fig:data_construction}. Starting from tens of thousands of hours of raw gameplay collected across different sources, we filter invalid or misaligned recordings and balance coverage across gameplay scenarios. For SFT, we curate approximately 1K hours of gameplay trajectories. GPT-5.6-Luna~\cite{gpt56card} generates sparse action reasoning, task segments with instructions and scene descriptions, and task-state memories. We organize these annotations into three instruction levels: local operations, intermediate subgoals, and extended task stages.

\subsection{Trajectory Preparation and Temporal Alignment}
\label{sec:data_harmonization}

\textbf{Data sources, filtering, and scenario balancing.}
We collect raw gameplay recordings with associated keyboard and mouse events from multiple sources. Within this collection, the HSR recording track contains roughly 68K sessions and 11K hours of video. Rather than annotating all available videos, we first select recordings with reliable visual and action supervision. We reject corrupted or duplicated sessions, missing or discontinuous frames, and trajectories whose video and action timestamps cannot be aligned consistently. We also remove invalid action traces, including mouse coordinates outside the valid screen or control bounds, malformed action states, and events that cross recording or interaction boundaries. Long inactive periods and recordings without meaningful interaction are filtered, while valid held inputs and task-related waiting periods are retained.

Filtering alone would leave the instruction corpus dominated by frequently recorded activities. We therefore balance scenario coverage by limiting repeated behaviors and retaining less frequent interactions and longer task transitions. 
After these quality and coverage checks, we obtain \textbf{approximately 3.2K sessions and 1K hours of video}, which we use as the source for SFT annotation.

\textbf{Action-observation alignment.}
We retain the original ordering of keyboard and mouse events and associate each action block with its corresponding frame using corrected recording timestamps. Each device-level action interval spans approximately $200$ ms and contains the six finer control blocks defined in Sec.~\ref{sec:unified_action}. Source videos are sampled to obtain the visual observations used for annotation. When frame indices are discontinuous, or timestamps show a recording gap, an action window is not allowed to cross that boundary. Short button presses are kept as short pulses rather than being dropped by temporal aggregation. This preserves movement duration and interaction timing while producing the same executable action interface used by the policy. 

\subsection{Sparse Reasoning and Visual Task Segmentation}
\label{sec:data_annotation}

The annotation process separates local action reasoning from longer visual task understanding. Short video-action windows identify key decisions, while longer visual clips reveal task structure and progress. We use GPT-5.6-Luna for both annotation passes and align their outputs by the original frame IDs and recording timestamps. 

\textbf{Decision-focused Thinking.}
We divide the action stream into windows of 50 device-level action intervals, corresponding to roughly 10 seconds of interaction. Each window contains 25 ordered game observations and the associated 50 action intervals. GPT-5.6-Luna identifies a sparse set of decision points, typically when the player selects or corrects a route, handles a new interaction prompt, changes objective, recovers from an unsuccessful attempt, or responds to a significant scene transition. Each \texttt{Think} annotation describes the observed outcome, the current decision, and an intended next action. Continuous movement, repeated controls, and minor camera corrections usually receive no separate reasoning target. Windows with neither meaningful input nor visual change can be skipped.
This process yields approximately \textbf{1.2M timestamp-aligned Think annotations}.

\textbf{Long-context instruction and scene annotation.}
We divide each session into the minimum number of equal-length visual clips of at most 288 seconds.
Frames are selected at approximately one-second intervals and arranged into $3\times2$ timestamped mosaics, yielding at most 49 mosaic images per clip. This lets GPT-5.6-Luna inspect continuous visual context without exceeding the image-input budget. The model identifies contiguous task segments and assigns each a task \texttt{instruction}, a gameplay \texttt{label}, and a short \texttt{scene} description of the visible evidence. These task segments provide the temporal and semantic basis for the instruction hierarchy.

We prompt the model to segment gameplay by the player's current task. Brief movements and reward collection are grouped with the task they support, while an explicit challenge restart begins a new segment. Postprocessing adjusts adjacent segment boundaries to form a continuous timeline and converts clip-relative start and end times to timestamps in the original recording. This yields approximately \textbf{63K task segments}, with a median duration of about 40 seconds. The activity labels cover seven gameplay categories: navigation, NPC dialogue, combat, minigames, system menus, Currency Wars, and other activities.

\subsection{Memory Annotation from Continuous Visual Context}
\label{sec:data_memory}

After task segmentation, we reuse the clip images and cleaned task segments in a second conversational turn to generate memory annotations. The model returns all summaries for the clip in one response, typically one per 16-second window. Each summary briefly describes, in the first person, the cumulative accomplishments and outcomes within the clip up to the window end. A final window shorter than 8 seconds is merged into the preceding window. Postprocessing associates each summary with the active task instruction and the latest image frame at or before the window end.

The dataset contains approximately \textbf{223K memory annotations}. At each training step, we condition the policy on the latest summary whose end timestamp is no later than the current observation. The annotation model receives the complete visual clip and is prompted to summarize only events up to each window end.

\subsection{Three-Level Instruction Construction}
\label{sec:multi-level_supervision}

Building on the first-pass task segments, we construct \textbf{easy, medium, and hard instructions} according to the scope of the objective and the number of annotated decisions it covers. These levels correspond to local operations, intermediate subgoals, and extended task stages, respectively. The same recorded interval can support instructions at multiple levels.

\textbf{Easy: local operations.}
Easy instructions cover one or two annotated decisions, such as approaching an NPC, selecting a menu option, or collecting an item. \textbf{Medium: intermediate subgoals.} Medium instructions combine related operations into a subgoal spanning three to five annotated decisions, such as navigating around an obstruction and completing the required interaction. \textbf{Hard: extended task stages.} Hard instructions connect multiple subgoals into a task stage spanning six or more annotated decisions, such as navigating through several areas and resolving intervening challenges.

We construct the instruction hierarchy bottom-up from temporally aligned annotations. Task-segment instructions and scene descriptions establish the activity and its boundaries, Think annotations identify key decisions, and memories track task progress. We group adjacent operations that advance a common subgoal, then combine related subgoals into extended task stages. Human-annotated examples guide the wording and scope of the generated instructions. All three levels share the \textbf{recorded actions and Think annotations}, and each provides an instruction and memory context appropriate to its objective.

\textbf{Horizon-dependent training targets.}
For a selected supervision horizon $h\in\{\mathrm{op},\mathrm{sub},\mathrm{task}\}$, an example is represented as
\begin{equation}
    x_t^{h}=\left(\mathcal{B},I^{h},\mathcal{M}_t^{h},\mathcal{H}_t\right),
    \qquad y_t=[r_t,a_t],
    \label{eq:multi-level_supervision}
\end{equation}
where $I^{h}$ denotes the task instruction at level $h$, $\mathcal{M}_t^{h}$ denotes its task memory, and $\mathcal{H}_t$ contains the current observation and recent interaction history. The reasoning text $r_t$ and action target $a_t$ are drawn from the same recorded trajectory and its annotations. Every step includes action supervision, including steps without Think annotations.

 Unlike GameHorizon\cite{wang2026gamehorizon}, \ourmodel builds high level instruction data from aligned gameplay videos and actions uses a separate annotation passes to identify task boundaries and label decision points. The first pass identifies task boundaries within visual clips. The second examines short video-action windows and adds sparse \texttt{Think} annotations at decision points (Sec.~\ref{sec:data_annotation}). Each training example provides supervision for keyboard and mouse actions at every step and for reasoning at selected decision points. Timestamped summaries supply context on task progress (Sec.~\ref{sec:data_memory}).


\section{Implementation Details}
\label{sec:implementation}

\subsection{Model and Training Configuration}
\label{sec:model_configuration}
\label{sec:training_configuration}
\label{sec:train_infer_consistency}

\ourmodel supports both dense and mixture-of-experts (MoE) architectures across Qwen3-VL and Qwen3.5 families. The trained models share the same autoregressive keyboard-mouse action interface without an additional output head. We first analyze CPT scaling across 120 controlled Qwen3.5 Dense configurations, referred to as \textbf{GameScaling}. Guided by these findings, we train \ourmodel from Qwen3-VL-8B with additional HSR gameplay data.

Training follows three stages while preserving the model architecture and action representation. CPT learns transferable control with key remapping and repeat-aware loss decay. SFT introduces multi-level task supervision, sparse reasoning, and cross-context memory. TA-GRPO then improves long-horizon behavior through online interaction, using environment-verifiable rewards and temporal credit assignment. Gameplay trajectories are sampled at 5~Hz, with six action blocks per step providing approximately 30~Hz control and a local visual history of 20 frames.

We extend Verl~\cite{sheng2024hybridflow} into a unified training and evaluation framework, using \textbf{Streaming-vLLM} for interactive evaluation and online RL rollouts. A sharded inference service pool backed by \textbf{64 NVIDIA H200 GPUs} serves 32 parallel game environments. Each RL iteration collects 128 rollouts in groups of 8, with environment-verifiable rewards and no learned process reward model. Updated policy weights are synchronized across the service pool before the next rollout collection, keeping policy versions aligned throughout online training. Streaming-vLLM also supports dense and MoE backbones on NVIDIA A800 and H200 GPUs. Further configuration details are provided in Table~\ref{tab:training_recipe} and Appendix~\ref{app:unified_training_inference}.

\begin{figure*}[!htbp]
    \centering
    \includegraphics[width=0.88\linewidth]{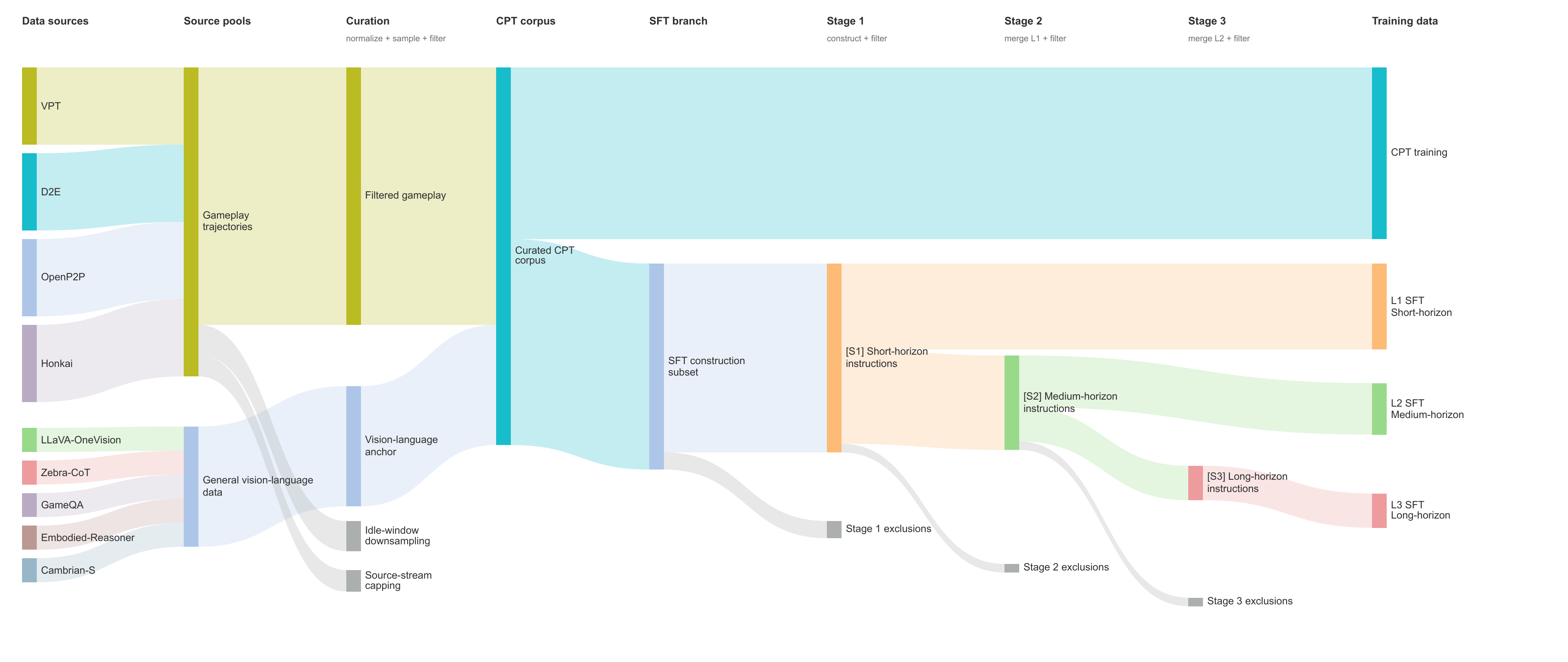}
    \caption{
    \textbf{Multi-stage data curation for training.}
    Gameplay trajectories and general vision-language data are cleaned, sampled, and filtered to construct the CPT corpus.
    Trajectories with aligned action annotations are further used to construct SFT examples at three levels: short-horizon operations, medium-horizon subgoals, and long-horizon tasks.
    For RL, online policy interaction provides trajectories and environment-verifiable outcomes.
    }
    \label{fig:training_data_flow}
\end{figure*}

The controlled scaling study spans 120 CPT configurations and approximately 45K H200 GPU-hours. Shorter training budgets are derived from shared Warmup-Stable-Decay trajectories with separate final decay phases. For the subsequent \ourmodel training stages, CPT uses a distributed H200 cluster, SFT uses a distributed A800 cluster, and online RL uses Megatron/MCore with BF16 optimization. Real-time rollouts run in parallel game environments served by a sharded Streaming-vLLM pool backed by 64 H200 GPUs. The pool receives synchronized policy checkpoints between rollout iterations, enabling consistent interaction data collection as training progresses. More details of training configurations and stage-specific settings are summarized in Table~\ref{tab:training_recipe}.

\subsection{Data Composition and Sampling}
\label{sec:data_composition}

Our training data combines cross-game interaction trajectories, general multimodal data, multi-level task supervision, and online policy experience. Figure ~\ref{fig:training_data_flow} summarizes how these sources are filtered and routed into the CPT, SFT, and RL stages, while Table~\ref{tab:data_summary} reports their corresponding usage and scale.

In the CPT stage, we perform controlled scaling experiments (i.e., GameScaling) using a mixture of VPT~\cite{baker2022vpt}, D2E~\cite{choi2026d2e}, and OpenP2P~\cite{yue2026openp2p} gameplay data while holding the token ratio at 85\% gameplay and 15\% general vision-language data (Appendix~\ref{app:cpt_diversity}). General multimodal data sources include LLaVA-OneVision~\cite{llava-o}, Zebra-CoT~\cite{li2026zebracot}, GameQA~\cite{tong2026gamerl}, Embodied-Reasoner~\cite{zhang2025embodiedreasoner}, and Cambrian-S~\cite{yang2025cambrians}. \ourmodel's CPT stage retains this data mixture ratio and adds additional HSR gameplay trajectories.

In the SFT stage, we annotate approximately 1K hours of curated HSR gameplay trajectories. As described in Sec.~\ref{sec:multi-level_supervision}, operation-level examples train local control and sparse reasoning. Subgoal-level examples add intermediate objectives and longer memory, while task-level examples cover complete long-horizon tasks. Examples at different annotation levels may come from the same trajectory and we count their derived samples and supervision tokens separately. 

To improve coverage across gameplay scenarios, we downsample overrepresented streams and excessively idle intervals while preserving meaningful movement, held controls, and waiting periods. In the HSR portion of the CPT corpus, filtering retains approximately 149K aligned clips, corresponding to around 332 hours of gameplay. Data splits are assigned at the recording or complete-task level before creating overlapping windows or multi-level examples, so derived samples from a source remain in the same split.

\begin{table*}[t]
\centering
\small
\caption{
\textbf{Overview of training data used for building \ourmodel.}
The table summarizes the major data sources, their role in each training stage, and the corresponding data scale.
}
\label{tab:data_summary}

\resizebox{0.98\linewidth}{!}{%
\begin{tabular}{lllcc}
\toprule

\textbf{Stage}
& \textbf{Data}
& \textbf{Supervision / Usage}
& \textbf{Size}
& \textbf{Sampling} \\

\midrule

\multirow{4}{*}{CPT}
& VPT, D2E, OpenP2P
& \multirow{2}{*}{Cross-game generalization}
& \multirow{2}{*}{ \ensuremath{\sim}2.4K hours}
&  \multirow{2}{*}{85\%} \\
& Honkai: Star Rail (HSR) 
& & & \\
\cmidrule{2-5}

& LLaVA-OneVision, Zebra-CoT, GameQA
  
& Vision–language 
& \multirow{2}{*}{\ensuremath{\sim}6.3B tokens}
& \multirow{2}{*}{15\%} \\

& Embodied-Reasoner, Cambrian-S
& Embodied and egocentric 
& & \\

\midrule

\multirow{3}{*}{SFT}
& Operation-Level HSR
& Local operations
& \ensuremath{\sim}261K samples
& 37.0\% \\

\cmidrule{2-5}

& Subgoal-Level HSR
& Intermediate subgoals
& \ensuremath{\sim}211K samples
& 29.8\% \\

\cmidrule{2-5}

& Task-Level HSR
& Long-horizon task
& \ensuremath{\sim}234K samples
& 33.1\% \\

\midrule

RL
& Verified Online Trajectories
& Policy optimization
& \ensuremath{\sim}14.6K rollouts
& -- \\

\bottomrule
\end{tabular}
}
\vspace{-4pt}
\end{table*}

\subsection{Automated RL Environment}
\label{sec:rl_environment}

\textbf{A shared infrastructure for learning, evaluation, and testing.}
Many existing game-RL benchmarks run as lightweight, self-contained simulators that can be reset cheaply and queried directly for rewards. Modern commercial games present a substantially harder setting. In particular, Honkai: Star Rail depends on a full game client, supporting online services, GPU-accelerated rendering, and complex game-state dependencies. Reproducing its native mechanics in a lightweight simulator would change the interaction setting, while simply replicating game clients does not provide controlled initialization or objective verification. These challenges extend beyond online RL to large-scale automated evaluation and gameplay testing, where hundreds of executions must start from comparable states and be judged against consistent success criteria. We therefore build a dedicated, engine-instrumented sandbox that makes this resource-intensive game reproducible, verifiable, and scalable without changing the agent's visual-control interface.

\textbf{Game-native rules as a unified source of feedback.}
We adapt the game engine to expose task-relevant state transitions and completion conditions to the environment backend. The resulting rule-based verifiers serve three complementary purposes. During online RL, they provide environment-verifiable rewards for TA-GRPO without a learned process reward model. During evaluation, they measure task completion objectively across models and policy checkpoints. For automated gameplay testing, the same checks provide reproducible pass/fail criteria for interactions and quest progression. This shared verification layer avoids relying on subjective assessments of generated text or inferring complex task outcomes from individual screenshots.

\textbf{GM-controlled initialization and reproducible execution.}
We integrate game-master (GM) commands to prepare predefined task states and restore the corresponding conditions before an episode. Each task pairs an initialization procedure with a target state, a completion check, and a termination rule. This allows the same objective to be executed repeatedly for policy optimization, fair comparison, and systematic testing of game behaviors. It also makes failed executions easier to reproduce and inspect. Crucially, all privileged engine access remains within the environment backend. The policy receives only screen observations and task context, and its actions are executed through the same keyboard--mouse interface used in real gameplay.

\textbf{Scalable execution across use cases.}
We construct verifiable environments spanning navigation, dialogue and interaction, and game-specific puzzle mechanics. For online RL, 32 parallel Windows game environments are served by a sharded Streaming-vLLM inference pool backed by 64 NVIDIA H200 GPUs. Each iteration collects 128 trajectories in groups of eight, with policy checkpoints synchronized before the next collection round. The same initialization, execution, and verification interfaces also support batched evaluation and automated regression tests without manually resetting or judging each attempt. Beyond enabling real-time policy optimization, this infrastructure provides a practical foundation for scalable assessment and testing in modern commercial games. Sec.~\ref{sec:evaluation_setup} introduces the held-out side-quest and minigame evaluations.

\section{Experiments}
\label{sec:experiments}

Our evaluation examines three complementary questions. First, we summarize the study in GameScaling, which evaluates 120 CPT configurations to characterize the effects of model scale, gameplay budget, and data composition. We then assess the executable cross-game performance of CPT policies built on Dense and MoE backbones. Next, we evaluate the separately trained HSR-augmented \ourmodel-CPT and its SFT and RL successors on HSR side quests and minigames, separating broad interaction skills from long-horizon task execution. Finally, we test whether real-time rollout collection supports online learning and whether Streaming-vLLM delivers the accuracy and latency needed for sustained interaction. All interactive evaluations use the shared keyboard-mouse interface.

\subsection{Evaluation Setup}
\label{sec:evaluation_setup}

\paragraph{Public game benchmarks.}
We evaluate game-control policies on MCU~\cite{zheng2023mcu}, ViZDoom~\cite{Wydmuch2019ViZdoom}, OmniGameArena~\cite{omnigamearena2026}, and GameWorld~\cite{gameworld2026}.
These benchmarks vary in how far they differ from the gameplay data used for CPT.
Specifically, MCU and ViZDoom evaluate game domains represented during CPT and primarily measure control learning within training-covered environments.
OmniGameArena evaluates transfer to unseen three-dimensional games with related interaction semantics.
GameWorld provides a more distant cross-game transfer setting with larger shifts in visual appearance, control mechanics, and task structure.

We follow the task definitions, environment settings, and native scoring protocols of the corresponding benchmarks.
Each benchmark retains its original evaluation metric.
We distinguish idealized time-paused (TP) evaluation from real-time (RT) interaction. TP pauses the environment during model inference, whereas RT allows the game to continue advancing. In the public game comparison, we report the RT score as the main value and its signed difference from TP (RT$-$TP) in parentheses.

Across game evaluations, we use the same set of strong multimodal baselines wherever they support the required visual-control interface.
These include GPT-6.1-Sol~\cite{openai2026gpt61sol}, Gemini-3.8-Flash~\cite{google2026gemini38flash}, GPT-5.6-Sol~\cite{gpt56card}, GPT-5.6-Luna~\cite{gpt56card}, Claude-Opus-5.0~\cite{anthropic2026opus5}, GLM-5.3-Flash~\cite{glm53flash}, and DeepSeek-V4-Flash-Vision~\cite{deepseekv4flashvision}.
The same baseline pool is reused in the Honkai: Star Rail side-quest and minigame evaluations to provide a consistent comparison across control, instruction following, and gameplay generalization. 

\begin{figure*}[!htbp]
    \centering
    \begin{minipage}[t]{0.80\linewidth}
        \centering
        \includegraphics[width=\linewidth]{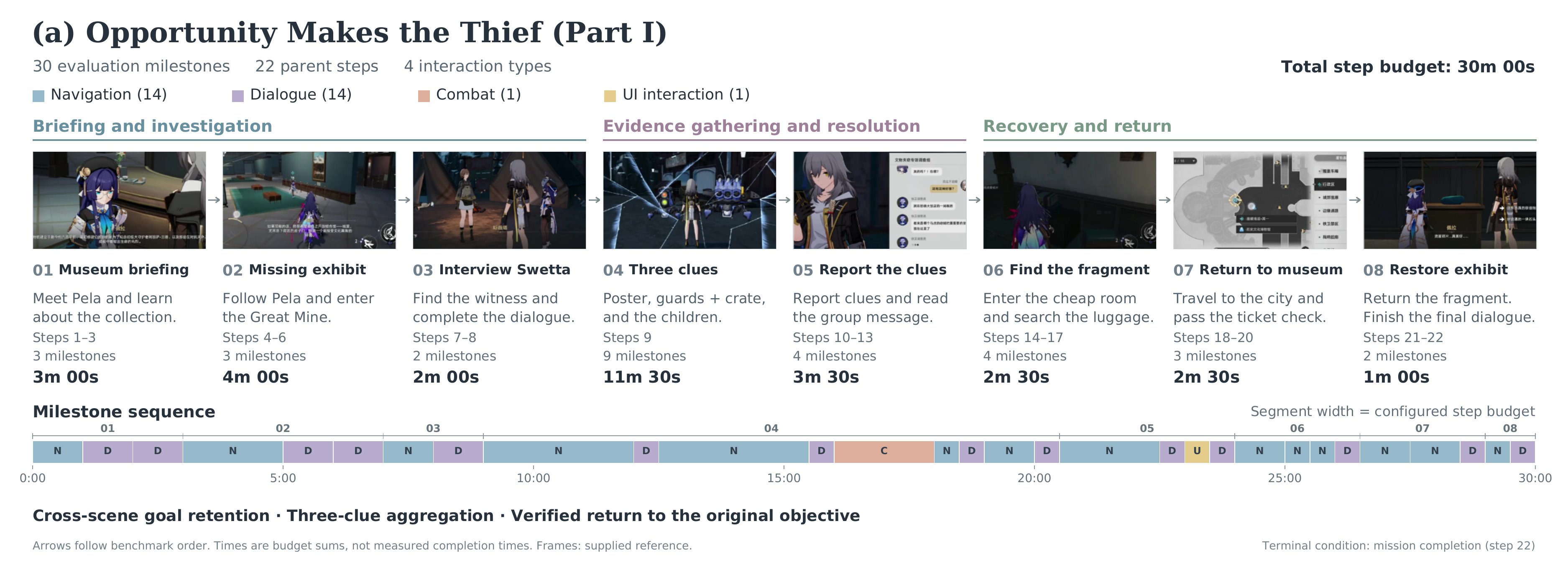}
        \vspace{-2mm}
        \centerline{\textbf{(a) Opportunity Makes the Thief (Part \uppercase\expandafter{\romannumeral1})}}
    \end{minipage}
    \\
    \vspace{2mm}
    \begin{minipage}[t]{0.80\linewidth}
        \centering
        \includegraphics[width=\linewidth]{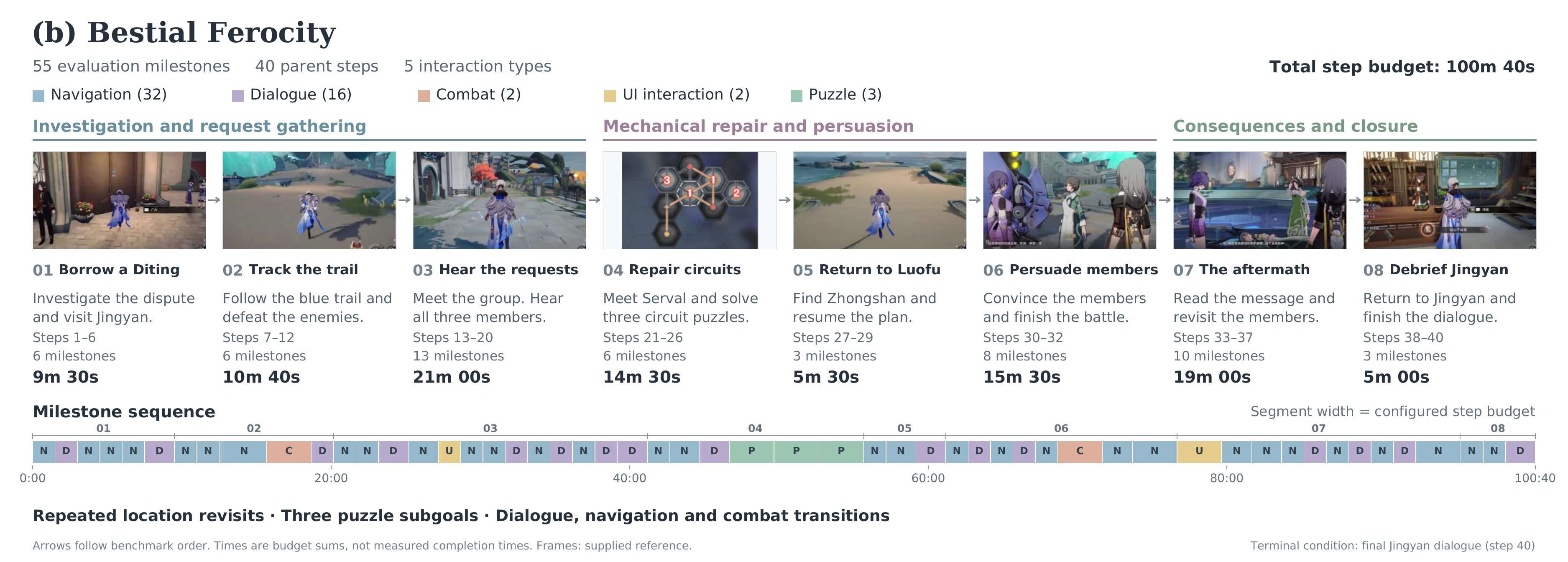}
        \vspace{-2mm}
        \centerline{\textbf{(b) Bestial Ferocity}}
    \end{minipage}
    \caption{
    \textbf{Representative long-horizon side quests in the Honkai: Star Rail evaluation suite.}
  (a) Opportunity Makes the Thief (Part I) contains 30 milestones with navigation, dialogue, combat, and UI interaction. (b) Bestial
Ferocity contains 55 milestones and additionally includes three mechanical-repair puzzles. Both panels show ordered milestone sequences and configured step budgets rather than measured completion times.
    }
    \label{fig:hsr_sidequests}
\end{figure*}

\paragraph{Honkai: Star Rail evaluation suite.}
We construct two complementary evaluation sets in Honkai: Star Rail, comprising side quests and minigame tasks held out from training (Figure~\ref{fig:hsr_sidequests}). Although the final CPT model has seen other gameplay from the same game, it has not been trained on these evaluation tasks or their target completions. This setting tests generalization to new tasks within a familiar game, rather than zero-shot transfer to an unseen game.

The first set contains two long-horizon \textbf{side quests} that test sustained progress on complex instructions. Each quest may combine navigation, dialogue, combat, puzzles, and item interactions across several intermediate objectives.
To finish a quest, the agent must remember quest information, track subgoal progress, and use that task state to guide low-level control.

The two evaluated quests are \emph{Opportunity Makes the Thief (Part I)} and \emph{Bestial Ferocity}. They contain 30 and 55 evaluation milestones, grouped into 22 and 40 parent steps, respectively. The first quest combines navigation, dialogue, combat, and UI interaction. The second also includes three automaton-repair puzzles and repeated visits to earlier locations. The total time budgets across milestones are 30 minutes and 100 minutes, respectively.

Each quest has a fixed initial state, task objective, milestone sequence, verifier, and termination condition. We report accuracy as the fraction of evaluated task executions that succeed, both overall and for each interaction type. The milestone inventory covers navigation, dialogue, combat, UI interaction, and, where applicable, puzzle solving. These rates measure verified progress across task executions rather than uninterrupted end-to-end quest completion. The complete milestones, types, and budgets are given in Appendix~\ref{app:hsr_protocol}.

\begin{figure*}[!htbp]
    \centering
    \includegraphics[width=0.87\linewidth]{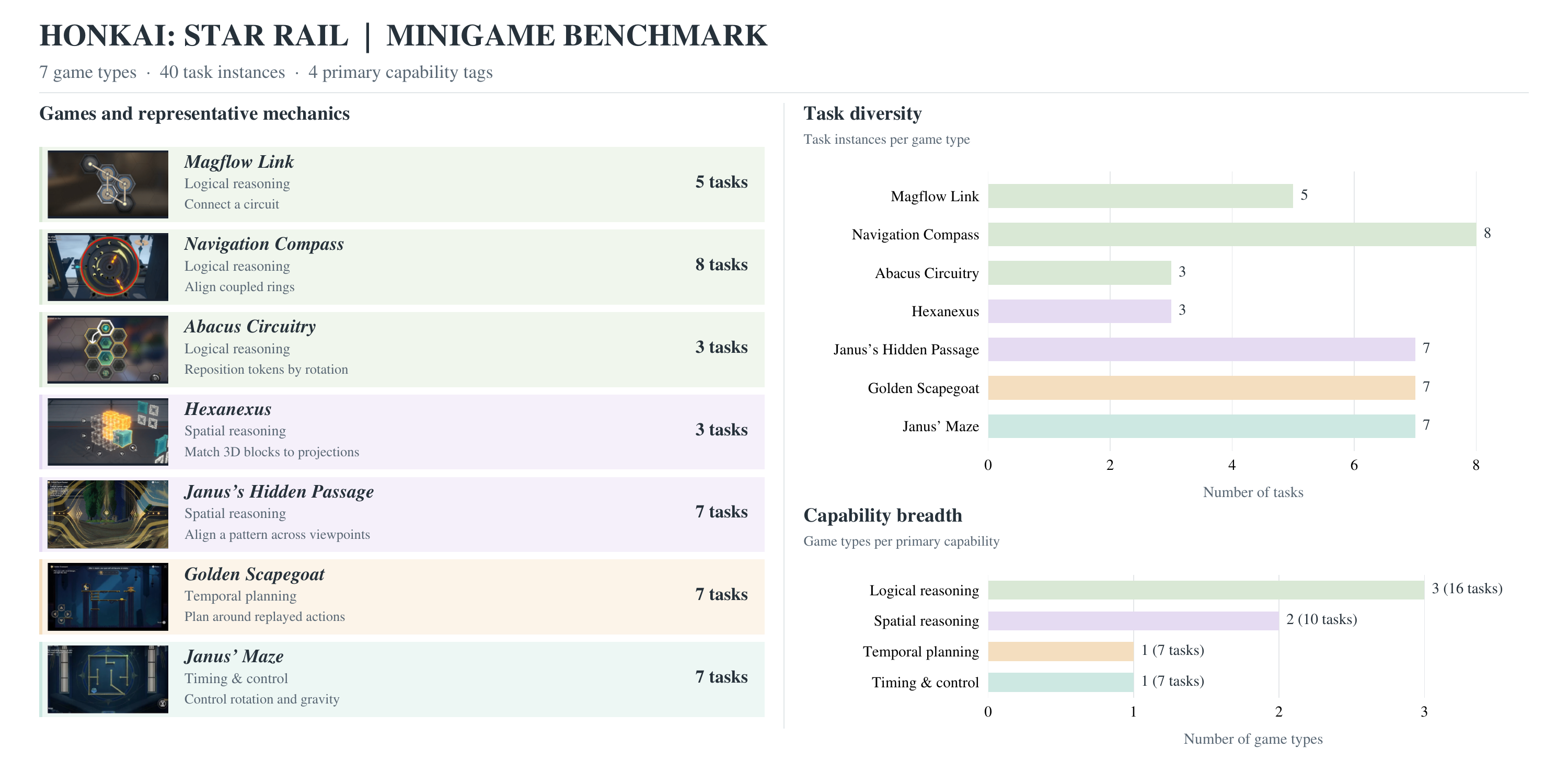}
    \caption{
    \textbf{Overview of the minigame benchmark in the Honkai: Star Rail evaluation suite.} The benchmark contains 40 task instances across seven game types. The left panel shows representative gameplay examples, instance counts, and mechanics. The upper-right chart shows the number of instances per type, while the lower-right chart groups minigames by four descriptive primary capability tags, with instance totals in parentheses. Each minigame type is assigned one primary tag for counting and may require additional abilities. Screenshots illustrate game types and are not necessarily drawn from the evaluated instances.
    }
    \label{fig:hsr_minigames}
\end{figure*}

The second subset contains 40 instances across 7 \textbf{minigame} types, covering pattern composition, logical reasoning, spatial reasoning, temporal planning, and timing-sensitive control. Each type contains between one and sixteen evaluated instances. Figure~\ref{fig:hsr_minigames} illustrates the minigame mechanics and instance distribution. We report the success rates for each type and compute overall accuracy as the unweighted mean of the seven type-level rates. Each type therefore contributes equally, regardless of its number of instances. Per-instance evaluation and taxonomy are described in Appendix~\ref{app:hsr_protocol}.

\subsection{Scaling Cross-Game Control with CPT}
\label{sec:exp_cpt_scaling}

\subsubsection{Controlled Scaling Setup}

The controlled study in GameScaling \cite{gamescaling} examines how model size, gameplay budget, and game composition affect CPT performance and cross-game generalization. These results guide further HSR-augmented \ourmodel-CPT recipes. We summarize its experimental setup here.

As summarized in Table~\ref{tab:cpt_scaling_setup}, the study comprises $4\times6\times5=120$ configurations spanning Qwen3.5 backbone sizes, gameplay budgets, and data mixtures. All other training settings are held fixed. Across configurations, gameplay trajectories account for 85\% of training tokens, while the remaining 15\% come from a fixed pool of general multimodal data. The five mixtures, S1--S5, vary only the composition of the gameplay portion: Minecraft accounts for 100\%, 51.4\%, 27.1\%, 2.8\%, and 0\%, respectively. The remaining gameplay data in S2--S5 are sampled from the same multi-game pool.

\begin{table}[t]
    \centering
    \small
    \setlength{\tabcolsep}{4pt}
    \renewcommand{\arraystretch}{1.08}
    \begin{tabular}{p{0.20\linewidth}p{0.40\linewidth}p{0.31\linewidth}}
        \toprule
        \textbf{Scaling Dimension}
        & \textbf{Controlled Configurations}
        & \textbf{Research Focus} \\
        \midrule
        Model size $N$
        & Qwen3.5 0.8B, 2B, 9B, 27B
        & Capacity and data efficiency \\
        
        Gameplay budget $D$
        & 50, 100, 200, 400, 800, 1,600 hours
        & Gains from additional interaction \\
        
        Game mixture
        & S1-S5
        & Effect of target-game coverage \\
        \bottomrule
    \end{tabular}
    \caption{
    \textbf{Experimental configurations for the GameScaling CPT study.} The study varies model size, gameplay budget, and game mixture across 120 configurations.
    }
    \label{tab:cpt_scaling_setup}
\end{table}

Let $w_{\mathrm{MC}}$ denote the Minecraft fraction of the gameplay mixture.
Consistent with Sec.~\ref{sec:cpt_scaling}, we use
$q=w_{\mathrm{MC}}$
for Minecraft validation loss, and
$q=1-w_{\mathrm{MC}}$
for validation over the multi-game training distribution.
The latter denotes the non-Minecraft fraction of the gameplay experience.

\subsubsection{Interface Adaptation and Scaling Behavior}

We summarize the interface-adaptation and scaling results reported in GameScaling~\cite{gamescaling}. The corresponding plots are provided in Appendix~\ref{app:gamescaling_results}.

\paragraph{Early CPT adapts the pretrained model to the shared control interface.}
Figure~\ref{fig:cpt_action_losses} shows that keyboard, mouse, and action-format losses decrease rapidly during the early stage of CPT, while directional prediction accuracy steadily improves.
Checkpoints trained for only 50 gameplay hours already reduce validation loss on both training and unseen games relative to their corresponding base models. This early improvement across distinct action components suggests that a substantial part of initial CPT learning concerns the mechanics of executable control. It also explains why interface adaptation can precede the larger gameplay-specific gains that emerge with additional interaction experience.

\paragraph{Later scaling gains increasingly depend on game composition.}
Figure~\ref{fig:cpt_scaling_losses} separates training fit from cross-game generalization.
Minecraft-heavy mixtures S1-S3 continue to reduce Minecraft validation loss as the interaction budget increases, while the low-Minecraft mixtures S4 and S5 tend to saturate.
On the remaining training games, multi-game mixtures continue to improve while Minecraft-only training eventually plateaus or degrades.
Adding the same amount of gameplay data can therefore produce different gains, depending on its coverage of the target distribution.

Larger models learn more efficiently early in training, but their gains still depend on data composition.
Under single-game training, stronger adaptation to Minecraft can coincide with degraded performance on other games.
Multi-game mixtures alleviate this effect.
For distant unseen games, validation loss typically improves during early interface adaptation and then plateaus or increases even while training loss continues to decrease.
These observations motivate considering model capacity, interaction budget, and target-game coverage jointly.

Evaluation results on game benchmarks complement the validation-loss analysis with executable gameplay outcomes. As shown in Figure~\ref{fig:cpt_benchmark_hours}, additional CPT experience generally improves performance in training-covered and behaviorally related environments, particularly at smaller model scales. The comparison across S1-S5 also shows why the amount of gameplay alone is an incomplete description of the training signal. Mixtures that expose the model to multiple interaction patterns provide broader coverage than single-game training and support transfer to related environments. For Gamer, these findings motivate scaling game diversity alongside data volume rather than treating all recorded gameplay hours as equally informative for every target game.

\subsubsection{Scaling-Law Fit and Predictive Range}

GameScaling~\cite{gamescaling} fits the empirical relation summarized in Sec.~\ref{sec:cpt_scaling},
\begin{equation}
\mathcal{L}_{\mathrm{val}}(N,D,q)
=
E
+
\frac{C_N}{N^a}
+
\frac{C_D q^\eta}{D^b}
+
\frac{C_q}{(q+\epsilon)^c},
\label{eq:exp_scaling_law}
\end{equation}
to characterize validation loss as a function of model size, gameplay budget, and target-dependent data composition.

The study reports that the standard model-and-data scaling relation fits training loss with $R^2=0.980$ and RMSE $=0.0041$.
For Minecraft and the multi-game training distribution, the composition-aware relation achieves the fit accuracy reported in Table~\ref{tab:scaling_fit}.

\begin{table}[t]
    \centering
    \small
    \setlength{\tabcolsep}{4pt}
    \begin{tabular}{lcccc}
        \toprule
        \textbf{Validation}
        & \textbf{$q$}
        & \textbf{$R^2$}
        & \textbf{RMSE}
        & \textbf{Within $R^2$} \\
        \midrule
        Minecraft
        & $w_{\mathrm{MC}}$
        & 0.982
        & 0.0020
        & 0.912 \\
        
        Training games
        & $1-w_{\mathrm{MC}}$
        & 0.995
        & 0.0012
        & 0.959 \\
        \bottomrule
    \end{tabular}
    \caption{
    \textbf{Fit accuracy of the composition-aware scaling relation.}
    Within-mixture $R^2$ measures variation explained by model size and data budget after removing mean differences across mixtures.
    }
    \label{tab:scaling_fit}
\end{table}

Both validation distributions yield a positive fitted $\eta$. At a fixed model size, increasing the interaction budget therefore produces larger gains when the training mixture contains more experience relevant to the validation distribution. The fit captures the same interaction between data quantity and composition visible in the measured curves. In particular, it distinguishes a model that has seen more gameplay from one that has seen more gameplay relevant to the target environment. This provides a practical basis for choosing CPT mixtures when expanding the training corpus.

GameScaling further performs block-wise holdout tests over model sizes, data budgets, and game mixtures.
Capacity and budget holdouts support prediction within the measured range.
Mixture prediction is more sensitive to the availability of nearby compositions.
The fitted relation summarizes the observed scaling regime across the evaluated distributions.
Fitting details and holdout results are provided in Appendix~\ref{app:cpt_scaling_fit}.

Distant unseen games provide a useful negative control.
Their validation loss rises again after early adaptation.
Fitting the same functional form yields a data exponent close to $b=0$.
The nominal $R^2=0.855$ is dominated by static differences across mixtures, while the within-mixture $R^2$ is only 0.079.
A high overall fit score therefore does not imply continued improvement from additional interaction data.

\subsubsection{Cross-Game Performance of CPT Policies}

While the controlled GameScaling study reveals how model size and gameplay data affect CPT, an important question remains: whether CPT can translate general-purpose visual understanding into executable control across diverse games. Unlike static visual reasoning, gameplay requires the agent to continuously interpret observations, execute precise actions, and adapt to evolving environments. We therefore evaluate whether CPT-trained policies can achieve effective real-time control across different games and backbone architectures. Table~\ref{tab:cpt_game_benchmarks} compares two CPT checkpoints, initialized from Qwen3-VL-8B Dense and Qwen3.5-35B-A3B MoE, with general-purpose multimodal baselines on four public game benchmarks. We report real-time (RT) scores as the primary results, together with their signed differences from time-paused (TP) evaluation.

\begin{table*}[t]
\centering
\small
\setlength{\tabcolsep}{4.5pt}
\renewcommand{\arraystretch}{1.06}
\caption{
\textbf{Cross-game performance of CPT policies under real-time interaction.}
The final two rows report CPT checkpoints based on Qwen3-VL-8B Dense and Qwen3.5-35B-A3B MoE. The main value is the real-time (RT) score, while parentheses show its difference from idealized time-paused (TP) evaluation, computed as RT$-$TP. Green indicates a positive difference, red a negative difference, and zero denotes no change. Higher scores are better. The performance of controlled GameScaling-S3 checkpoints is not included in this table.
}
\label{tab:cpt_game_benchmarks}

\resizebox{0.92\linewidth}{!}{
\begin{tabular}{lccccc}
\toprule
\textbf{Model}
& \textbf{MCU $\uparrow$}
& \textbf{ViZDoom $\uparrow$}
& \textbf{OmniGameArena $\uparrow$}
& \textbf{GameWorld $\uparrow$}
& \textbf{Average $\uparrow$} \\
\midrule

GPT-5.6-Sol
& 8.7 (\textcolor{red}{$-$73.3})
& 2.7 (\textcolor{red}{$-$24.0})
& 14.5 (\textcolor{red}{$-$31.1})
& 44.3 (\textcolor{red}{$-$9.6})
& 17.6 (\textcolor{red}{$-$34.5}) \\

GPT-5.6-Luna
& 5.3 (\textcolor{red}{$-$48.7})
& 2.9 (\textcolor{red}{$-$21.2})
& 13.2 (\textcolor{red}{$-$13.5})
& 41.6 (\textcolor{red}{$-$3.3})
& 15.8 (\textcolor{red}{$-$21.6}) \\

Claude-Opus-5.0
& 8.7 (\textcolor{red}{$-$48.0})
& 2.0 (\textcolor{red}{$-$18.8})
& 16.4 (\textcolor{red}{$-$28.0})
& 42.0 (\textcolor{red}{$-$10.8})
& 17.3 (\textcolor{red}{$-$26.4}) \\

GLM-5.3-Flash
& 7.3 (\textcolor{red}{$-$40.0})
& 1.9 (\textcolor{red}{$-$17.0})
& 12.3 (\textcolor{red}{$-$11.2})
& 29.6 (\textcolor{red}{$-$4.4})
& 12.8 (\textcolor{red}{$-$18.1}) \\

DeepSeek-V4-Flash-Vision
& 7.3 (\textcolor{red}{$-$28.7})
& 3.3 (\textcolor{red}{$-$9.7})
& 11.6 (\textcolor{red}{$-$15.5})
& 28.5 (\textcolor{red}{$-$13.3})
& 12.7 (\textcolor{red}{$-$16.8}) \\

\midrule

Qwen3.5-0.8B
& 0.0 ($+$0.0)
& 1.0 (\textcolor{green}{$+$0.1})
& 1.8 (\textcolor{green}{$+$0.1})
& 15.1 (\textcolor{green}{$+$1.3})
& 4.5 (\textcolor{green}{$+$0.4}) \\

Qwen3.5-2B
& 0.0 ($+$0.0)
& 1.5 (\textcolor{green}{$+$0.6})
& 1.7 (\textcolor{red}{$-$0.1})
& 16.3 (\textcolor{red}{$-$3.5})
& 4.9 (\textcolor{red}{$-$0.7}) \\

Qwen3.5-9B
& 3.3 ($+$0.0)
& 3.2 (\textcolor{green}{$+$0.4})
& 4.1 (\textcolor{red}{$-$1.1})
& 27.7 (\textcolor{red}{$-$1.5})
& 9.6 (\textcolor{red}{$-$0.5}) \\

Qwen3.5-27B
& 4.0 (\textcolor{red}{$-$5.3})
& 2.9 (\textcolor{red}{$-$2.7})
& 7.6 (\textcolor{red}{$-$2.1})
& 25.2 (\textcolor{red}{$-$4.1})
& 9.9 (\textcolor{red}{$-$3.6}) \\

\midrule

\ourmodel-8B-CPT
& 47.7 (\textcolor{red}{$-$13.0})
& 14.4 (\textcolor{red}{$-$2.2})
& 21.9 (\textcolor{red}{$-$3.0})
& 34.0 (\textcolor{green}{$+$0.7})
&  29.5 (\textcolor{red}{$-$4.4})\\

\ourmodel-35B-A3B-CPT
& 65.3 (\textcolor{green}{$+$18.6})
& 11.9 (\textcolor{green}{$+$1.7})
& 18.7 (\textcolor{red}{$-$6.7})
& 30.2 (\textcolor{red}{$-$0.1})
&  31.5 (\textcolor{green}{$+$3.4}) \\

\bottomrule
\end{tabular}
}
\end{table*}

\textbf{CPT supports effective cross-game control across Dense and MoE backbones.}
The Qwen3-VL-8B CPT policy achieves an average real-time score of 29.5, while the Qwen3.5-35B-A3B MoE CPT policy reaches 31.5. Both exceed the strongest listed general-purpose baseline. The difference is especially clear on MCU, where the two CPT policies score 47.7 and 65.3, respectively. Their results also show complementary strengths across games. The 8B Dense policy achieves 21.9 on OmniGameArena and 34.0 on GameWorld, while the MoE policy performs particularly strongly on MCU. These results extend the scaling analysis from action prediction and validation loss to executable gameplay, and show that the CPT approach is applicable to both Dense and MoE model architectures.

\textbf{Real-time evaluation provides a practical measure of control ability.} The paired RT and TP results further highlight the importance of real-time inference for interactive game control. While most general-purpose models experience substantial performance degradation when the game continues advancing during inference, our 8B CPT policy shows only a modest decrease, demonstrating its ability to maintain effective control under real-time constraints. Notably, the MoE CPT policy achieves an average RT--TP improvement of +3.4 points. This positive gain is consistent with its faster inference, which enables more timely interaction with the evolving environment and reduces the accumulation of control errors. These results suggest that low-latency inference is important not only for preserving gameplay performance but also for fully realizing the capabilities learned through CPT. Further details and results of the 120-configuration GameScaling study are provided in \citet{gamescaling}. Table~\ref{tab:cpt_game_benchmarks} complements the scaling analysis with direct evidence of executable cross-game control. We further investigate the inference optimizations enabling real-time interaction in Sec.~\ref{sec:exp_streaming}.

\begin{table*}[!ht]
\centering
\caption{
\textbf{Long-horizon instruction following across two side quests.}
To improve evaluation robustness, we evaluate each task three times and report the averaged success rate (\%) of evaluated tasks for each interaction type.
Overall accuracy (Acc) divides successful task executions by all evaluated task
executions, pooling repeated attempts over the 30 or 55 milestone units.
Opportunity Makes the Thief (Part I) and Bestial Ferocity contain
30 and 55 milestones, respectively.
}
\label{tab:sidequest_results}

\setlength{\tabcolsep}{4pt}
\renewcommand{\arraystretch}{1.12}

\resizebox{0.92\linewidth}{!}{%
\begin{tabular}{l*{11}{c}}
\toprule

\multirow{2}{*}{\textbf{Model}}
& \multicolumn{5}{c}{\textbf{Opportunity Makes the Thief (I)}}
& \multicolumn{6}{c}{\textbf{Bestial Ferocity}} \\

\cmidrule(lr){2-6}
\cmidrule(lr){7-12}

& Nav.
& Dialogue
& Combat
& UI
& \textbf{Acc $\uparrow$}
& Nav.
& Dialogue
& Combat
& UI
& Puzzle
& \textbf{Acc $\uparrow$} \\

\midrule

Gemini-3.8-Flash
& 0.0 & 35.7 & 0.0 & 0.0 & 16.7
& 31.2 & 81.2 & 50.0 & 50.0 & 0.0 & 45.5 \\

GPT-5.6-Sol
& 7.1 & 28.6 & 0.0 & 0.0 & 16.7
& 28.1 & 68.8 & 50.0 & 100.0 & 0.0 & 41.8 \\

GPT-5.6-Luna
& 7.1 & 14.3 & 0.0 & 0.0 & 10.0
& 18.8 & 43.8 & 0.0 & 0.0 & 0.0 & 23.6 \\

Claude-Opus-5.0
& 7.1 & 21.4 & 0.0 & 0.0 & 13.3
& 15.6 & 56.2 & 50.0 & 50.0 & 0.0 & 29.1 \\

GLM-5.3-Flash
& 7.1 & 21.4 & 0.0 & 0.0 & 13.3
& 15.6 & 43.8 & 50.0 & 0.0 & 0.0 & 23.6 \\

DeepSeek-V4-Flash-Vision
& 14.3 & 35.7 & 0.0 & 0.0
& 23.3
& 15.6 & 56.2 & 50.0 & 0.0 & 0.0 & 27.3 \\

\midrule

Qwen3-VL-8B (Base)
& 0.0 & 0.0 & 0.0 & 0.0 & 0.0
& 12.5 & 18.8 & 33.3 & 0.0 & 0.0 & 13.9 \\

\ourmodel-CPT
& 14.3 & 50.0 & 100.0 & 33.3 & 34.4
& 36.5 & 70.8 & 100.0 & 66.7 & 0.0 & 47.9 \\

\ourmodel-SFT
& 47.6 & \textbf{92.9} & 100.0 & 100.0 & 72.2
& 55.2 & 97.9 & 100.0 & 100.0 & \textbf{22.2} & 69.1 \\

\textbf{\ourmodel-RL}
& \textbf{69.1} & 88.1 & \textbf{100.0} & \textbf{100.0} & \textbf{80.0}
& \textbf{80.2} & \textbf{100.0} & \textbf{100.0} & \textbf{100.0} & 11.1 & \textbf{83.6} \\

\bottomrule
\end{tabular}%
}
\end{table*}

\subsection{Long-Horizon Instruction Following in Side Quests}
\label{sec:exp_sft}
\label{sec:exp_rl}

The two side quests require the agent to carry out multi-stage tasks while retaining information from earlier interactions. Table~\ref{tab:sidequest_results} reports verified task-execution success rates on \ourmodel.

\textbf{SFT substantially improves long-horizon task execution.}
On \emph{Opportunity Makes the Thief (Part I)}, overall accuracy increases from 0.0\% for Qwen3-VL-8B to 34.4\% after HSR-augmented CPT and 72.2\% after SFT. On \emph{Bestial Ferocity}, the corresponding scores are 13.9\%, 47.9\%, and 69.1\%. Thus, SFT adds 37.8 and 21.2 percentage points over the respective CPT checkpoints. The improvement is visible across distinct interaction types. For example, navigation rises from 14.3\% to 47.6\% on the first quest, while dialogue increases from 70.8\% to 97.9\% on the second. These outcomes are consistent with the objective of multi-level instruction tuning, which trains the same action policy to follow goals that span local operations, intermediate subgoals, and longer task sequences.

Both quests involve repeated scene transitions, UI interactions, and changes in objectives. To make progress, the agent must retain relevant information from earlier interactions and use it to decide what to do next. SFT improves performance on held-out tasks, suggesting that task-conditioned reasoning and memory can support this behavior in a familiar game, even when the specific evaluation objectives were excluded from training.

\textbf{Online RL further strengthens long-horizon execution.}
We perform online RL on the basis of the SFT checkpoint. Relative to its SFT initialization, TA-GRPO improves overall task-execution accuracy from 72.2\% to 80.0\% on \emph{Opportunity Makes the Thief (Part I)} and from 69.1\% to 83.6\% on \emph{Bestial Ferocity}. These correspond to gains of \textbf{7.8} and \textbf{14.5 percentage points}, respectively. Navigation accuracy rises from 47.6\% to 69.1\% on the first quest and from 55.2\% to 80.2\% on the second. On \emph{Bestial Ferocity}, dialogue reaches 100\% after RL. These results suggest that verified interaction feedback helps refine the policy's ability to advance toward task goals over extended trajectories.

The gains across both side quests highlight the complementary roles of instruction supervision and online policy optimization. SFT provides task-conditioned decisions and memory, while RL allows the policy to learn from the consequences of its own actions under game-native verification. Importantly, these improvements are obtained without changing the underlying action interface or model architecture. Appendix~\ref{app:rl_rollouts} provides representative training and held-out development rollouts that illustrate how navigation and final-target interactions evolve across checkpoints.

\begin{table*}[t]
\centering
\caption{
\textbf{Instruction generalization across seven minigame types.}
To improve evaluation robustness, we evaluate each task three times and report the averaged task success rates (\%) on a reduced benchmark of 40 task instances.
Each task score is the mean of its valid evaluation attempts.
Acc is the unweighted average across the seven game types.
ML, NC, AC, HX, JHP, GS, and JM denote Magflow Link, Navigation Compass,
Abacus Circuitry, Hexanexus, Janus's Hidden Passage, Golden Scapegoat,
and Janus's Maze, respectively.
}
\label{tab:minigame_results}
\vspace{4pt}
\setlength{\tabcolsep}{6pt}
\renewcommand{\arraystretch}{1.12}
\resizebox{0.75\linewidth}{!}{%
\begin{tabular}{l*{8}{c}}
\toprule
\multirow{2}{*}{\textbf{Model}} & \multicolumn{7}{c}{\textbf{Minigame Types}} & \multirow{2}{*}{\textbf{Acc $\uparrow$}} \\
\cmidrule(lr){2-8}
& ML & NC & AC & HX & JHP & GS & JM & \\
\midrule
GPT-6.1-Sol & 20.0 & 37.5 & 0.0 & 33.3 & 14.3 & 0.0 & 14.3 & 17.1 \\
Gemini-3.8-Flash & 0.0 & 0.0 & 0.0 & 0.0 & 0.0 & 0.0 & 14.3 & 2.0 \\
GPT-5.6-Sol & 0.0 & 12.5 & 0.0 & 0.0 & 57.1 & 0.0 & 14.3 & 12.0 \\
GPT-5.6-Luna & 0.0 & 37.5 & 0.0 & 0.0 & 28.6 & 0.0 & 14.3 & 11.5 \\
Claude-Opus-5.0 & 0.0 & 12.5 & 0.0 & 0.0 & 57.1 & 0.0 & 0.0 & 9.9 \\
GLM-5.3-Flash & 0.0 & 0.0 & 0.0 & 0.0 & 57.1 & 0.0 & 0.0 & 8.2 \\
DeepSeek-V4-Flash-Vision & 0.0 & 0.0 & 0.0 & 0.0 & 0.0 & 0.0 & 14.3 & 2.0 \\
\midrule
Qwen3-VL-8B (Base)
& 0.0 & 8.3 & 0.0 & 0.0
& 0.0 & 4.8 & 9.5 & 3.2 \\

\ourmodel-CPT
& 20.0 & 37.5 & 0.0 & 0.0
& 23.8 & 9.5 & 14.3 & 15.0 \\

\ourmodel-SFT
& 40.0 & 50.0
& 11.1 & 11.1
& 76.2 & 33.3
& \textbf{52.4} & 39.2 \\

\textbf{\ourmodel-RL}
& \textbf{53.3} & \textbf{95.8}
& \textbf{22.2} & \textbf{33.3}
& \textbf{81.0} & \textbf{38.1}
& 23.8 & \textbf{49.7} \\
\bottomrule
\end{tabular}%
}
\end{table*}

\subsection{Instruction Generalization across Diverse Minigames}
\label{sec:exp_minigames}

The minigame suite emphasizes instruction following across different gameplay mechanics. This broader coverage is important because the tasks require different combinations of spatial reasoning, rule interpretation, visual matching, and device-level interaction rather than a single repeated control routine. Table~\ref{tab:minigame_results} reports success for each of the seven types and their unweighted macro-average.

\textbf{Instruction tuning broadens success across distinct minigame mechanics.}
The seven-type macro-average increases from 3.2\% for the base Qwen3-VL-8B to 15.0\% for \ourmodel-CPT and 39.2\% for \ourmodel-SFT, a further gain of 24.2 percentage points after instruction tuning. SFT reaches 76.2\% on Janus's Hidden Passage, 52.4\% on Janus's Maze, and 50.0\% on Navigation Compass. It also obtains non-zero mean success in all seven types, compared with three for the base model. The CPT checkpoint already improves the macro-average, providing a useful foundation for responding to gameplay observations and actions. SFT then connects these control capabilities to task-specific goals, which is reflected in improvements across puzzles with different interfaces and completion rules. The evaluation tasks are held out from training, so the comparison measures generalization to new instances and objectives within HSR.

\textbf{Online RL further improves performance across diverse mechanics.}
Following SFT, \ourmodel-RL raises the seven-type macro-average from 39.2\% to \textbf{49.7\%}, an additional gain of \textbf{10.5 percentage points}. The improvement is particularly pronounced on Navigation Compass, where accuracy increases from 50.0\% to 95.8\%. RL also improves Magflow Link from 40.0\% to 53.3\%, Hexanexus from 11.1\% to 33.3\% and Janus’s Hidden Passage from 76.2\% to 81.0\%, with additional gains on Abacus Circuitry and Golden Scapegoat. These results extend the benefits of online policy optimization beyond long-horizon quests to tasks involving diverse rules and interaction patterns. Together, the CPT, SFT, and RL stages progressively improve the model's ability to interpret instructions and execute actions across different game mechanics.

To prevent game types with more tasks from dominating the aggregate, the overall metric first averages attempts within each task and then gives each of the seven types equal weight. This makes the reported improvement a measure of breadth across mechanics rather than only the most frequent minigame.

\subsection{Impact of Real-Time Rollouts on Online RL}
\label{sec:exp_realtime_rl}

\textbf{Real-time interaction is critical for online RL in modern games.}
Unlike lightweight simulators, modern games involve continuously evolving environments, complex gameplay mechanics, and state-dependent interactions. In this setting, inference latency affects not only the responsiveness of the deployed agent but also the quality and efficiency of online policy learning. Delayed control can cause subsequent actions to operate on outdated observations, making it harder to obtain timely feedback from the environment. We therefore examine the effect of rollout frequency on RL training, comparing real-time Streaming-vLLM against standard vLLM rollouts. The game continues to advance under both settings.

\begin{figure*}[!htbp]
    \centering
    \includegraphics[width=0.85\linewidth]{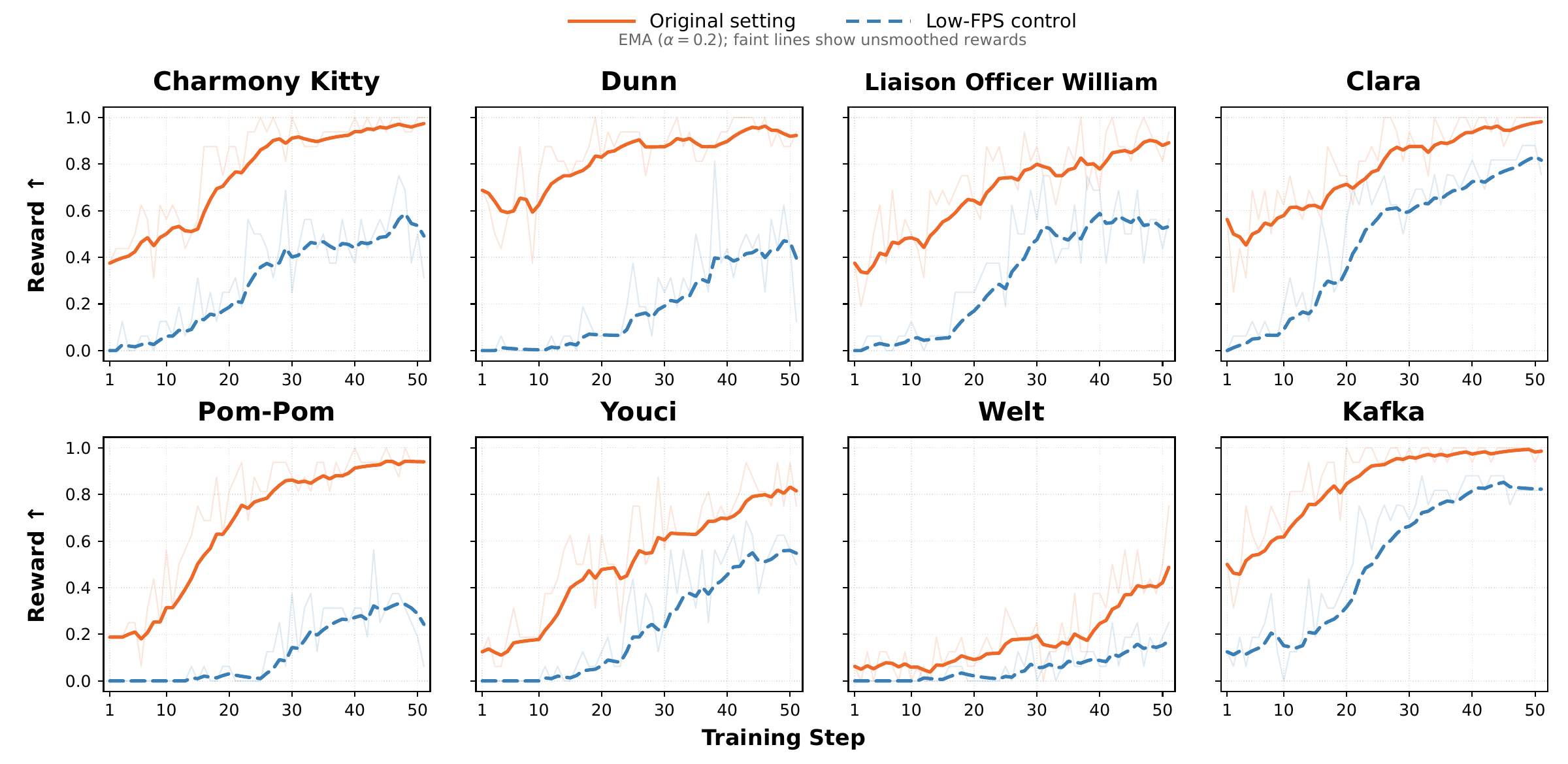}
    \caption{\textbf{Real-time inference improves online RL learning across eight navigation tasks.} Task labels indicate navigation targets. For example, ``Pom-Pom'' refers to the task of finding the NPC Pom-Pom. Orange solid lines show rewards under the original real-time Streaming-vLLM setting, while blue dashed lines show the 0.6-FPS control representing standard vLLM rollout speed. Real-time inference typically achieves earlier reward improvement and stronger learning over the first 50 plotted training iterations.}
    \label{fig:rl_reward}
\end{figure*}

\textbf{Timely control supports more effective online policy learning.}
As shown in Figure~\ref{fig:rl_reward}, real-time Streaming-vLLM consistently demonstrates more favorable learning dynamics than the standard vLLM setting across eight navigation tasks. First, real-time rollouts achieve higher rewards from the early stages of training. Timely observation and action execution reduce accumulated interaction errors, allowing the agent to reach intended states more reliably and obtain informative reward feedback more efficiently. Second, real-time training achieves higher reward levels, particularly on tasks involving time-sensitive navigation and interactions, where delayed control can prevent the agent from completing the required action sequences. More importantly, several low-FPS runs exhibit reward collapse in the later stages of training. This behavior suggests that accumulated execution errors can introduce increasingly biased trajectories into online optimization, undermining subsequent policy updates and training stability. In contrast, real-time interaction maintains more reliable environment feedback and supports sustained policy improvement. These observations highlight that low-latency inference is not merely a deployment optimization but an important prerequisite for effective and stable RL training in complex commercial games.

\subsection{Streaming-vLLM}
\label{sec:exp_streaming}

\paragraph{Streaming video benchmarks.}
In addition to low latency, an inference engine must preserve the model's ability to understand and respond to continuously arriving visual input.
To verify the inference accuracy of our Streaming-vLLM, we further evaluate streaming inference on StreamingBench~\cite{lin2026streamingbench} and OVO-Bench~\cite{niu2025ovo}.
For StreamingBench, we use the Real-Time Visual Understanding subset.
OVO-Bench evaluates current-event understanding through \textbf{Real}, retrieval of earlier information through \textbf{Backward}, and delayed response after sufficient evidence through \textbf{Forward}.

We first test whether stateful streaming preserves online visual understanding.
The controlled comparison uses the same backbone checkpoint, hardware, numerical precision, decoding configuration, and cache budget across inference methods. We compare Streaming-vLLM against DSCache~\cite{pang2026dscache}, InfiniPot-V~\cite{kim2025infinipotv}, and a uniform streaming-cache baseline corresponding to the caching strategy used by StreamingVLM~\cite{xu2026streamingvlm}.
The evaluation reports StreamingBench Real-Time accuracy, OVO-Bench Overall and its Real, Backward, and Forward subsets, query latency, and peak GPU memory.

\begin{table*}[t]
    \centering
    \caption{
        \textbf{Accuracy, latency, and memory consumption of streaming video understanding.}
        StreamingBench Real-Time and OVO-Bench scores are reported with query latency and peak GPU memory. Latency and memory consumption are measured on an NVIDIA H200.
    }
    \label{tab:streaming_video_results}

    \resizebox{0.95\linewidth}{!}{%
        \begin{tabular}{llcccc}
            \toprule
            \multirow{2}{*}{\textbf{Method}}
            & \multirow{2}{*}{\textbf{Backbone}}
            & \textbf{StreamingBench$\uparrow$}
            & \textbf{OVO-Bench$\uparrow$}
            & \multirow{2}{*}{\textbf{Latency$\downarrow$}}
            & \multirow{2}{*}{\textbf{Memory}} \\
            & & \textbf{(Real-time)} & \textbf{(Real / Backward / Forward)} & & \\
            \midrule
            
            ReKV~\cite{di2025rekv}
            & \multirow{3}{*}{LLaVA-OV-7B}
            & 71.06
            & 53.9 (62.0 / 48.5 / 51.2)
            & 0.23s
            & 21 GB \\

            StreamingVLM~\cite{xu2026streamingvlm}
            &
            & 76.92
            & 55.6 (68.4 / 46.0 / 52.9)
            & 0.21s
            & 16 GB \\

            DSCache~\cite{pang2026dscache}
            & 
            & 79.12
            & 57.5 (71.5 / 47.8 / 53.1)
            & 0.30s
            & 16 GB \\

            \midrule

            InfiniPot-V~\cite{kim2025infinipotv}
            & \multirow{3}{*}{Qwen2.5-VL-7B}
            & 76.40
            & 53.6 (65.9 / 47.6 / 47.9)
            & 0.37s
            & 17 GB \\

            StreamingVLM~\cite{xu2026streamingvlm}
            &
            & 78.56
            & 52.2 (67.8 / 42.3 / 46.6)
            & 0.31s
            & 17 GB \\

            DSCache~\cite{pang2026dscache}
            & 
            & 82.32
            & 55.4 (71.7 / 44.8 / 49.7)
            & 0.49s
            & 17 GB \\

            \midrule

            DSCache~\cite{pang2026dscache}
            & \multirow{2}{*}{Qwen3-VL-8B}
            & 83.21
            & 58.2 (78.8 / 46.2 / 49.5)
            & 0.53s
            & 21 GB \\

            \textbf{Streaming-vLLM}
            &
            & \textbf{83.29}
            & \textbf{62.3} (\textbf{80.4} / \textbf{54.2} / \textbf{52.3})
            & \textbf{0.08s}
            & 22 GB \\

            \midrule

            DSCache~\cite{pang2026dscache}
            & \multirow{2}{*}{Qwen3.5-9B}
            & 80.56
            & 56.3 (74.8 / 50.9 / 43.2)
            & 1.56s
            & 21 GB \\

            \textbf{Streaming-vLLM}
            &
            & \textbf{83.77}
            & \textbf{60.9} (\textbf{78.6} / \textbf{57.6} / \textbf{46.3})
            & \textbf{0.16s}
            & 22 GB \\
            
            \bottomrule
        \end{tabular}%
    }
\end{table*}

\textbf{Streaming-vLLM preserves visual understanding while reducing inference latency.}
Table~\ref{tab:streaming_video_results} presents matched-backbone comparisons on two complementary streaming benchmarks. With Qwen3-VL-8B, Streaming-vLLM scores 83.29 on StreamingBench and 62.3 on OVO-Bench, reducing query latency from 0.53 to 0.08~s relative to DSCache. With Qwen3.5-9B, it reaches 83.77 and 60.9, respectively, while latency falls from 1.56 to 0.16~s. The OVO-Bench improvements appear in its Real, Backward, and Forward subsets, suggesting that the retained context remains useful for both incoming visual evidence and information from earlier video frames.

The results are particularly relevant to interactive agents, as reducing latency is only useful if the system can still interpret the observations needed for its next decision. These comparisons show that the serving optimizations do not require sacrificing the measured streaming-video accuracy on either backbone. They also establish a practical connection between efficient context reuse and the real-time policy inference used in our game experiments. Peak memory remains in the same general range as the matched-backbone baselines, with 22~GB reported for Streaming-vLLM. For OVO-Bench Forward, the latency measure excludes the time spent waiting for sufficient visual evidence, separating model computation from the benchmark's delayed-answer protocol.

\subsubsection{Inference Efficiency}

We next evaluate Streaming-vLLM on the actual keyboard-mouse interaction workload used by the Verl-based training and evaluation framework. This is the inference path through which policy actions are generated for online TA-GRPO rollouts as well as real-time testing. Table~\ref{tab:streaming_vllm_ablation} summarizes a controlled step-by-step engineering ablation on a fixed short-horizon interaction sequence on Hopper using the Qwen3.5-35B-A3B MoE model. To evaluate transferability of the serving design beyond this ablation benchmark, Appendix~\ref{app:streaming_latency} separately conducts a multi-turn cross-hardware sweep comparing the Qwen3-VL-8B Dense model and Qwen3.5-35B-A3B MoE model across datacenter Ampere and Hopper GPUs, multiple tensor-parallel degrees, and BF16 or W8A8 precision.

The cumulative configurations cover stateful serving, tensor parallelism, operator fusion, parallel overlap, CUDA Graphs, a low-overhead execution backend, quantization, and FSM-guided decoding with BF16 target-model verification.

\begin{table*}[t]
    \centering
    \small
    \caption{
    \textbf{Cumulative Streaming-vLLM optimization on NVIDIA H200 (Hopper) with Qwen3.5-35B-A3B MoE.} Latencies are in milliseconds. SR is the step rate, and CR is the theoretical six-block control rate derived from SR. The change from TP1 to TP4 incurs communication and synchronization overhead, so per-component speedups need not be monotonic.
    }
    \label{tab:streaming_vllm_ablation}
    \setlength{\tabcolsep}{5pt}
    \renewcommand{\arraystretch}{1.12}
    \resizebox{0.92\linewidth}{!}{%
    \begin{tabular}{lcccccc}
        \toprule
        \textbf{Configuration}
        & \textbf{ViT $\downarrow$}
        & \textbf{Prefill $\downarrow$}
        & \textbf{Decode $\downarrow$}
        & \textbf{Latency (ms) $\downarrow$}
        & \textbf{SR (Hz) $\uparrow$}
        & \textbf{CR (Hz) $\uparrow$} \\
        \midrule
        Baseline (TP1)
        & 27.90 & 94.70 & 1087.00 & 1306.00 & 0.77 & 4.62 \\
        \midrule
        + Stateful Streaming Engine (TP4)
        & 68.89 & 121.85 & 874.07 & 1264.76 & 0.79 & 4.74 \\
        + Basic Fusion
        & 60.67 & 154.80 & 630.51 & 1042.76 & 0.96 & 5.76 \\
        + Parallel Overlap
        & 55.74 & 144.48 & 507.37 & 831.82 & 1.20 & 7.20 \\
        + CUDA Graph
        & 64.06 & 140.31 & 179.79 & 536.66 & 1.86 & 11.16 \\
        \midrule
        + Megakernel
        & 16.61 & 24.19 & 35.97 & 87.68 & 11.41 & 68.46 \\
        + Partial W8A8 Quantization
        & 16.38 & 23.29 & 35.23 & 85.87 & 11.65 & 69.90 \\
        \textbf{+ Guided Decoding}
        & \textbf{16.20} & \textbf{23.06} & \textbf{15.27}
        & \textbf{65.12} & \textbf{15.36} & \textbf{92.16} \\
        \bottomrule
    \end{tabular}%
    }
    \vspace{2pt}
    \begin{minipage}{\linewidth}
    \end{minipage}
\end{table*}

\textbf{End-to-end performance depends on optimizing the full interaction path.}
Table~\ref{tab:streaming_vllm_ablation} traces successive runtime configurations for the Qwen3.5-35B-A3B MoE model on H200. The initial TP4 streaming configuration offers only a modest change in total latency relative to the TP1 baseline. Vision processing, prefill, and autoregressive decoding impose different computational and communication costs, so accelerating one component does not necessarily reduce the complete request time by the same proportion. This motivates the use of architecture-aware kernels and overlapping execution rather than relying on tensor parallelism alone. Because some configurations also change profiling boundaries, the table reports cumulative system improvements rather than isolated single-factor effects.

\textbf{Cumulative runtime optimizations yield a substantial latency reduction.}
Operator fusion, parallel overlap, CUDA Graph replay, and the low-overhead megakernel backend progressively reduce request latency. FSM-guided drafting further shortens action generation, while BF16 target-model verification preserves the target policy's acceptance decision. Together, these cumulative optimizations reduce end-to-end latency from approximately 1.31~s to 65~ms, which is a \textbf{20.1$\times$ speedup} on the measured H200 MoE configuration. The largest gains emerge when the runtime addresses vision processing, context handling, and decoding as a single interactive workload. Appendix~\ref{app:streaming_latency} extends the system evaluation to representative Dense and MoE configurations on A800 and H200 GPUs.

The latency reduction also increases the capacity to produce executable action packets within a fixed wall-clock interval. Table~\ref{tab:streaming_vllm_ablation} reports the corresponding step rate (SR) and a theoretical control rate (CR) obtained by multiplying SR by the six action blocks per packet. These are throughput-derived capacities, not direct measurements of the client's realized game frame rate, which follows its own execution schedule. This distinction allows the system results to quantify available inference headroom without conflating request throughput with observed gameplay cadence.

\section{Conclusion}

\ourmodel learns to control diverse and complex games through a shared keyboard-and-mouse interface. It combines large-scale continued pre-training, task-conditioned supervised fine-tuning, and temporal-aware reinforcement learning to learn transferable control, long-horizon reasoning, and persistent memory. Streaming-vLLM supports both training and deployment, enabling real-time interaction and efficient online rollouts while reducing latency-induced bias in reinforcement learning. At inference time, \ourmodel acts through visual observations and device-level controls without privileged game state or game-engine modifications. Environment-side task verification is used separately to provide online RL rewards. These capabilities could effectively reduce development costs through automated testing, faster game iteration, and new gameplay mechanics. By supporting real-time interaction without game-specific infrastructure, \ourmodel offers a basis for agents that learn and improve across complex 3D games and other interactive environments.

\section{Limitations and Future Work}

Several main directions remain for extending Gamer’s scope and accessibility. First, our evaluation covers a selected set of environments and task structures. The shared observation-action interface provides a basis for extending joint post-training and systematic evaluation to broader interactive settings. Our preliminary results suggest the potential for improving performance and generalization by scaling relevant gameplay data across a broader range of environments. These benefits depend on the coverage and composition of the training experience. 

Second, Streaming-vLLM enables real-time interaction on the evaluated GPU platforms, while further reductions in serving costs would support wider deployment. Model compression and additional runtime optimizations offer promising directions for extending this efficiency to more resource-constrained settings while preserving responsiveness and task performance.

Third, although Gamer improves the trade-off between response latency and task performance in the evaluated settings, reasoning capability may remain a bottleneck in environments that demand deeper deliberation. Fast, continuous interaction alone does not ensure the ability to solve complex strategic problems. Our framework provides a foundation for exploring hierarchical game agents that combine a larger model for deliberative System 2 reasoning with a fast model for reactive System 1 execution. Such architectures could coordinate longer-term planning with timely control, extending the range of tasks that agents can handle under real-time constraints.

\renewcommand{\footnoterule}{}

\makeatletter
\begingroup
\renewcommand{\thefootnote}{}
\long\def\@makefntext#1{%
  \parindent=0pt
  \noindent#1%
}
\footnotetext{%
  \textsuperscript{$\dagger$}
  Work done while interning at miHoYo.
}
\endgroup
\makeatother

\section*{Contributions}
\label{contribution}
All core contributors contributed equally to this work.

\subsection*{Core Contributors}
Run Luo$^{1,2,\dagger}$,
Kuan Zhang$^{1,3,\dagger}$,
Qi Deng$^{1}$,
Yukun Chen$^{1,4,\dagger}$,
Zhihao Yang$^{1,4,\dagger}$,
Yufei Zhang$^{1}$,
Renrong Hu$^{1}$,
Yi Zhang$^{1,5,\dagger}$

\subsection*{Contributors}
Guiyu Ma$^{6,\dagger}$,
Tianyu Fan$^{1}$,
Longze Chen$^{4,\dagger}$,
Hao Sun$^{3}$,
Jiaming Li$^{4,\dagger}$,
Jiaming Yang$^{1}$,
Xiaohao Liu$^{2}$,
Tao Tan$^{1}$,
Minzheng Wang$^{4,\dagger}$,
Hangtian Jia$^{1}$,
Shuai Yuan$^{1}$,
Lvdi Wang$^{1}$,
Xiao Liu$^{1}$,
Chao Ma$^{1}$,
Yuting Zou$^{1}$

\subsection*{Supervisors}
Yinhe Zheng$^1$, Xiaobo Xia$^5$, Dawei Jiang$^1$, Shaoqing Ren$^5$

\subsection*{Affiliations}
$^1$miHoYo Honkai AI R\&D Team,
$^2$National University of Singapore,
$^3$Tsinghua University,
$^4$University of Chinese Academy of Sciences,
$^5$University of Science and Technology of China,
$^6$Xi’an Jiaotong University

\subsection*{Core Contribution Details}
\subsubsection*{Continual Pre-Training}
Run Luo, Kuan Zhang
\vspace{-5pt}
\subsubsection*{Supervised Fine-Tuning}
Run Luo, Yukun Chen
\vspace{-5pt}
\subsubsection*{Reinforcement Learning}
Run Luo, Qi Deng, Zhihao Yang
\vspace{-5pt}
\subsubsection*{Data Collection and Construction}
Run Luo, Kuan Zhang, Qi Deng, Yukun Chen, Zhihao Yang
\vspace{-5pt}
\subsubsection*{Training and Inference Infrastructure}
Run Luo, Qi Deng, Yufei Zhang
\vspace{-5pt}
\subsubsection*{Benchmarking and Evaluation}
Run Luo, Kuan Zhang, Qi Deng, Yukun Chen, Zhihao Yang, Renrong Hu, Yi Zhang

\subsection*{Acknowledgment}
We thank all members of the Gamer project for their dedication and collaboration throughout this work. We also thank our colleagues, including Kongyang Liu, Wangqin Fu, Qi Qin, Yupeng Zhu, Rui Dong, Haiyun Gu, Xiyu Zhao, Ran Fang, and Ange Tang, for their valuable support. We are grateful to teams across miHoYo for the technical and organizational support that made this research possible.

\bibliographystyle{plainnat}
\bibliography{main}

\clearpage

\renewcommand{\cftbeforesecskip}{16pt}      
\renewcommand{\cftbeforesubsecskip}{12pt}

\renewcommand{\contentsname}{Appendix}
\addtocontents{toc}{\protect\setcounter{tocdepth}{2}}
\appendix
{\small
\tableofcontents
}
\clearpage

\section{Related Work}

\subsection{Generalist Game Agents}

Research on game-playing agents has expanded from environment-specific policies to general models combining perception, reasoning, memory, and executable control. Large-scale behavioral pretraining provides one route: VPT~\cite{baker2022vpt} learns Minecraft control from internet videos labeled by an inverse dynamics model, while STEVE-1~\cite{lifshitz2023steve1} extends this paradigm to language-conditioned behavior. Other systems, including DEPS~\cite{wang2023deps}, Voyager~\cite{wang2023voyager}, JARVIS-1~\cite{wang2025jarvis1}, Optimus-1~\cite{li2024optimus1}, and OmniJARVIS~\cite{wang2024omnijarvis}, incorporate planning, reusable skills, and memory to support increasingly complex tasks. 

Recent work further targets generalization across tasks and worlds. Gato~\cite{reed2022gato} and SIMA~\cite{simateam2024sima} study shared policies across heterogeneous environments, while Game-TARS~\cite{wang2025gametars} and Lumine~\cite{tan2025lumine} combine reasoning, memory, and native keyboard–mouse control for extended gameplay. NitroGen~\cite{magne2026nitrogen} and OpenP2P ~\cite{yue2026openp2p} instead emphasize scalable visual policy learning from broad gameplay data. These approaches address different aspects of generalist gameplay, typically emphasizing cross-environment reactive control or extended reasoning under more specialized training settings. We combine cross-game continued pre-training with task-oriented policy optimization for sustained interaction.

\subsection{Scalable Gameplay Learning}

Progress in game agents increasingly relies on large and heterogeneous interaction data. MineRL~\cite{guss2019minerl} and MineDojo~\cite{fan2022minedojo} provide structured demonstrations and multimodal resources, while VPT~\cite{baker2022vpt} scales behavioral learning through inverse dynamics. D2E~\cite{choi2026d2e}, NitroGen~\cite{magne2026nitrogen}, and OpenP2P~\cite{yue2026openp2p} broaden supervision through desktop trajectories, visible control overlays, and inverse-dynamics-labeled gameplay. GameHorizon~\cite{wang2026gamehorizon} further introduces multi-level annotations spanning operations, subgoals, and strategies. 

Transfer also depends on model size and data composition. General scaling and continual-pretraining studies show that performance depends jointly on model capacity, data budget, and domain mixture. In gameplay, VPT~\cite{baker2022vpt} demonstrates the value of large-scale behavioral pretraining, while OpenP2P~\cite{yue2026openp2p} studies scaling through both predictive performance and causal behavior. GameScaling varies model size, interaction budget, and game composition to distinguish between gains obtained by learning a shared control interface and gains that depend on target-relevant gameplay. Scale alone is not sufficient for task-oriented gameplay. Effective policy learning also requires supervision that connects local actions with higher-level objectives. GameHorizon~\cite{wang2026gamehorizon} introduces temporal hierarchy into gameplay supervision, while OmniJARVIS~\cite{wang2024omnijarvis} and Lumine~\cite{tan2025lumine} combine reasoning, memory, and staged training. We train the policy with supervision at the operation, subgoal, and task levels, and retain compact task summaries across context boundaries.

\subsection{Memory and Long-Horizon Interaction}

Sustained gameplay also requires agents to retain task-relevant information after earlier interactions leave the active context. Existing agents address this through reusable skills, retrieved experience, or compressed interaction histories. Voyager stores executable skills, JARVIS-1~\cite{wang2025jarvis1} and Optimus-1~\cite{li2024optimus1} retrieve accumulated experience, and Game-TARS~\cite{wang2025gametars} combines recent interactions with compressed textual history. OmniJARVIS~\cite{wang2024omnijarvis} and Lumine~\cite{tan2025lumine} further integrate memory with reasoning and action generation. We explicitly supervise task summaries and use them as persistent task state during supervised training, online policy optimization, and streaming inference. 

Feedback-based optimization provides a complementary mechanism for improving extended behavior. GRPO~\cite{shao2024deepseekmath} optimizes relative advantages across groups of sampled outputs, while GUI-R1~\cite{luo2025gui} and Game-RL~\cite{tong2026gamerl} adapt reinforcement learning to executable GUI actions and multimodal game reasoning. Game-TARS~\cite{wang2025gametars} also refines thought–action trajectories through action-consistency-based filtering. We optimize finite trajectory segments linked by compressed memory, updating task-state maintenance and executable behavior together. Long-horizon interaction must also operate under bounded context and latency. StreamingLLM~\cite{xiao2024streamingllm} maintains stable inference with a fixed attention window, while InfiniPot-V~\cite{kim2025infinipotv}, DSCache~\cite{pang2026dscache}, and StreamingVLM~\cite{xu2026streamingvlm} explore compressed or structured memory for continuous visual streams. Serving techniques such as PagedAttention~\cite{kwon2023pagedattention} and speculative decoding further reduce autoregressive inference cost. These directions are largely studied separately from long-horizon policy optimization. Our streaming engine uses bounded memory, cache reuse, and the task representation learned during training to sustain interaction within context and latency limits.

\ourmodel combines cross-game behavioral scaling, multi-level task supervision, and memory-supported long-horizon interaction in one policy.

\section{CPT Scaling and Cross-Game Generalization}
\label{app:cpt_scaling}
\subsection{Data Mixtures and Validation Distributions}
\label{app:cpt_diversity}

The controlled CPT grid holds its source inventory fixed while varying the share of Minecraft trajectories. S1 uses Minecraft exclusively. S2-S4 draw from the same 80 canonical game labels with different sampling weights, and S5 removes Minecraft, leaving 79 labels. You can find more details in Table~\ref{tab:cpt_mixture_diversity}. Consequently, the five mixtures should not be interpreted as a monotonic increase in the number of game worlds. We summarize diversity using the effective number of worlds, $\exp(-\sum_w p_w\log p_w)$, which accounts for the balance of gameplay samples across game labels.

\begin{table}[t]
\centering\small
\caption{\textbf{Measured diversity of the CPT game mixtures.} World labels are canonicalized across sources. Effective worlds denotes exponential Shannon diversity, while the inverse Simpson index emphasizes high-probability worlds.}
\label{tab:cpt_mixture_diversity}
\setlength{\tabcolsep}{4pt}
\begin{tabular}{lrrrrr}
\toprule
Mixture & Worlds & MC share & Largest share & Eff. worlds & Inv. Simpson\\
\midrule
S1 & 1 & 100.0\% & 100.0\% & 1.0 & 1.0\\
S2 & 80 & 51.4\% & 51.4\% & 14.5 & 3.7\\
S3 & 80 & 27.1\% & 27.1\% & 35.2 & 11.9\\
S4 & 80 & 2.8\% & 4.2\% & 60.1 & 52.6\\
S5 & 79 & 0.0\% & 4.3\% & 58.7 & 51.0\\
\bottomrule
\end{tabular}
\end{table}

The fixed non-Minecraft pool contains 124 source-stream labels mapping to 79 canonical game labels. The three validation distributions separate Minecraft held-out clips, training-covered game clips, and title-disjoint unseen-game clips. The latter is game-balanced, with 28 unseen titles and 160 clips per title (4,480 clips in total). Grouping by source recording and removing title aliases limit overlap between training and validation. These splits distinguish transfer of the action interface from gains limited to games in the training mixture.

\subsection{Scaling Results from GameScaling}
\label{app:gamescaling_results}

This subsection collects the scaling results from GameScaling~\cite{gamescaling} summarized in Sec.~\ref{sec:exp_cpt_scaling}. These results characterize how model size, gameplay budget, and game composition affect CPT within the evaluated configurations and provide context for Gamer's CPT stage.

\begin{figure*}[!htbp]
    \centering
    \includegraphics[width=0.85\linewidth]{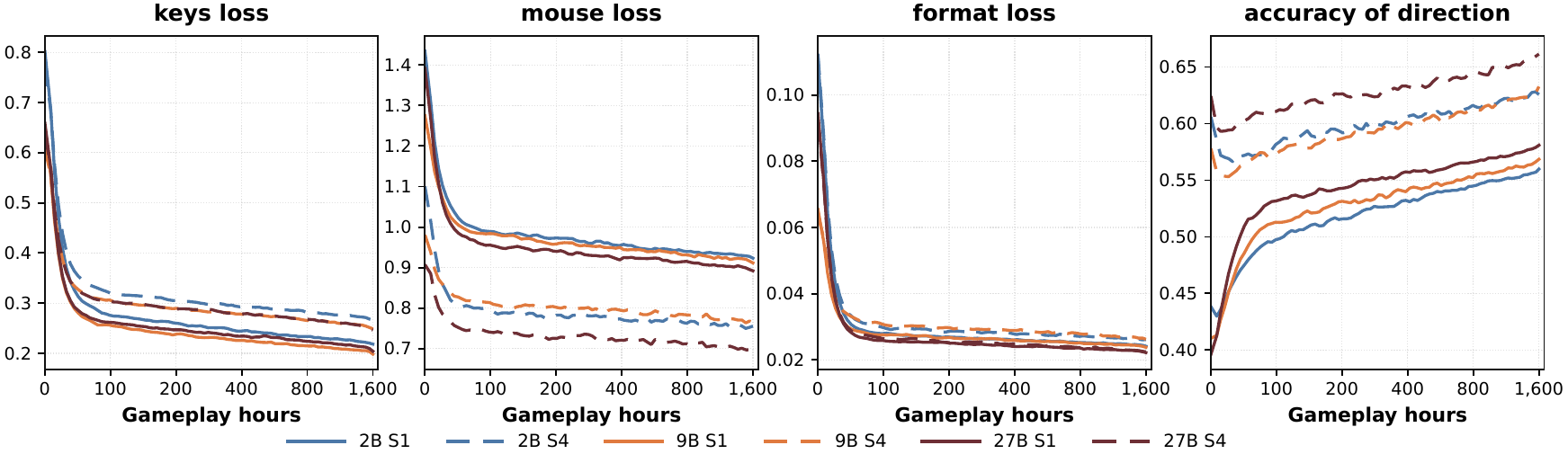}
    \caption{
    \textbf{Early adaptation to the shared keyboard-mouse interface.}
    The figure reports keyboard, mouse, and action-format losses together with directional prediction accuracy for 2B, 9B, and 27B models under the S1 and S4 mixtures.
    The horizontal axis denotes cumulative gameplay hours. Results from GameScaling~\cite{gamescaling}.
    }
    \label{fig:cpt_action_losses}
\end{figure*}

\paragraph{Interface adaptation.}
Figure~\ref{fig:cpt_action_losses} reports keyboard, mouse, and action-format losses together with directional prediction accuracy. The rapid early decrease in these losses and the improvement in directional accuracy suggest that initial CPT learning partly concerns adaptation to the shared keyboard-mouse interface.

\begin{figure*}[!t]
    \centering
    \includegraphics[width=0.83\linewidth]{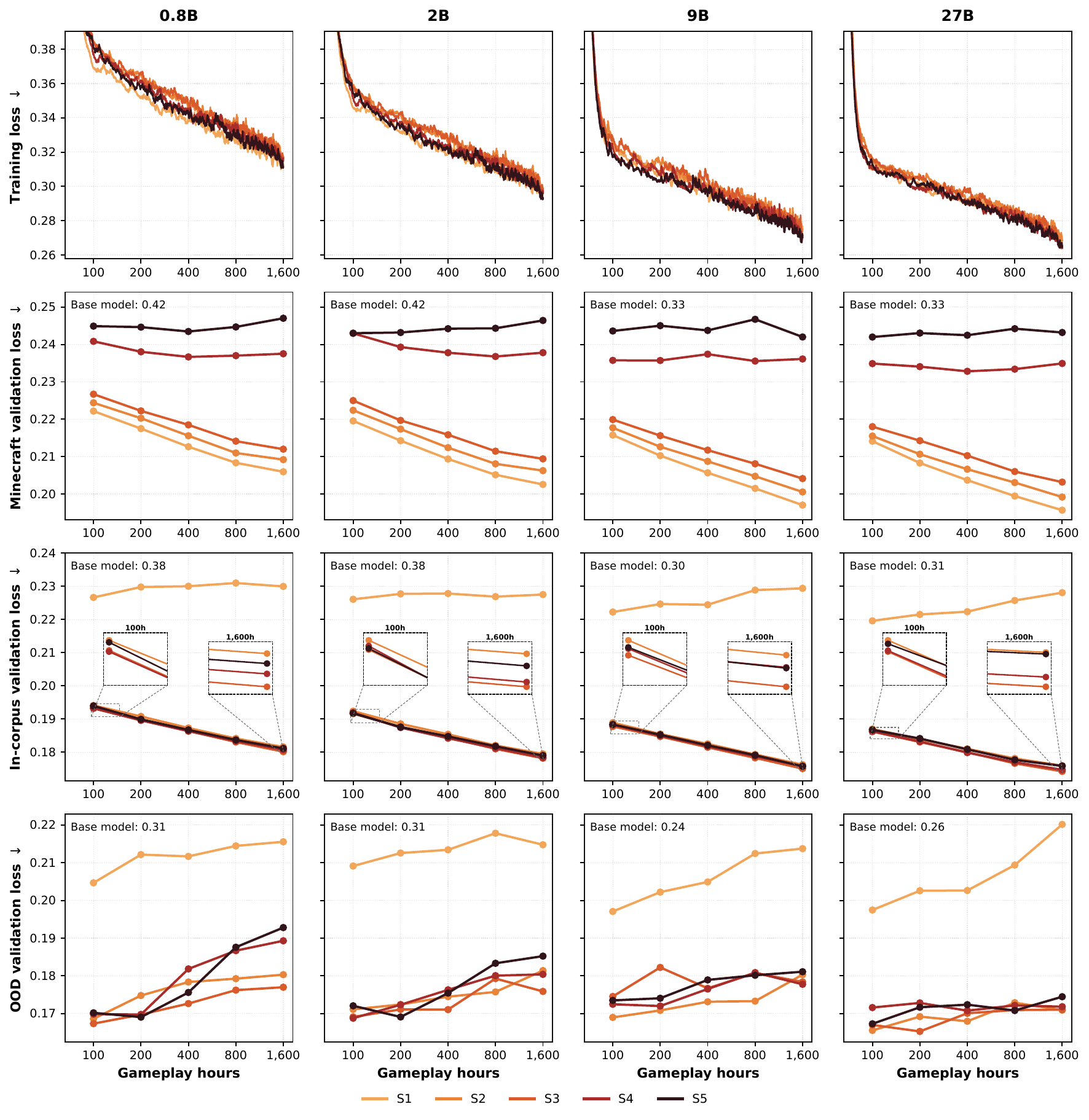}
    \caption{
    \textbf{CPT scaling depends on model capacity, interaction budget, and game composition.}
    The rows show training loss, Minecraft validation loss, validation loss on other training games, and validation loss on unseen games.
    Columns correspond to model scale, while S1-S5 denote different game mixtures.
    The final column compares fitted and measured losses. Results from GameScaling~\cite{gamescaling}.
    }
    \label{fig:cpt_scaling_losses}
\end{figure*}

\paragraph{Training and validation losses.}
Figure~\ref{fig:cpt_scaling_losses} reports training loss and validation losses on Minecraft, other training games, and unseen games. The curves show that gains from additional gameplay data depend on model capacity and game composition. In particular, validation loss on distant unseen games plateaus or increases after early adaptation, even while training loss continues to decrease.

\begin{figure*}[!t]
    \centering
    \includegraphics[width=0.77\linewidth]{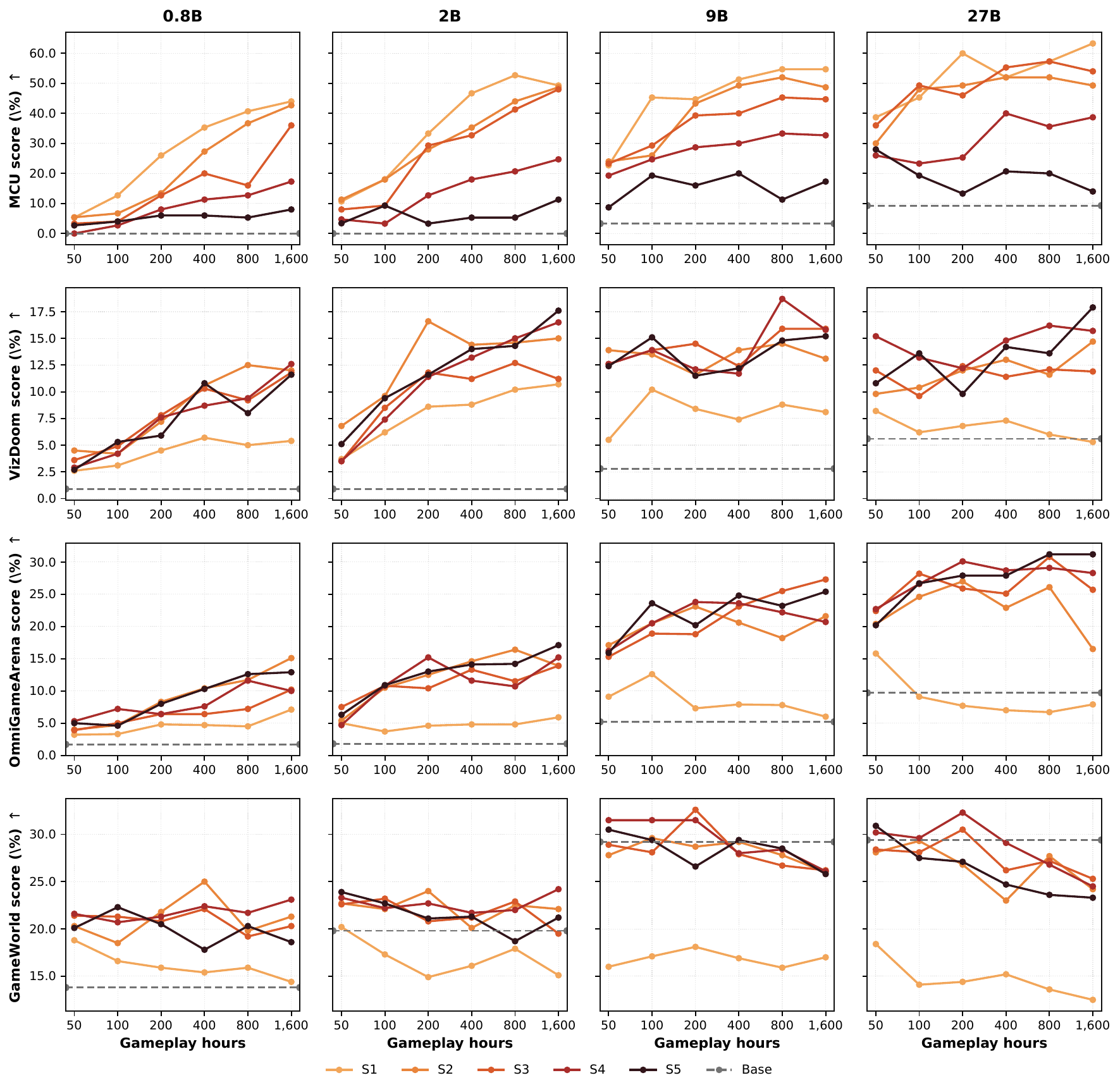}
    \caption{\textbf{Gameplay benchmark scaling across training budgets.} Rows show MCU, ViZDoom, OmniGameArena, and GameWorld, and columns show the four Qwen3.5 Dense scales. Colored curves compare the S1-S5 training mixtures across 50-1,600 gameplay hours, with dashed lines denoting the unadapted backbone. Multi-game mixtures support transfer to related environments. Scores are measured under time-paused interaction. Results from GameScaling~\cite{gamescaling}.}
    \label{fig:cpt_benchmark_hours}
\end{figure*}

\paragraph{Gameplay benchmark performance.}
Figure~\ref{fig:cpt_benchmark_hours} reports performance on MCU, ViZDoom, OmniGameArena, and GameWorld under time-paused interaction. These results complement the validation-loss analysis with executable gameplay outcomes. Within the evaluated configurations, mixed-game training supports transfer to related environments, while gains on more distant games remain limited.

\subsection{Composition-Aware Scaling-Law Fitting}
\label{app:cpt_scaling_fit}

This subsection summarizes the fitting procedure and results reported in GameScaling~\cite{gamescaling}, using the notation of Sec.~\ref{sec:cpt_scaling}.
The composition-aware scaling relation is fitted separately to Minecraft validation loss and validation loss over the remaining training-game distribution:
\begin{equation}
\mathcal{L}_{\mathrm{val}}(N,D,q)
=
E
+
\frac{C_N}{N^a}
+
\frac{C_D q^\eta}{D^b}
+
\frac{C_q}{(q+\epsilon)^c}.
\label{eq:appendix_scaling_law}
\end{equation}

The fitted parameter vector is
\[
\boldsymbol{\phi}
=
\left(
E,
C_N,
C_D,
C_q,
a,
b,
c,
\eta,
\epsilon
\right).
\]

Model size $N$ is measured in billions of parameters and data budget $D$ in hours of gameplay interaction.
For Minecraft validation, we use
\[
q=w_{\mathrm{MC}},
\]
where $w_{\mathrm{MC}}$ is the Minecraft fraction of the gameplay mixture.
For validation over the other training games, we use
\[
q=1-w_{\mathrm{MC}}.
\]

The complete fit uses the 120 controlled configurations described in Sec.~\ref{sec:exp_cpt_scaling}.
All validation losses are measured from checkpoints that complete the corresponding training budget and decay phase.

\textbf{Fitting objective.}
Let $\mathcal J_{\mathrm{fit}}$ denote the set of configurations used for fitting and let $L_j^{\mathrm{obs}}$ denote the observed validation loss of configuration $j$.
We define the log residual as
\[
e_j(\boldsymbol{\phi})
=
\log
\mathcal{L}_{\mathrm{val}}
\left(
N_j,D_j,q_j,
\boldsymbol{\phi}
\right)
-
\log
L_j^{\mathrm{obs}}.
\]

The fitted parameters minimize a Huber objective:
\begin{equation}
\boldsymbol{\phi}^{\star}
=
\arg\min_{\boldsymbol{\phi}}
\sum_{j\in\mathcal J_{\mathrm{fit}}}
h_{\delta}
\left(
e_j(\boldsymbol{\phi})
\right),
\label{eq:scaling_fit_objective}
\end{equation}
where $h_{\delta}$ is quadratic for small residuals and linear for large residuals.
We use $\delta=10^{-3}$.
The logarithmic residual emphasizes relative prediction error, while the Huber objective reduces sensitivity to individual outlier configurations.

We optimize Eq.~\eqref{eq:scaling_fit_objective} with L-BFGS-B~\cite{byrd1995limited} and retain the solution with the lowest objective among 192 initializations.
The positive scale parameters $E$, $C_N$, $C_D$, and $C_q$ are optimized in log space.
We constrain
\[
a,b\in[0,3],
\qquad
\eta\in[-3,3],
\qquad
c\in[0,5],
\qquad
\epsilon\in[10^{-4},1].
\]

\textbf{Fitted parameters.}
Table~\ref{tab:scaling_parameters} reports the point estimates obtained from the complete controlled grid.

\begin{table}[t]
    \centering
    \small
    \setlength{\tabcolsep}{5pt}
    \renewcommand{\arraystretch}{1.08}
    \begin{tabular}{lcc}
        \toprule
        \textbf{Parameter}
        & \textbf{Minecraft}
        & \textbf{Training Games} \\
        & $q=w_{\mathrm{MC}}$
        & $q=1-w_{\mathrm{MC}}$ \\
        \midrule
        
        $E$
        & 0.047
        & 0.136 \\
        
        $C_N$
        & 0.019
        & 0.014 \\
        
        $a$
        & 0.15
        & 0.17 \\
        
        $C_D$
        & 0.106
        & 0.078 \\
        
        $b$
        & 0.16
        & 0.14 \\
        
        $\eta$
        & 0.24
        & 0.10 \\
        
        $C_q$
        & 0.0987
        & 0.0031 \\
        
        $c$
        & 0.13
        & 1.00 \\
        
        $\epsilon$
        & 0.011
        & 0.039 \\
        
        \bottomrule
    \end{tabular}
    \caption{
    \textbf{Fitted parameters of the composition-aware CPT scaling relation.}
    Separate parameter sets are estimated for Minecraft validation loss and validation loss over the remaining training-game distribution.
    }
    \label{tab:scaling_parameters}
\end{table}

\textbf{Block-wise holdout evaluation.}
To evaluate predictive performance beyond the configurations used for fitting, we remove complete blocks of model sizes, data budgets, or game mixtures.
For each holdout setting, the scaling relation is refit using the remaining configurations, and RMSE is computed only over the held-out block. The holdout results show that predictions over model size and interaction budget remain accurate near the measured range.
Mixture prediction is more sensitive to the availability of nearby compositions, particularly for Minecraft when S4 is held out.

\begin{table}[t]
    \centering
    \small
    \setlength{\tabcolsep}{4pt}
    \renewcommand{\arraystretch}{1.08}
    \begin{tabular}{lccc}
        \toprule
        \textbf{Held-Out Block}
        & \textbf{Configurations Used for Fitting}
        & \textbf{Minecraft RMSE}
        & \textbf{Training-Game RMSE} \\
        \midrule
        
        27B model
        & 0.8B, 2B, 9B
        & 0.0026
        & 0.0016 \\
        
        1,600 hours
        & 50-800 hours
        & 0.0027
        & 0.0018 \\
        
        50 hours
        & 100-1,600 hours
        & 0.0025
        & 0.0022 \\
        
        Mixture S2
        & S1, S3, S4, S5
        & 0.0020
        & 0.0016 \\
        
        Mixture S3
        & S1, S2, S4, S5
        & 0.0021
        & 0.0008 \\
        
        Mixture S4
        & S1, S2, S3, S5
        & 0.0057
        & 0.0005 \\
        
        \bottomrule
    \end{tabular}
    \caption{
    \textbf{Block-wise holdout prediction error for the CPT scaling relation.}
    Each row excludes the indicated block during fitting and evaluates RMSE only on that held-out configuration.
    }
    \label{tab:scaling_holdout}
\end{table}

Different parameter combinations can produce similar loss surfaces, making it hard to identify individual exponents and offsets separately.
The predicted loss surface is substantially more stable than the individual fitted parameters.
We therefore assess the relation using the observed scaling curves, within-mixture explanatory power, and block-wise holdout error, without relying on any single coefficient.

We additionally estimate fitting uncertainty using 300 stratified bootstrap resamples within each game mixture.
These intervals characterize uncertainty conditional on the observed training runs and do not include variation across independent training seeds.

\subsection{Scaling Robustness and Predictive Uncertainty}
\begin{figure*}[t]
    \centering
    \includegraphics[width=0.97\linewidth]{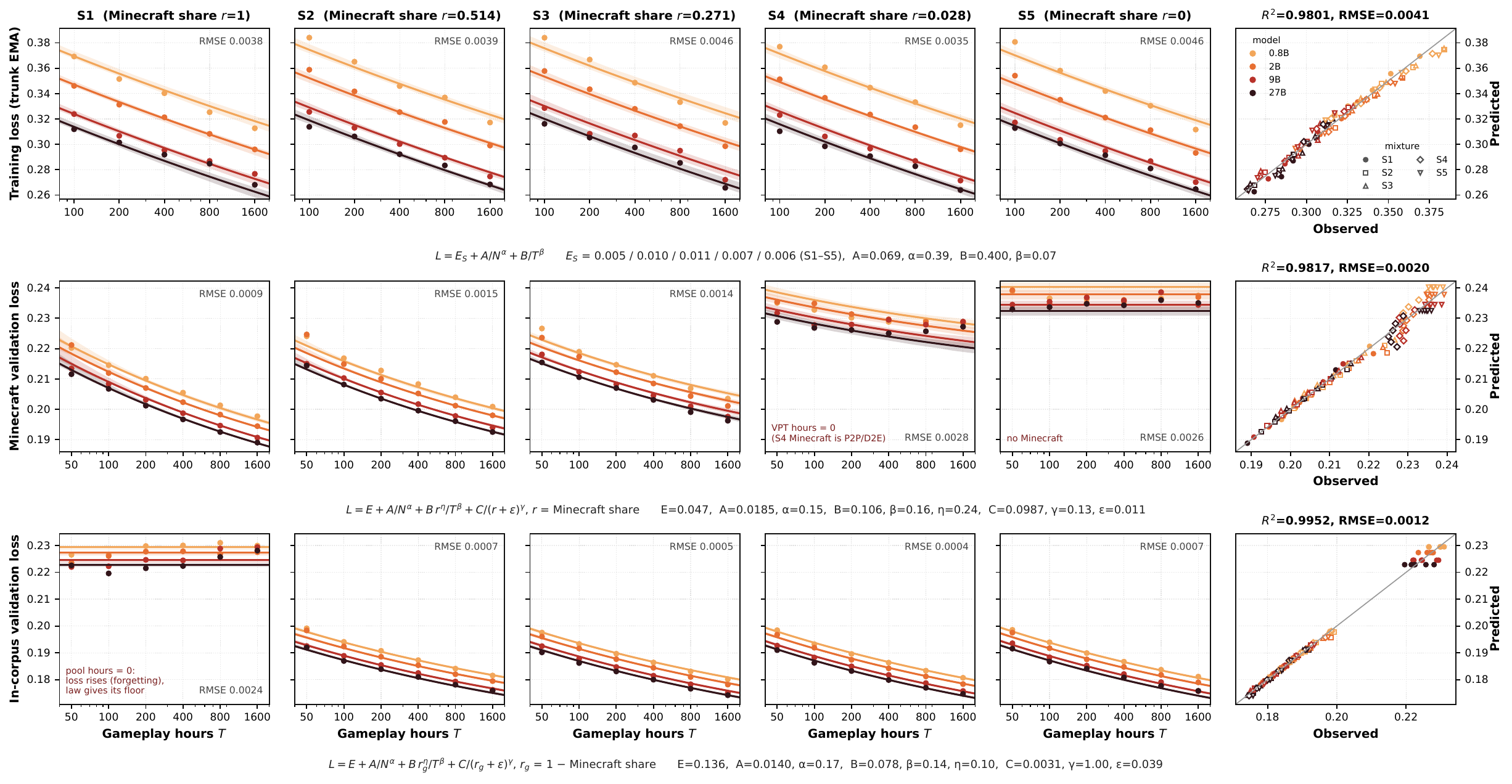}
    \caption{\textbf{Scaling-law fits with predictive uncertainty.} The panels show the training, Minecraft, and in-corpus loss laws across S1-S5, with measurements, predictions, and stratified-bootstrap bands. Prediction bands reflect variation across resampled experimental configurations within the observed grid, not seed-to-seed variance in model training.}
    \label{fig:gamescaling_uncertainty}
\end{figure*}

The composition-aware relation explains both differences among mixtures and their different slopes with respect to gameplay hours. An equally parameterized categorical-offset alternative cannot predict a held-out mixture and yields higher training-grid errors: Minecraft RMSE 0.0044 versus 0.0020 and in-corpus RMSE 0.0035 versus 0.0012. By contrast, distant unseen-game losses generally rise after the early interface phase, and their within-mixture scaling fit is weak. We therefore interpret the fitted relation as an empirical description of the measured in-corpus data regime rather than a universal law for unseen environments. Figure~\ref{fig:gamescaling_uncertainty} and Table~\ref{tab:scaling_holdout} demonstrate the fit's accuracy and limitations.

\subsection{Frontier-Model Progress and CPT Transfer}
\label{app:cpt_frontier_progress}

\begin{figure*}[t]
    \centering
    \includegraphics[width=0.92\linewidth]{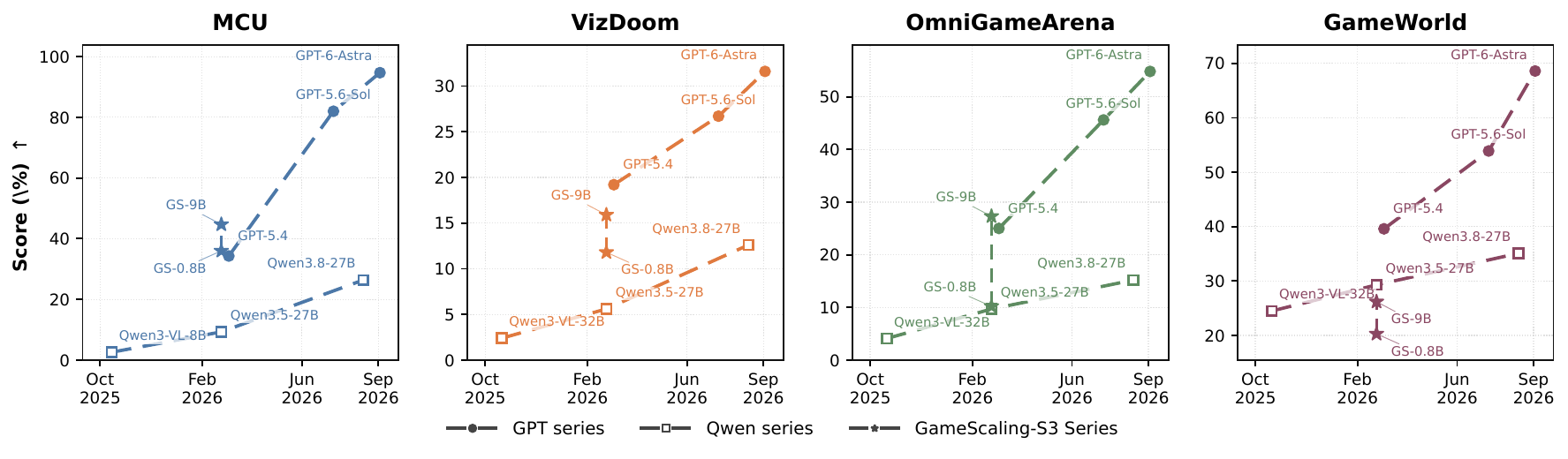}
    \caption{\textbf{Frontier-model progress on four public game benchmarks.} Scores are plotted against model release date under time-paused evaluation. Circles indicate the GPT series, squares the Qwen series, and stars the GameScaling-S3 checkpoints at 1,600 gameplay hours (marked GS in the original plot). These results are from the controlled scaling study, not the HSR-augmented \ourmodel-CPT model. The comparison shows gains from gameplay adaptation alongside the contribution of general backbone capabilities, particularly on distant worlds.}
    \label{fig:cpt_frontier_progress}
\end{figure*}

Figure~\ref{fig:cpt_frontier_progress} places the GameScaling CPT checkpoints alongside successive general-purpose model releases on MCU, ViZDoom, OmniGameArena, and GameWorld. The controlled GameScaling results show that gameplay pretraining enables smaller backbones to achieve substantial gains on the source and related 3D environments. However, improvements across generations of general-purpose models are also pronounced, particularly on GameWorld, where distant-world transfer remains limited for the CPT checkpoints. The trends reflect progress from both game-specific adaptation and improvements to the underlying multimodal model. All points in this comparison use time-paused evaluation, so they characterize gameplay capability without conflating it with the real-time latency effects examined in the main experiments.

\subsection{Test-Time Interaction Budgets}
\label{app:cpt_test_time}
\begin{figure}[!htbp]
    \centering
    \includegraphics[width=0.74\linewidth]{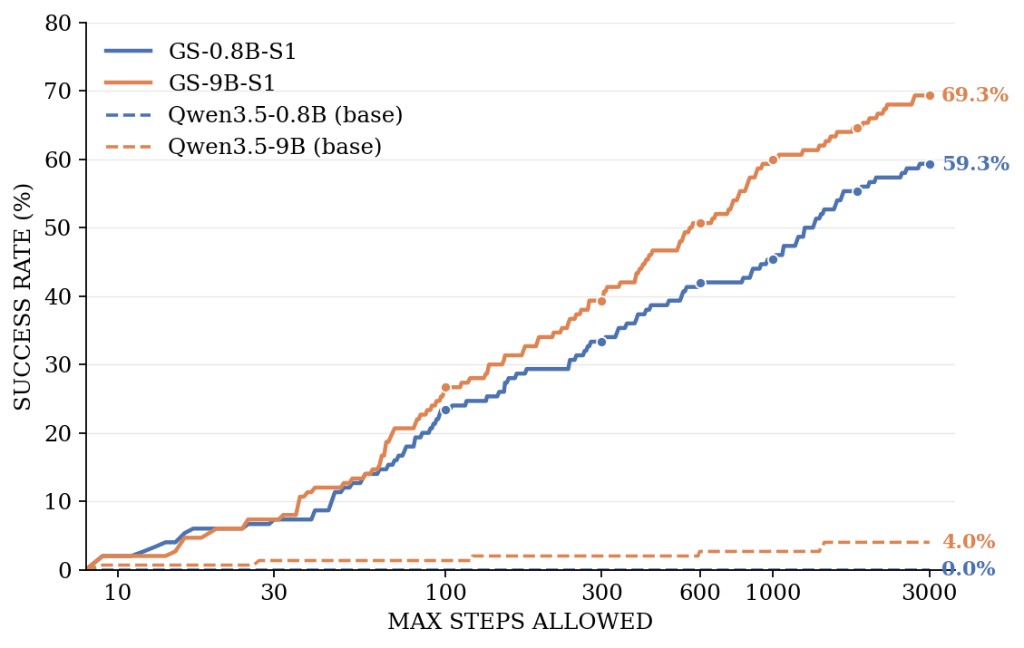}
    \caption{\textbf{Test-time scaling on MCU.} Cumulative task success versus available interaction steps (up to 3,000) for CPT checkpoints and unadapted Qwen3.5 backbones. The primary MCU protocol uses a 600-step cap. This supplementary sweep illustrates how additional interaction opportunities affect task completion.}
    \label{fig:gamescaling_test_time}
\end{figure}

As illustrated in Figure~\ref{fig:gamescaling_test_time}, increasing the test-time budget from 600 to 3,000 steps raises success for the CPT models in this comparison. The two reported checkpoints reach 59.3\% and 69.3\% at 3,000 steps, while the unadapted backbones remain near zero. A larger interaction budget helps policies that have already learned useful control, but cannot replace CPT. These curves are supplementary to the fixed-budget primary MCU scores.

\section{Streaming-vLLM Implementation Details}
\label{app:streaming_details}
\subsection{FSM-Guided Drafting and BF16 Verification}
\label{app:speculative_sampling}

\textbf{FSM grammar coverage.} The SM90 fast FSM recognizes the boundaries \texttt{<think>}, \texttt{</think>}, \texttt{<|action\_start|>}, and \texttt{<|action\_end|>}. It can draft the end of an empty reasoning span or the beginning of an action packet after a finished reasoning span. Within the XYZ mouse-coordinate fields, the FSM drafts formatting tokens following numerical values, not the coordinate values themselves. In the six subsequent key groups, it drafts predictable delimiters, blank-key templates, and the action terminator. Candidate tokens are proposed opportunistically when the current grammar state supports them.

\textbf{Target-model verification.} The drafts are checked by the target model before being committed. A candidate prefix is accepted only where each drafted token equals the token that target-model verification selects at the corresponding position under the active sampling rule (greedy argmax or temperature sampling); rejected suffixes and their speculative cache states are discarded. Throughout generation, the FSM simultaneously enforces grammar-level constrained decoding and speculative drafting: at every decoding and verification position, it restricts token selection to valid protocol transitions via state-dependent vocabulary bitmasks in the fused sampler---including for tokenizers whose action delimiters span multiple subword pieces---while proposing predictable structural continuations as drafts for parallel target-model verification, without replacing the policy's semantic action decisions over coordinate values or valid key combinations. 

\subsection{Incremental RoPE Cache Refresh}
\label{app:rope_refresh}

\begin{figure*}[t]
    \centering
    \includegraphics[width=0.85\linewidth]{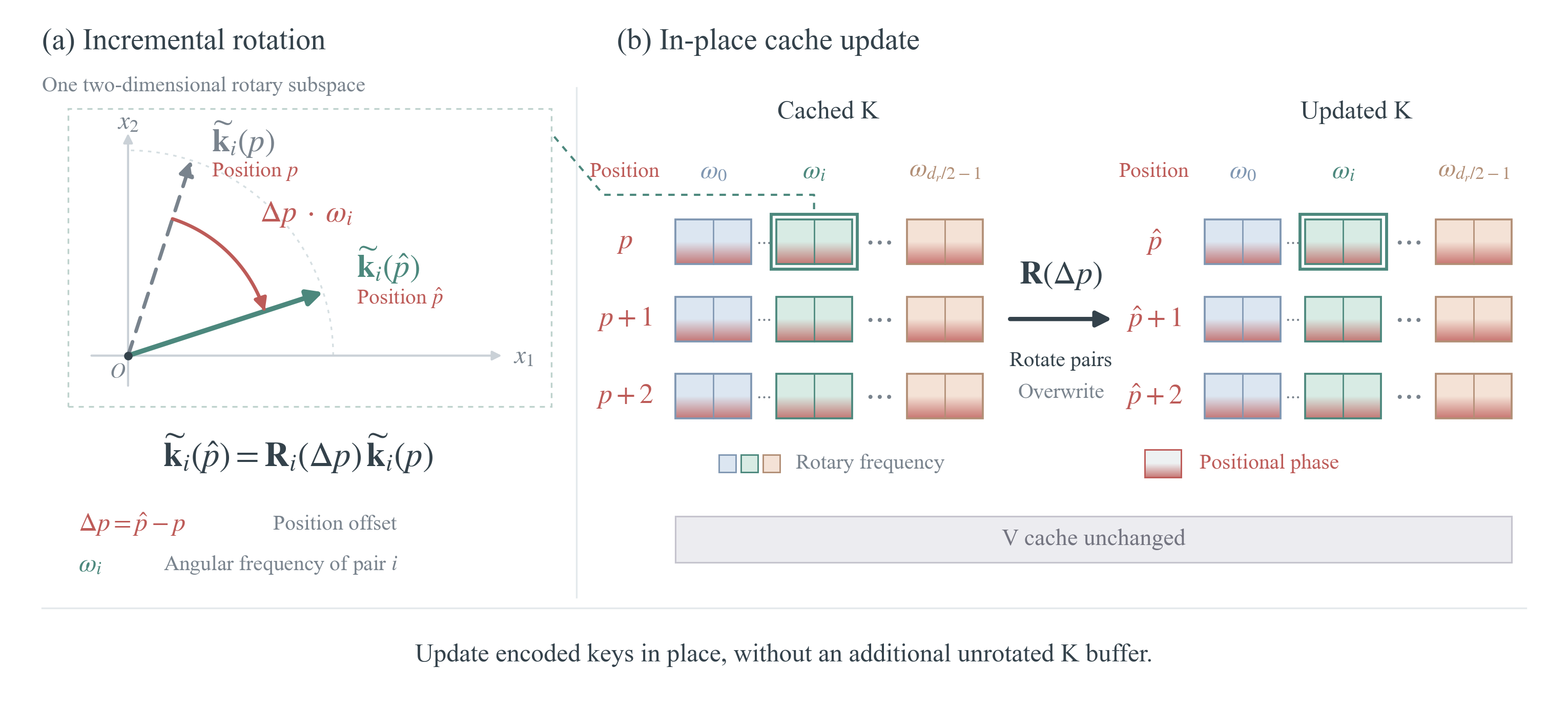}
    \caption{
    \textbf{Incremental RoPE refresh for retained KV states.}
    A cached key moved from logical position $p$ to $\widehat p$ is updated by an additional rotation determined by $\Delta p=\widehat p-p$.
    The same positional offset is applied independently to each RoPE frequency pair and written back in place, while the value cache remains unchanged.
    The figure illustrates $\Delta p<0$, corresponding to a leftward shift after context compaction.
    }
    \label{fig:rope_refresh_geometry}
\end{figure*}

Figure~\ref{fig:rope_refresh_geometry} illustrates the incremental positional refresh used for retained KV states after context compaction.
A cached key originally associated with logical position $p$ may be reassigned to a new position $\widehat p$.
Rather than recomputing its key projection, Streaming-vLLM applies only the positional difference
\[
\Delta p
=
\widehat p-p.
\]

For a pair of cached key components $(u,v)$ associated with rotary frequency $\omega_i$, let
\[
\phi
=
\Delta p\,\omega_i.
\]
The incremental update is implemented as
\begin{equation}
\begin{aligned}
u'
&=
u\cos\phi-v\sin\phi,\\
v'
&=
u\sin\phi+v\cos\phi.
\end{aligned}
\label{eq:rope_pairwise_update}
\end{equation}

Both updated components are computed from the original pair before being written back to the cache.
The implementation follows the component-pair layout used by the underlying model and applies the corresponding update independently to all dimensions participating in RoPE.

Tokens whose logical positions remain unchanged have $\Delta p=0$ and therefore require no modification.
The update is performed in place on the retained key cache.
No unrotated key representation is stored, and the value cache is left unchanged.
For RoPE variants with multiple positional axes, the same correction is applied independently to the component groups associated with each axis.

This update corrects only the rotary positional phase.
As discussed in Sec.~\ref{sec:streaming_context}, retained KV states continue to follow the stateful streaming semantics of Streaming-vLLM rather than being reconstructed through full re-encoding after memory compaction.

\subsection{Hardware and Model-Architecture Coverage}
\label{app:streaming_architecture}

\textbf{Hardware mapping and evaluated backbones.} The Ampere measurements use NVIDIA A800 GPUs, while the Hopper measurements use NVIDIA H200 GPUs. Figure~\ref{fig:streaming_vllm_configuration_latency} compares Qwen3-VL-8B, a representative dense visual-language model, and Qwen3.5-35B-A3B, a representative mixture-of-experts model. Both are run with multiple tensor-parallel configurations and numerical precision settings. The Qwen3-VL and Qwen3.5 families are supported at additional dense and MoE sizes listed in Table~\ref{tab:training_recipe}, but this figure should not be read as an exhaustive per-checkpoint benchmark.

\textbf{Representative cross-hardware behavior.} Under TP4 and BF16, Qwen3-VL-8B has a mean request latency of 147.64 ms on datacenter Ampere and 75.66 ms on Hopper. Qwen3.5-35B-A3B has corresponding latencies of 205.77 ms and 110.08 ms. On Hopper, TP4 FP8 W8A8 further yields 74.69 ms and 92.19 ms, respectively. These figures come from the multi-turn latency-composition sweep and are not matched compute-cost comparisons across GPU generations. These results show that the same stateful serving design supports dense and MoE backbones on both architectures. The best parallelism and precision settings vary with the model and hardware.

\subsection{Latency Composition and Control-Rate Interpretation}
\label{app:streaming_latency}

\begin{figure*}[t]
    \centering
    \includegraphics[width=\linewidth]{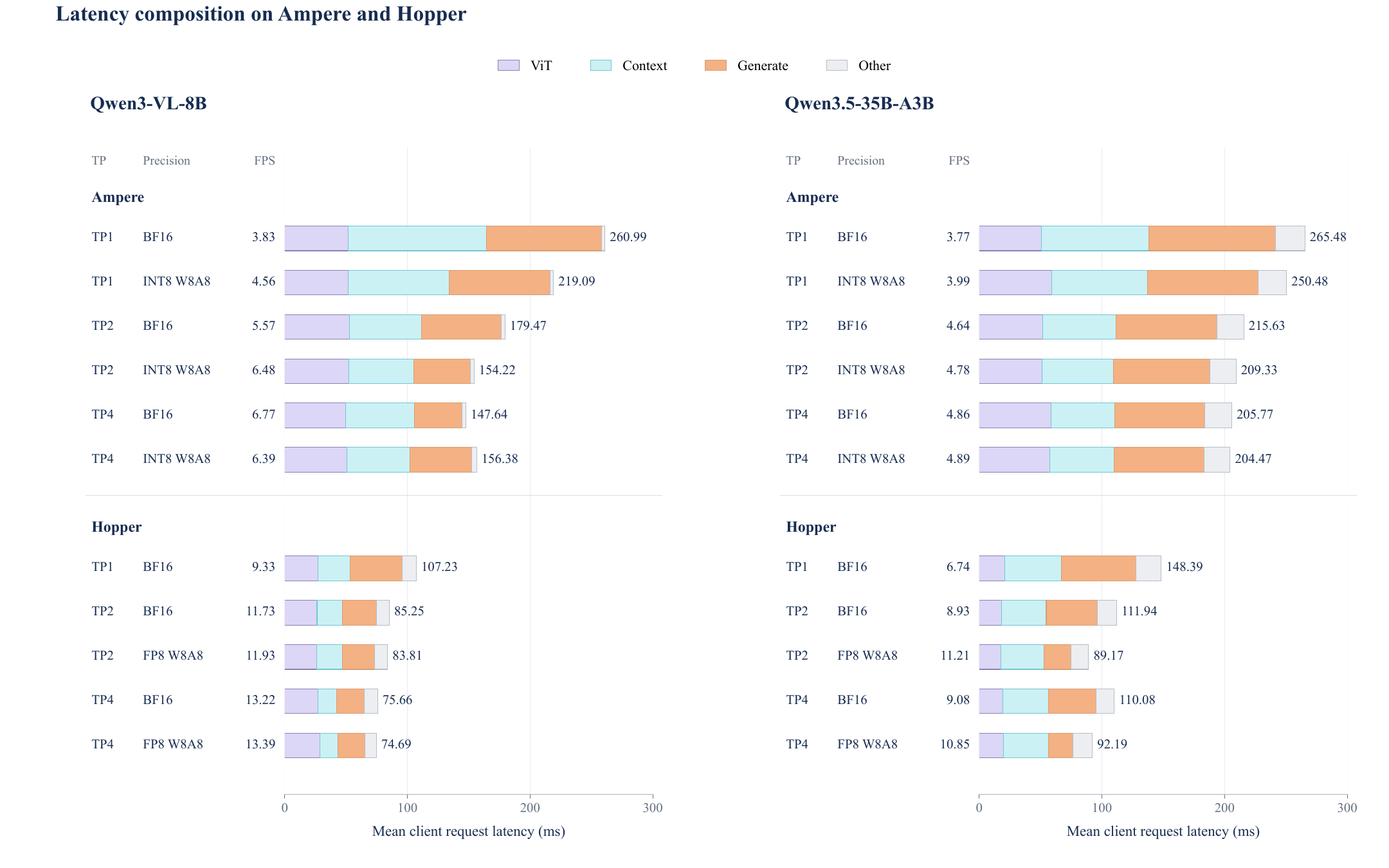}
    \caption{\textbf{Cross-backbone request latency on NVIDIA A800 (Ampere) and H200 (Hopper).} The existing sweep compares Qwen3-VL-8B (Dense) and Qwen3.5-35B-A3B (MoE) across tensor-parallel degrees and BF16 or W8A8 execution. Stacked values separate vision processing, context (prefill), generation (decode), and remaining overhead, measured as mean request latency. These results characterize representative hardware and backbone coverage rather than all supported model sizes. The theoretical control rate in Table~\ref{tab:streaming_vllm_ablation} is derived from request time, not measured at the client.}
    \label{fig:streaming_vllm_configuration_latency}
\end{figure*}

The end-to-end request time comprises vision encoding, context processing, action generation, and runtime overhead. In the figure, Ampere denotes NVIDIA A800, and Hopper denotes NVIDIA H200. Figure~\ref{fig:streaming_vllm_configuration_latency} compares these components across model architectures, hardware generations, parallel settings, and numerical precision. The system ablation reports step rate (SR), the reciprocal of end-to-end request latency, and a theoretical control rate (CR) obtained by multiplying SR by the six low-level blocks in each action packet. CR is a latency-derived upper bound, not a direct measurement of the client execution cadence. The latter uses approximately 33~ms per block (30~Hz) when actions arrive in time.

\FloatBarrier

\section{Honkai: Star Rail Evaluation Protocol and Taxonomy}
\label{app:hsr_protocol}

\subsection{World Diversity and Interactive Complexity}
\label{app:hsr_world_diversity}

\begin{figure*}[t]
    \centering
    \includegraphics[width=0.92\textwidth]{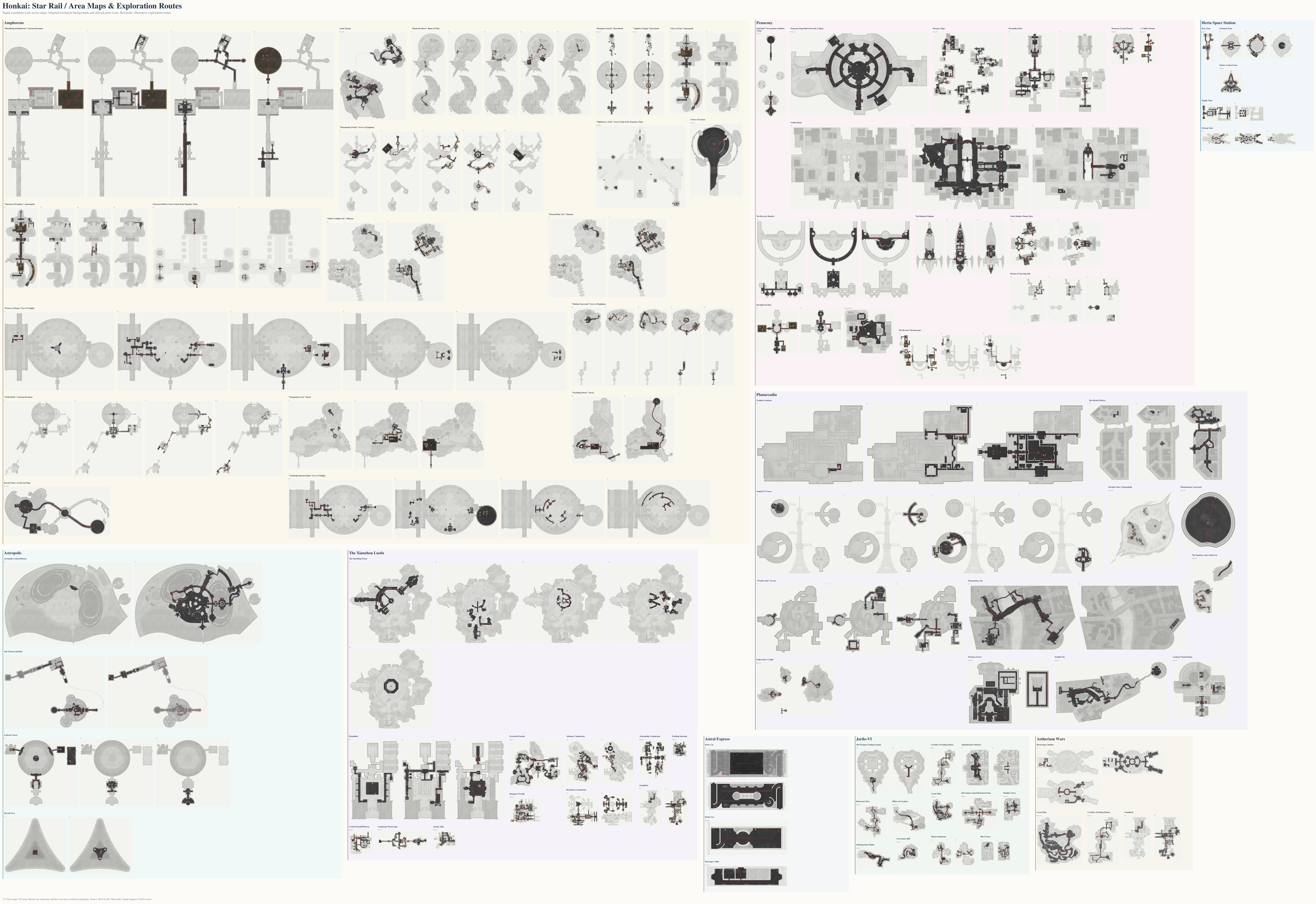}
    \caption{\textbf{Exploration diversity in \textit{Honkai: Star Rail}.}
    Map layouts vary in geometry, connectivity, and floor structure, and point icons indicate resources and other interactive locations. Red paths illustrate the possible exploration routes. }
    \label{fig:hsr_world_maps}
\end{figure*}

\textit{Honkai: Star Rail} provides a useful setting for studying generalist game agents because its complexity comes from combining familiar interactions across a large, varied world, rather than from a single difficult mechanic. Figure~\ref{fig:hsr_world_maps} brings together 171 floor maps from 85 areas across the game's major worlds. The maps range from compact rooms and branching corridors to large open layouts, interconnected structures, and multi-floor environments. Even when presented at a common coordinate scale, their differences are apparent. An agent cannot assume that a route learned in one area will work in another. It must adjust to new geometry, identify transitions between areas or floors, and preserve its destination as the scene changes.

\FloatBarrier

The many point icons also reveal how exploration and task execution are interleaved. Individual areas contain resources and other points of interest alongside locations relevant to progression and interaction. A route can pass several optional objects before reaching the target, and the shortest visible path need not be a feasible sequence of game actions. The red paths in Figure~\ref{fig:hsr_world_maps} are \textbf{schematic exploration routes}, not gameplay-verified optimal solutions. They illustrate the need to plan through branching spaces, but do not establish that any particular route is the fastest or most reliable. The annotated maps are provided only to describe the environment and are \textbf{not given to the evaluated agents} as observations or privileged navigation input.

\textbf{Accessible gameplay does not imply easy autonomous control.} For a human player, quest objectives, map markers, and familiar interaction prompts make many individual steps understandable. A vision-language-action agent receives only the rendered screen and must turn this guidance into timed keyboard and mouse inputs. It must recognize where it is, distinguish relevant objects from nearby distractions, navigate while the camera and scene change, and determine when an interaction has actually completed. Longer tasks add another difficulty: navigation, dialogue, UI manipulation, combat, and local puzzles may appear within the same quest. Failure at one step can invalidate a later plan, and an agent must remember earlier progress even after those observations leave its active context. The challenge is therefore not simply to play any one minigame, but to coordinate heterogeneous skills over time.

This combination makes the game useful for both research and practical evaluation. Short interactive challenges probe whether a policy can adapt to different local rules, while side quests test whether those skills support sustained, instruction-conditioned progress. The same interface also reflects conditions faced by automated game testing: new content introduces routes, objects, objectives, and mechanics that cannot all be anticipated by hand-written scripts. A generalist agent able to execute these tasks through screen observations and device-level actions could support exploration, quest verification, and regression testing without modifications to the game engine. These are \textbf{potential applications}, not deployment benefits measured by the map itself. Our experiments cover a defined subset of this broader environment. Although \ourmodel-CPT is trained on other \textit{Honkai: Star Rail} gameplay, neither of the two evaluated side quests nor any of the 40 scored minigame tasks is included in training. This split tests generalization to held-out tasks within a familiar game world, not zero-shot transfer to a game unseen during training.

\subsection{Combat Mechanics and Strategic Decision-Making}
\label{app:hsr_combat}

\begin{figure}[H]
    \centering
    \includegraphics[width=0.70\linewidth]{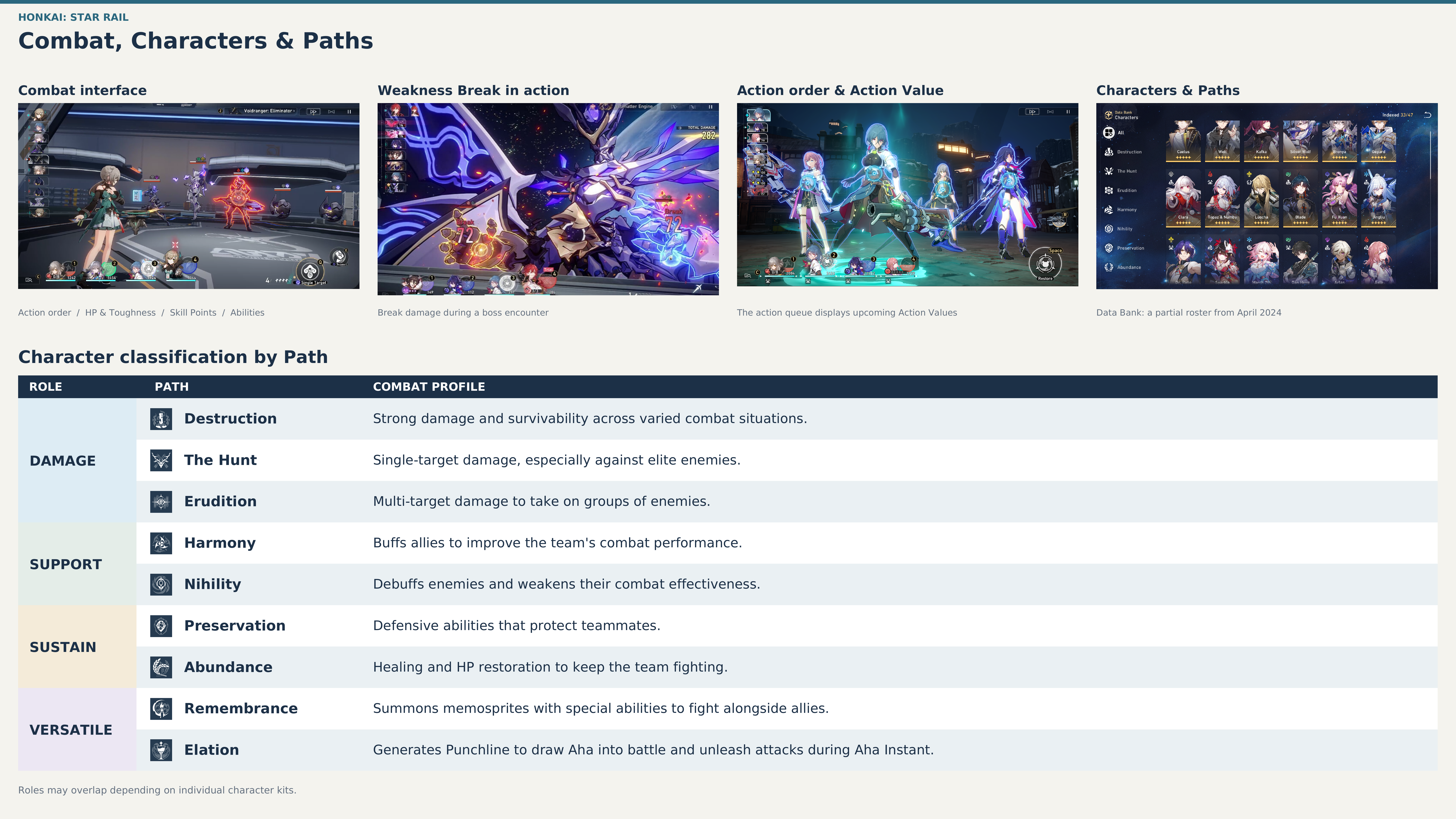}
    \caption{\textbf{Combat mechanics and character roles in \textit{Honkai: Star Rail}.} The examples illustrate (1) the combat interface, including action order, HP and Toughness, team status, skill points, and available abilities, (2) a Weakness Break during a boss encounter, (3) the Action Value display, and (4) a character Data Bank snapshot. The lower panel summarizes the nine Paths by representative combat roles, including damage, support, sustain, and more specialized mechanics. Path groupings summarize representative combat roles, and the figure provides game-mechanics context rather than a measured policy rollout.}
    \label{fig:hsr_combat_mechanics}
\end{figure}

Exploration is only one part of \textit{Honkai: Star Rail}. Combat adds a different decision-making problem within the same game interface. The game uses turn-based encounters, but choosing an action still depends on information that changes throughout the fight. As shown in Figure~\ref{fig:hsr_combat_mechanics}, the combat screen combines the action-order queue with character HP, enemy Toughness, team status, skill points, and available abilities. The action-order display and Action Value indicate when combatants will act, while the Toughness system creates opportunities for Weakness Break. The agent must use these cues together, rather than treating each turn as an isolated image-to-action decision.

Combat decisions involve coupled constraints over multiple turns. Attacking a vulnerable target may advance a Weakness Break, but the agent must also consider the available skill points, the health and status of its team, and which combatant acts next. Character roles introduce further variation. The Paths shown in Figure~\ref{fig:hsr_combat_mechanics} span single-target and multi-target damage, buffs and debuffs, protection and healing, summons, and other specialized combat effects. A useful action can therefore depend on the team composition and on how one character's choice changes the options available to others. Although human players can read these cues from a familiar interface, an autonomous policy must recognize them visually, choose a suitable action, and issue the corresponding executable inputs.

This makes combat complementary to the game's navigation, dialogue, and puzzle mechanics. Navigation requires spatial grounding and route execution, whereas combat requires state-dependent target selection and coordination of limited resources across turns. Both demand that the agent preserve its objective while the available actions and relevant observations change. The held-out side-quest benchmark includes combat milestones alongside other interaction types. The atlas in Figure~\ref{fig:hsr_combat_mechanics} provides context for those milestones and the broader game setting, rather than constituting a separate combat benchmark or evidence of a model's combat performance. Together with the exploration maps, it illustrates why generalist gameplay requires more than a collection of fixed action scripts: the same visual-control policy must adapt to different rules, time scales, and notions of progress within one evolving game world.

\subsection{Long-Horizon Milestones}

The long-horizon benchmark consists of two fixed quest specifications. \emph{Opportunity Makes the Thief (Part I)} contains 30 evaluation milestones grouped into 22 parent steps, with 14 navigation, 14 dialogue, one combat, and one UI milestone. \emph{Bestial Ferocity} contains 55 milestones grouped into 40 parent steps, with 32 navigation, 16 dialogue, two combat, two UI, and three puzzle milestones. Configured time budgets are attached to each required milestone. They are limits for evaluation, not observations of how long an agent took to complete a step. The ordering and terminal conditions are listed in Tables~\ref{tab:museum_milestones} and~\ref{tab:bestial_milestones}.

For each model, the milestone-type rate is the number of verified passes divided by the total milestones of that type in the respective quest. Quest accuracy uses all required milestones as its denominator rather than averaging type-level rates. If a run resumes through a scripted transition, a verified milestone pass does not establish uninterrupted autonomous progress. Missing evaluations are not counted as failures. The published report should state the exact number of valid attempts per model and task when final results are available.

\begin{table}[t]
\centering
\caption{\footnotesize \textbf{Opportunity Makes the Thief (Part I).} The 30 required milestones span 22 parent steps and cover navigation, dialogue, combat, and exhibit recovery. Read down the left block, then the right block.}
\label{tab:museum_milestones}
\vspace{5pt}
\footnotesize
\setlength{\tabcolsep}{2pt}
\renewcommand{\arraystretch}{1.06}
\begin{minipage}[t]{0.485\linewidth}
\vspace{0pt}
\centering
\textbf{First 15 milestones}\par
\vspace{4pt}
\begin{tabular*}{\linewidth}{@{}p{0.12\linewidth}@{\hspace{5pt}}p{0.66\linewidth}@{\extracolsep{\fill}}cr@{}}
\noalign{\hrule height 0.7pt}
\rule{0pt}{12pt}\textbf{ID} & \textbf{Milestone} & \textbf{Type} & \textbf{s} \\[2pt]
\hline
\noalign{\vskip 3pt}
1 & \raggedright Follow the quest marker to locate Pela. & N & 60 \\[0.5pt]
2 & \raggedright Talk to Pela and transition into the museum. & D & 60 \\[0.5pt]
3 & \raggedright Continue the conversation with Pela inside the museum. & D & 60 \\[0.5pt]
4 & \raggedright Follow Pela to the exhibit. & N & 120 \\[0.5pt]
5 & \raggedright Complete the conversation and transition to the Underworld. & D & 60 \\[0.5pt]
6 & \raggedright Finish the current conversation. & D & 60 \\[0.5pt]
7 & \raggedright Follow the quest marker to locate Swetta. & N & 60 \\[0.5pt]
8 & \raggedright Complete the conversation with Swetta. & D & 60 \\[0.5pt]
9.2a & \raggedright Locate and inspect the poster. & N & 180 \\[0.5pt]
9.2b & \raggedright Advance the dialogue associated with the poster. & D & 30 \\[0.5pt]
9.3a & \raggedright Approach the robot guards. & N & 180 \\[0.5pt]
9.3b & \raggedright Advance the dialogue until combat begins. & D & 30 \\[0.5pt]
9.3c & \raggedright Defeat the robot guards. & C & 120 \\[0.5pt]
9.3d & \raggedright Locate the crate containing a clue. & N & 30 \\[0.5pt]
9.3e & \raggedright Complete the dialogue associated with the crate. & D & 30 \\[0.5pt]
\noalign{\vskip 2pt}
\noalign{\hrule height 0.7pt}
\end{tabular*}
\end{minipage}
\hfill
\begin{minipage}[t]{0.485\linewidth}
\vspace{0pt}
\centering
\textbf{Final 15 milestones}\par
\vspace{4pt}
\begin{tabular*}{\linewidth}{@{}p{0.12\linewidth}@{\hspace{5pt}}p{0.66\linewidth}@{\extracolsep{\fill}}cr@{}}
\noalign{\hrule height 0.7pt}
\rule{0pt}{12pt}\textbf{ID} & \textbf{Milestone} & \textbf{Type} & \textbf{s} \\[2pt]
\hline
\noalign{\vskip 3pt}
9.1a & \raggedright Find the little girl and listen to the conversation. & N & 60 \\[0.5pt]
9.1b & \raggedright Complete the associated dialogue. & D & 30 \\[0.5pt]
10 & \raggedright Follow the quest marker to locate Pela. & N & 120 \\[0.5pt]
11 & \raggedright Advance the dialogue until the phone message interface opens. & D & 30 \\[0.5pt]
12 & \raggedright Complete the conversation in the phone message interface. & U & 30 \\[0.5pt]
13 & \raggedright Finish the subsequent conversation. & D & 30 \\[0.5pt]
14 & \raggedright Open the map and teleport to the next quest location. & N & 60 \\[0.5pt]
15 & \raggedright Locate and enter the cheap room. & N & 30 \\[0.5pt]
16 & \raggedright Interact with the clue inside the room. & N & 30 \\[0.5pt]
17 & \raggedright Complete the associated dialogue. & D & 30 \\[0.5pt]
18 & \raggedright Teleport back to the Administrative District. & N & 60 \\[0.5pt]
19 & \raggedright Locate the museum ticket inspector. & N & 60 \\[0.5pt]
20 & \raggedright Complete the conversation and enter the museum. & D & 30 \\[0.5pt]
21 & \raggedright Reach the target location and return the Meteor Stone fragment. & N & 30 \\[0.5pt]
22 & \raggedright Complete the final dialogue and finish the mission. & D & 30 \\[0.5pt]
\noalign{\vskip 2pt}
\noalign{\hrule height 0.7pt}
\end{tabular*}
\end{minipage}
\par\vspace{7pt}
\begin{minipage}{\linewidth}
\footnotesize
\textbf{Total: 30 milestones}\hfill\textbf{Time budget: 1,800 s}\par
\vspace{4pt}
\raggedright
\textbf{Types.} N: navigation (14), D: dialogue (14), C: combat (1), U: UI interaction (1).\par
\vspace{2pt}
\textbf{Notes.} IDs, order, and types follow the benchmark specification. Times are configured per-milestone budgets in seconds, not measured completion times.
\end{minipage}
\end{table}

\begin{table}[t]
\centering
\caption{\footnotesize \textbf{Bestial Ferocity.} The 55 required milestones span 40 parent steps and include three automaton-repair puzzles. Read down the left block, then the right block.}
\label{tab:bestial_milestones}
\vspace{5pt}
\footnotesize
\setlength{\tabcolsep}{2pt}
\renewcommand{\arraystretch}{1.06}
\begin{minipage}[t]{0.485\linewidth}
\vspace{0pt}
\centering
\textbf{First 28 milestones}\par
\vspace{4pt}
\begin{tabular*}{\linewidth}{@{}p{0.12\linewidth}@{\hspace{5pt}}p{0.66\linewidth}@{\extracolsep{\fill}}cr@{}}
\noalign{\hrule height 0.7pt}
\rule{0pt}{12pt}\textbf{ID} & \textbf{Milestone} & \textbf{Type} & \textbf{s} \\[2pt]
\hline
\noalign{\vskip 3pt}
1 & \raggedright Locate the unidentified NPC in Central Starskiff Haven. & N & 90 \\[0.5pt]
2 & \raggedright Complete the initial NPC dialogue. & D & 90 \\[0.5pt]
3 & \raggedright Teleport to Exalting Sanctum. & N & 90 \\[0.5pt]
4 & \raggedright Enter the Realm-Keeping Commission Chancery. & N & 90 \\[0.5pt]
5 & \raggedright Locate Jingyan. & N & 90 \\[0.5pt]
6 & \raggedright Complete the dialogue with Jingyan. & D & 120 \\[0.5pt]
7 & \raggedright Teleport to Scalegorge Waterscape. & N & 90 \\[0.5pt]
8 & \raggedright Reach the marked quest destination. & N & 90 \\[0.5pt]
9 & \raggedright Activate the Diting tracking trail. & N & 10 \\[0.5pt]
10 & \raggedright Follow the blue trail to the enemies. & N & 180 \\[0.5pt]
11 & \raggedright Advance the dialogue and defeat the enemies. & C & 180 \\[0.5pt]
12 & \raggedright Complete the post-encounter dialogue. & D & 90 \\[0.5pt]
13 & \raggedright Teleport to the Alchemy Commission. & N & 90 \\[0.5pt]
14 & \raggedright Reach Zhongshan and complete the dialogue. & N & 90 \\[0.5pt]
15 & \raggedright Talk to Zhongshan again. & D & 120 \\[0.5pt]
16 & \raggedright Reach the highlighted quest location. & N & 120 \\[0.5pt]
17 & \raggedright Open the phone and finish the messages. & U & 90 \\[0.5pt]
18 & \raggedright Find Zhongshan and continue the dialogue. & N & 90 \\[0.5pt]
19.1a & \raggedright Approach Heathmann to hear his request. & N & 90 \\[0.5pt]
19.1b & \raggedright Complete the dialogue with Heathmann. & D & 90 \\[0.5pt]
19.2a & \raggedright Approach Tonya to hear her request. & N & 90 \\[0.5pt]
19.2b & \raggedright Complete the dialogue with Tonya. & D & 90 \\[0.5pt]
19.3a & \raggedright Approach Benna to hear her request. & N & 90 \\[0.5pt]
19.3b & \raggedright Complete the dialogue with Benna. & D & 90 \\[0.5pt]
20 & \raggedright Finish the remaining Zhongshan dialogue. & D & 120 \\[0.5pt]
21 & \raggedright Teleport to the Silvermane Guard Restricted Zone. & N & 90 \\[0.5pt]
22 & \raggedright Locate Serval. & N & 120 \\[0.5pt]
23 & \raggedright Complete the dialogue with Serval. & D & 120 \\[0.5pt]
\noalign{\vskip 2pt}
\noalign{\hrule height 0.7pt}
\end{tabular*}
\end{minipage}
\hfill
\begin{minipage}[t]{0.485\linewidth}
\vspace{0pt}
\centering
\textbf{Final 27 milestones}\par
\vspace{4pt}
\begin{tabular*}{\linewidth}{@{}p{0.12\linewidth}@{\hspace{5pt}}p{0.66\linewidth}@{\extracolsep{\fill}}cr@{}}
\noalign{\hrule height 0.7pt}
\rule{0pt}{12pt}\textbf{ID} & \textbf{Milestone} & \textbf{Type} & \textbf{s} \\[2pt]
\hline
\noalign{\vskip 3pt}
24 & \raggedright Repair the Automaton Grizzly. & P & 180 \\[0.5pt]
25 & \raggedright Repair the Automaton Spider. & P & 180 \\[0.5pt]
26 & \raggedright Repair the Automaton Direwolf. & P & 180 \\[0.5pt]
27 & \raggedright Return to Scalegorge Waterscape. & N & 90 \\[0.5pt]
28 & \raggedright Follow the quest marker to Zhongshan. & N & 120 \\[0.5pt]
29 & \raggedright Complete the dialogue with Zhongshan. & D & 120 \\[0.5pt]
30.1a & \raggedright Approach Benna to discuss the mechanical creatures. & N & 90 \\[0.5pt]
30.1b & \raggedright Complete the dialogue with Benna. & D & 90 \\[0.5pt]
30.2a & \raggedright Approach Heathmann to discuss the mechanical creatures. & N & 90 \\[0.5pt]
30.2b & \raggedright Complete the dialogue with Heathmann. & D & 90 \\[0.5pt]
30.3a & \raggedright Approach Tonya to discuss the mechanical creatures. & N & 90 \\[0.5pt]
30.3b & \raggedright Complete the battle and its preceding and following dialogues. & C & 180 \\[0.5pt]
31 & \raggedright Reach Zhongshan and initiate dialogue. & N & 120 \\[0.5pt]
32 & \raggedright Complete the group dialogue. & N & 180 \\[0.5pt]
33 & \raggedright Complete the text-message conversation. & U & 180 \\[0.5pt]
34 & \raggedright Reach Nannan and initiate dialogue. & N & 120 \\[0.5pt]
35 & \raggedright Complete the Nannan and Zhongshan dialogue. & N & 120 \\[0.5pt]
36.1a & \raggedright Visit Benna to check on her situation. & N & 90 \\[0.5pt]
36.1b & \raggedright Complete the dialogue with Benna. & D & 90 \\[0.5pt]
36.2a & \raggedright Visit Tonya to check on her situation. & N & 90 \\[0.5pt]
36.2b & \raggedright Complete the dialogue with Tonya. & D & 90 \\[0.5pt]
36.3a & \raggedright Visit Heathmann to check on his situation. & N & 90 \\[0.5pt]
36.3b & \raggedright Complete the dialogue with Heathmann. & D & 90 \\[0.5pt]
37 & \raggedright Complete the dialogue with Zhongshan. & N & 180 \\[0.5pt]
38 & \raggedright Re-enter the Realm-Keeping Commission Chancery. & N & 90 \\[0.5pt]
39 & \raggedright Locate Jingyan. & N & 90 \\[0.5pt]
40 & \raggedright Complete the final dialogue with Jingyan. & D & 120 \\[0.5pt]
\noalign{\vskip 2pt}
\noalign{\hrule height 0.7pt}
\end{tabular*}
\end{minipage}
\par\vspace{7pt}
\begin{minipage}{\linewidth}
\footnotesize
\textbf{Total: 55 milestones}\hfill\textbf{Time budget: 6,040 s}\par
\vspace{4pt}
\raggedright
\textbf{Types.} N: navigation (32), D: dialogue (16), C: combat (2), U: UI interaction (2), P: puzzle (3).\par
\vspace{2pt}
\textbf{Notes.} IDs, order, and types follow the benchmark specification. Times are configured per-milestone budgets in seconds, not measured completion times.
\end{minipage}
\end{table}

\subsection{Minigame Coverage and Visualization}

\begin{figure*}[t]
    \centering
    \includegraphics[width=\linewidth]{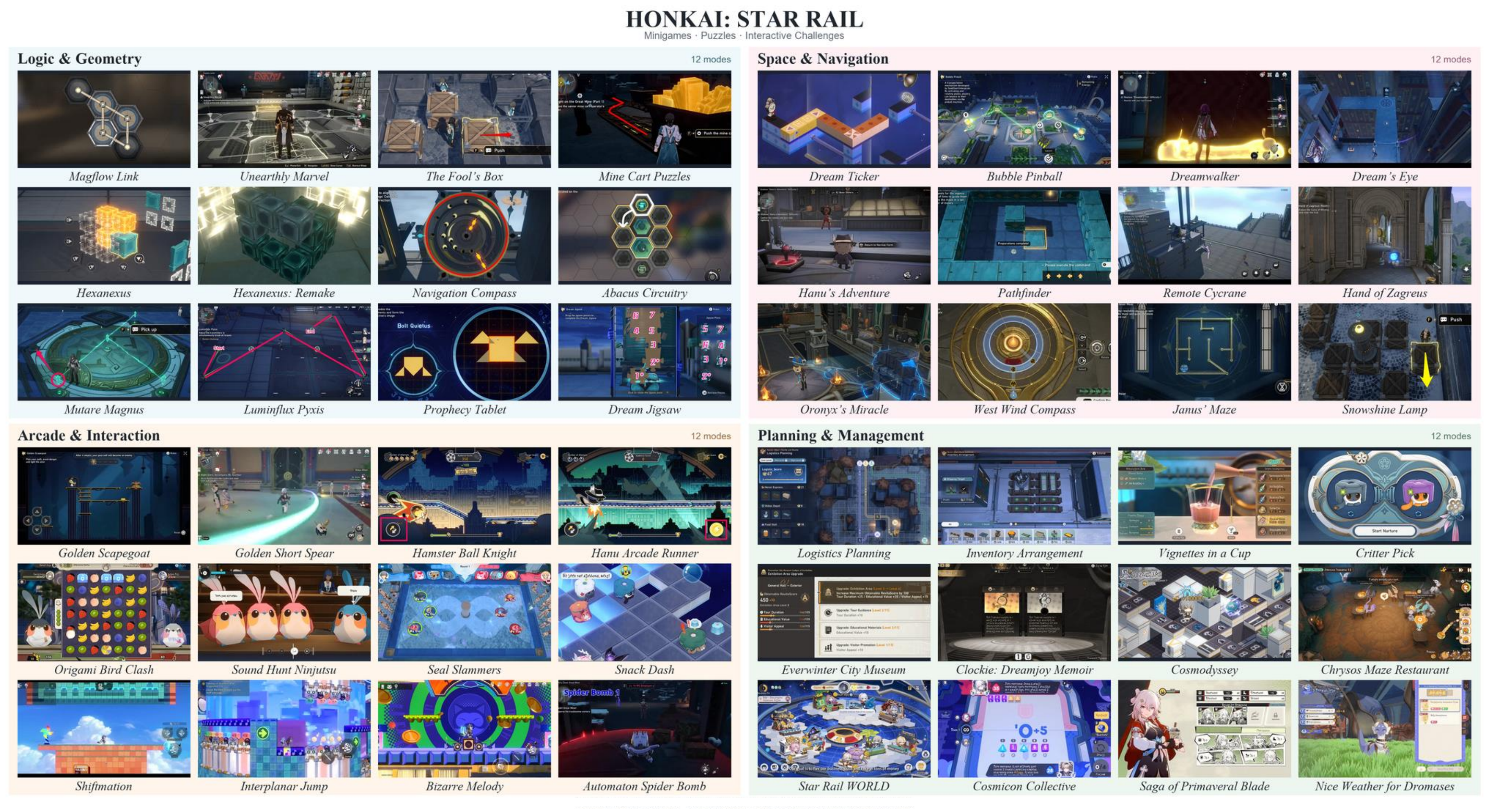}
    \caption{
    \textbf{Representative minigames and interaction challenges in the Honkai: Star Rail evaluation suite.}
    This extended screenshot gallery illustrates the range of game interactions in Honkai: Star Rail. It is a qualitative visualization rather than the scored minigame inventory. The quantitative benchmark contains seven scored minigame types and 40 tasks, summarized in Figure~\ref{fig:hsr_minigames} and Table~\ref{tab:minigame_results}.
    }
    \label{fig:hsr_minigames_all}
\end{figure*}

The scored minigame benchmark contains \textbf{40 tasks across seven types}: Magflow Link (ML), Navigation Compass (NC), Abacus Circuitry (AC), Hexanexus (HX), Janus's Hidden Passage (JHP), Golden Scapegoat (GS), and Janus's Maze (JM). The extended gallery in Figure~\ref{fig:hsr_minigames_all} illustrates other mechanics encountered in the game, but those examples are not additional scored types in the reported benchmark. Capability tags describe the main reasoning or control challenge for visualization and do not enter the accuracy calculation. Understanding diverse and complex gameplay mechanics is essential for general game agents, as modern commercial games such as Honkai: Star Rail typically combine multiple mechanics within a single game.

We compute accuracy by first averaging the success outcomes of valid attempts for each task, then averaging task scores within each game type, and finally taking an unweighted mean over the seven type-level scores. Unlike an any-success-at-three metric, this score retains fractional success rates when a task succeeds in only a subset of its attempts. The seven-type macro-average prevents categories with more tasks from dominating the result.

\paragraph{Extended Main-Story Evaluation.}
Beyond side quests and minigames, we further evaluate \ourmodel on the main storyline of Honkai: Star Rail, covering five major story arcs with 19 quest chains and 160 main-story mission entries. Across approximately 50 hours of gameplay evaluation, our model operates autonomously for over 95\% of the total evaluation time, demonstrating sustained gameplay proficiency across diverse environments, complex interactions, and long sequences of evolving objectives. This result highlights the potential of generalist game agents to support extended autonomous gameplay beyond isolated benchmark tasks. With further reinforcement learning and model scaling, generalist game agents could serve as foundational tools for large-scale automated evaluation and testing of large modern games.

\FloatBarrier

\section{Rollout Visualization and Qualitative Case Study}
\label{app:rl_rollouts}

The reward curves in Figure~\ref{fig:rl_reward} show how policy optimization progresses during training. To complement these aggregate measurements, we visualize how navigation and target interaction change across checkpoints. We compare the SFT initialization (CP 0) with five TA-GRPO checkpoints on a training task and a held-out development task. Each checkpoint is evaluated in 16 attempts from the same task and initial scene. The \textbf{Task successes} value on the left of each row reports the number of successful attempts, while twelve ordered frames depict one representative rollout. Executed actions and selected translated reasoning excerpts appear below the frames. The two views are complementary. The counts summarize reliability across attempts, and the screenshots make route selection, obstacle handling, and goal interaction easier to inspect.

\begin{figure*}[!t]
    \centering
    \includegraphics[width=0.98\textwidth]{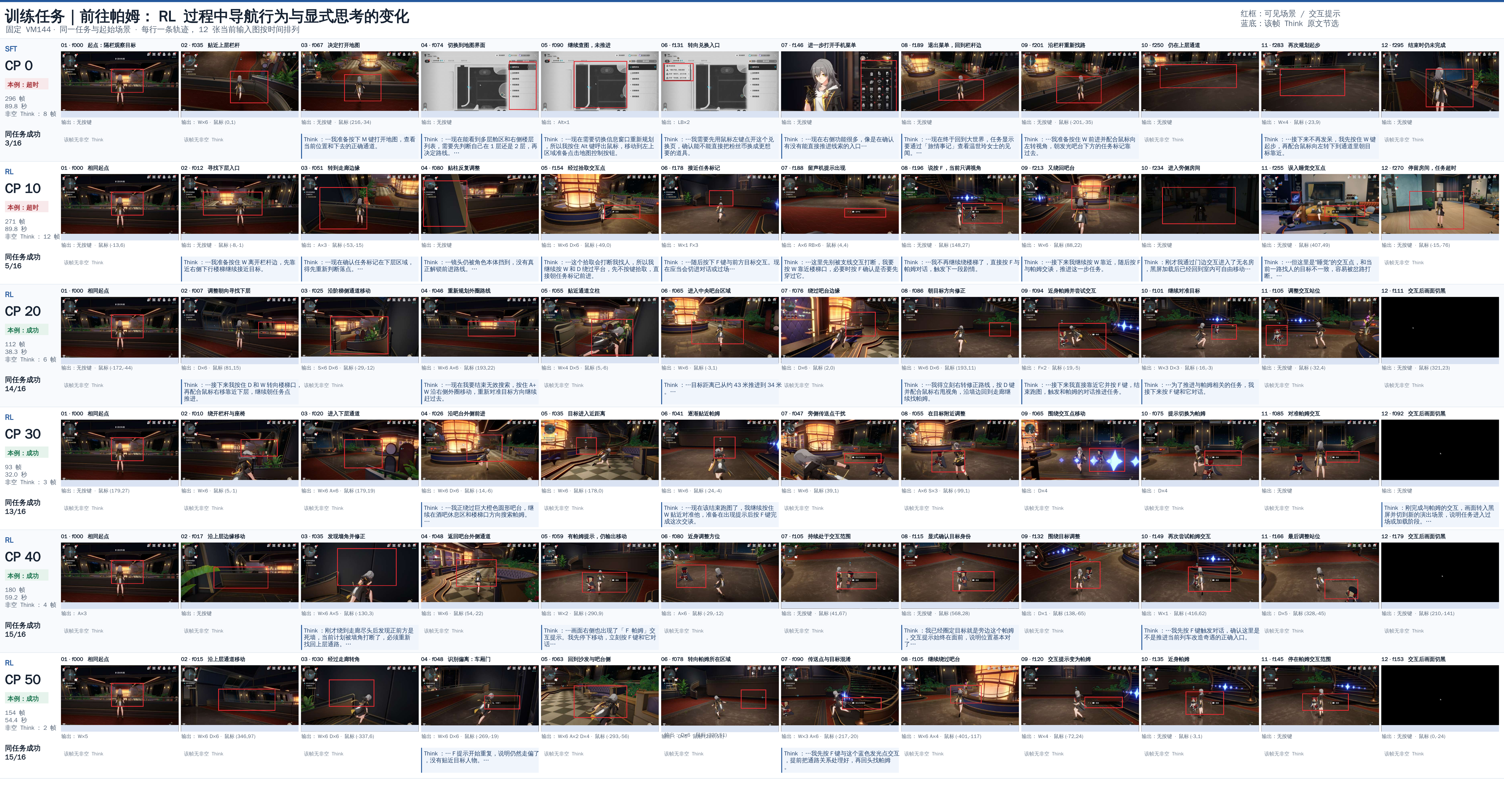}
    \caption{\textbf{Navigation across RL checkpoints on a training task (Reach Pom-Pom).} Each row shows one representative rollout with twelve temporally ordered observations, actions, and translated reasoning excerpts. \textbf{Task successes} reports successful attempts out of 16 for each checkpoint: 3, 5, 14, 13, 15, and 15 for CP 0-50, respectively. Red boxes mark scene details or interaction prompts, and blue shading marks reasoning excerpts. The illustrated CP 0 and CP 10 attempts time out, while CP 20-50 reach the target and complete the interaction. Frame counts and completion times describe only the displayed rollout.}
    \label{fig:rl_train_rollout}
\end{figure*}

\textbf{Training rollout: Reach Pom-Pom.} Figure~\ref{fig:rl_train_rollout} follows the policy through a multi-level interior with railings, corridors, and nearby interactable objects. The number of successful attempts grows from 3/16 for the SFT initialization to 14/16 at CP 20 and 15/16 at both CP 40 and CP 50. In the illustrated CP 20 rollout, the policy takes a route around the bar and completes the target interaction in 38.3 s. Later checkpoints also complete the displayed task. This case illustrates a useful progression from recognizing the destination to navigating the intervening geometry and executing the interaction that verifies completion. The visual examples and aggregated attempts together indicate more reliable goal-directed behavior after online training.

\begin{figure*}[!t]
    \centering
    \includegraphics[width=0.98\textwidth]{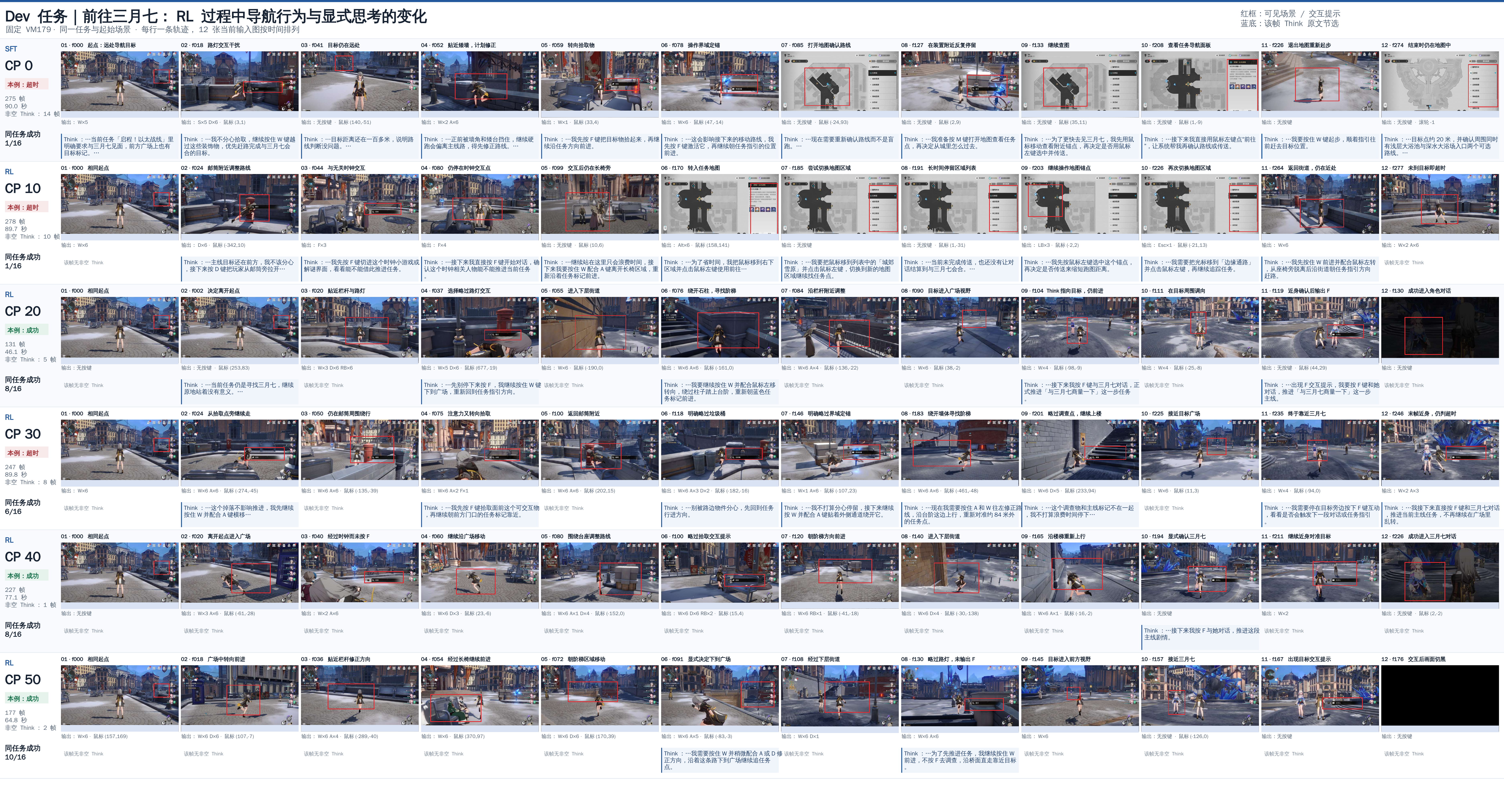}
    \caption{\textbf{Navigation across RL checkpoints on a development task (Reach March 7th).} Each row presents one representative rollout from the same task and initial scene, with twelve ordered observations, actions, and translated reasoning excerpts. \textbf{Task successes} counts successful attempts out of 16: 1, 1, 8, 6, 8, and 10 for CP 0-50, respectively. The illustrated CP 0, CP 10, and CP 30 attempts time out, whereas CP 20, CP 40, and CP 50 complete the target dialogue. Aggregate successes and representative trajectories provide complementary evidence of navigation and interaction progress.}
    \label{fig:rl_dev_rollout}
\end{figure*}

\textbf{Development rollout: Reach March 7th.} Figure~\ref{fig:rl_dev_rollout} examines whether similar behavior emerges on a held-out navigation objective. Successful attempts increase from 1/16 at the SFT initialization to 8/16 at CP 20 and 10/16 at CP 50. The illustrated CP 20 policy follows the lower street, navigates around a pillar, and initiates the required dialogue in 46.1\,s. CP 40 and CP 50 also complete the target interaction in their displayed rollouts. Because this task is not the RL training objective, the comparison provides complementary evidence that the learned navigation behavior can extend to a different destination and interaction sequence within the same game.

Across both cases, the increase in successful attempts is accompanied by more goal-directed behavior in the illustrated trajectories. Later policies more consistently select routes around obstacles, avoid unrelated interactions, and carry through to the final task-verifying action rather than stopping near the target location. The 16-attempt counts quantify the change in completion reliability, while the examples help interpret what that improvement looks like during execution. Together with the held-out side-quest results, they illustrate how online training can strengthen navigation and sustained goal pursuit.

\FloatBarrier

\section{Training Recipe and Reproducibility Details}
\label{app:training_recipe_details}

\subsection{Unified Verl-Based Training and Inference}
\label{app:unified_training_inference}

We extend Verl~\cite{sheng2024hybridflow} to use a shared game-interaction interface across CPT, SFT, and TA-GRPO. The training framework handles the stage-specific objectives and model updates, while Streaming-vLLM supplies autoregressive policy inference for interactive evaluation and online RL rollouts. The stages share the keyboard-mouse action grammar, observation-action history representation, and game-facing execution interface. CPT and SFT learn from recorded trajectories under their respective objectives, and TA-GRPO receives environment-verifiable feedback from interactions generated by the current policy. Consequently, the final policy is evaluated through the same type of executable control path used to collect online training experience.

\textbf{Consistency scope.} The system keeps action serialization, token boundaries, and environment timing consistent between training-stage evaluation and online policy interaction. It does not imply that every forward pass uses the same numerical kernel or cached context. Training computes gradients on supervised or rollout-derived sequences, whereas Streaming-vLLM uses stateful KV reuse, memory compaction, and architecture-dependent serving kernels. The RoPE phase adjustment in Appendix~\ref{app:rope_refresh} corrects retained cache positions but does not reconstruct the output of a fully re-encoded context. Consistency therefore refers to the policy interface and interaction semantics, not bitwise equality of training and inference logits.

\subsection{Training Configuration and Data Separation}

\textbf{Controlled CPT configurations.} The scaling study consists of four backbone sizes, five game mixtures, and six completed training budgets. It uses twenty primary Warmup-Stable-Decay (WSD) trunks, one per size-mixture pair. The shorter-budget checkpoints branch from their shared trunk and complete an independent decay phase within the stated total budget, yielding 120 evaluated configurations. All controlled runs use an 85\% gameplay and 15\% general vision-language token ratio, which is also retained for the separately trained HSR-augmented \ourmodel-CPT checkpoint. The CPT recipe uses BF16 AdamW with $\beta_1=0.9$, $\beta_2=0.95$, weight decay $0.01$, gradient clipping at $1.0$, and peak learning rates of $10^{-5}$ for the language model and $7\times10^{-6}$ for the unfrozen vision encoder. The WSD allocation is 3\% warmup, 87.3\% stable training, and 9.7\% cosine decay. The controlled grid uses 854$\times$480 observations, twenty visual turns per context, and approximately $2.3\times10^{5}$ supervised target tokens per update.

\textbf{Cross-stage implementation.} The same autoregressive device-level action representation is retained in CPT, SFT, and TA-GRPO, while instruction and memory conditioning are added in the later stages. The full stage-by-stage recipe, including the image resolution actually configured for each stage, is summarized in Table~\ref{tab:training_recipe}. For the real-time RL stage, the rollout-worker and training configuration are recorded separately from the controlled CPT grid. The short-term history uses 20 dialogue turns, the memory budget is 512 tokens, and each TA-GRPO segment is bounded by 20 dialogue turns with temporal factor $\beta=0.9$.

\textbf{Evaluation separation.} Training/validation/test splits are defined at the complete recording or task-source level before extracting overlapping action windows or multi-level examples. For interaction benchmarks, time-paused (TP) evaluation freezes the simulated environment during policy inference, whereas real-time (RT) evaluation lets the environment advance. These settings use the same action serialization but different timing semantics. Real-time performance comparisons should therefore report both the serving configuration and environment timing rather than treating model size or score as a proxy for latency.

\begin{table*}[t!]
\vspace{-6pt}
\begin{center}
\caption{
\textbf{Training recipes for \ourmodel.}
Stage I reports the controlled GameScaling reference on four Qwen3.5 Dense sizes. The HSR experiments use a separately trained \ourmodel-CPT checkpoint with additional \emph{Honkai: Star Rail} data and the same 85/15 gameplay-VLM token mixture, followed by SFT and TA-GRPO. RL run-dependent settings are reported for navigation (nav) and puzzle training (puzzle). RL token totals count generated response tokens, whereas supervised target tokens per update follow a different accounting convention.
}
\label{tab:training_recipe}

\resizebox{\linewidth}{!}{
\begin{tabular}{lccc}
& \texttt{Stage I} & \texttt{Stage II} & \texttt{Stage III} \\
\toprule[0.95pt]
Phase & CPT & SFT & RL \\

\midrule[0.6pt]
\multicolumn{4}{c}{\textbf{Model and Context Configuration}} \\
\midrule[0.6pt]
\textbf{Qwen3-VL (supported)} & \multicolumn{3}{c}{Dense: 2B, 4B, 8B, 32B, MoE: 30B-A3B} \\
\textbf{Qwen3.5 (supported)} & \multicolumn{3}{c}{Dense: 0.8B, 2B, 9B, 27B, MoE: 35B-A3B} \\
Controlled CPT Scales & 0.8B / 2B / 9B / 27B & -- & -- \\
Image Resolution & $854\times480$ & $854\times480$ & $854\times480$ \\
Visual Observation Rate & 5 Hz & 5 Hz & 5 Hz \\
Low-Level Control Rate & 30 Hz & 30 Hz & 30 Hz \\
Local Visual History & 20 observation turns & 20 observation turns & 20 observation turns \\
Maximum Context Length & 45K & 45K & 23K \\
Short-Term History Budget $L_H$ & -- & 20 dialogue turns & 20 dialogue turns \\
Long-Term Memory Budget $L_M$ & -- & 512 tokens & 512 tokens \\

\midrule[0.6pt]
\multicolumn{4}{c}{\textbf{Training Hyper-Parameters}} \\
\midrule[0.6pt]
Initialization & Base & CPT & SFT \\
Optimizer & AdamW & AdamW & Adam (MCore) \\
Optimizer Hyper-Parameters & $\beta_1,\beta_2=0.9,0.95$ & $\beta_1,\beta_2=0.9,0.95$ & $\beta_1,\beta_2=0.9,0.999$ \\
\multirow{2}{*}{Learning Rate}
& $1\times10^{-5}$ (LLM) & $1\times10^{-5}$ (LLM) & $1\times10^{-6}$ (LLM) \\
& $7\times10^{-6}$ (vision encoder) & $1\times10^{-5}$ (vision encoder) & $1\times10^{-6}$ (vision encoder) \\
Weight Decay & 0.01 & 0.1 & 0.01 \\
Gradient Clipping & 1.0 & 1.0 & 1.0 \\
Training Precision & BF16 & BF16 & BF16 \\
Learning Rate Scheduler & Warmup-Stable-Decay & Warmup-Decay & Constant \\
Warmup / Stable / Decay & 3\% / 87.3\% / 9.7\% & 5\% / -- / 95\% & 0\% / 100\% / 0\% \\
Minimum LR Ratio & 0.1 & 0.1 & -- (constant LR) \\
Global Supervised Tokens / Update & $2.3\times10^{5}$ & $2.3\times10^{5}$ & -- \\
Training Iterations & 6,839 & 1,696 & 61 (nav) / 53 (puzzle) \\
Optimizer Updates & -- & -- & 122 (nav) / 106 (puzzle) \\
Total Training Tokens & $5.45\times10^{10}$ & $3.9\times10^{8}$ & $3.1\times10^{7}$ (nav) / $2.8\times10^{7}$ (puzzle) \\

\midrule[0.6pt]
\multicolumn{4}{c}{\textbf{Stage-Specific Configuration}} \\
\midrule[0.6pt]
Random Key Remapping & \checkmark & -- & -- \\
Repeat Decay $\gamma$ & 0.9 & -- & -- \\
Multi-level Supervision & -- & \checkmark & \checkmark \\
Cross-Context Memory & -- & \checkmark & \checkmark \\
Parallel Environments & -- & -- & 32 Windows VMs \\
Inference Services & -- & -- & Sharded pool (64 H200) \\
Tasks per Iteration & -- & -- & 8 (nav) / 16 (puzzle) \\
Rollouts per Iteration & -- & -- & 128 \\
Rollouts per Task & -- & -- & 16 (nav) / 8 (puzzle) \\
Group Size $G$ & -- & -- & 8 \\
Maximum Segment Length & -- & -- & 20 dialogue turns \\
Temporal Factor $\beta$ & -- & -- & 0.9 \\
Policy Clip $\epsilon_{\mathrm{clip}}$ & -- & -- & 0.2 \\
Updates per Rollout Batch & -- & -- & 2 (disjoint half-batches) \\

\midrule[0.6pt]
\multicolumn{4}{c}{\textbf{Training Data}} \\
\midrule[0.6pt]
Primary Data & Cross-Game $+$ General VLM & Multi-Level HSR & Online Interaction \\
Supervision & Actions & Reasoning + actions & Verifiable binary task rewards \\
Data Size & $\sim$2.4K h & $\sim$1K h (annotated pool) & $\sim$7.8K (nav) / $\sim$6.8K (puzzle) rollouts \\
\bottomrule[0.95pt]
\end{tabular}
}
\end{center}
\vspace{-4pt}
\end{table*}

\subsection{Execution Protocol and Reproducibility}
\label{app:runtime_manifest}

\textbf{Rollout and evaluation protocol.}
Interactive inference produces six keyboard-mouse control blocks from the current visual observation, task instruction, and memory context. The client executes these blocks while the environment continues to advance. Online RL runs across 32 parallel game environments served by a sharded Streaming-vLLM pool with 64 NVIDIA H200 GPUs. Each iteration gathers 128 rollouts in groups of eight from the active navigation or puzzle task pool. Two disjoint rollout subsets support the policy updates, and environment task verification provides rewards without a learned process reward model. SFT and RL evaluations use matched task initializations, milestone verifiers, action semantics, and inference settings as described in Sec.~\ref{sec:exp_rl}.

\textbf{Reproducibility boundary.}
After the two policy updates, the resulting checkpoint is synchronized across the inference services, and the next rollout iteration starts only after the workers acknowledge the expected policy version. Our run audit records the trainer and inference revisions, checkpoints, action serialization, tokenizer, image preprocessing, and environment settings. Inference uses BF16 with a maximum generation length of 512 tokens and temperature 0.9. The game screenshots are captured at $1280\times720$ and resized to $854\times480$ for model input. TA-GRPO uses 20-turn segments and temporal factor $\beta=0.9$. These settings, together with the checkpoint and service synchronization protocol, specify the interaction conditions used in the reported experiments.

\end{document}